\documentclass[10pt,twocolumn,letterpaper]{article}
\pdfoutput=1
\usepackage{cvpr}              %

\usepackage{nicefrac}
\usepackage{graphicx, multicol, afterpage, amsmath, tikz, pgfplots, caption, subcaption, amsfonts, bm, breakcites, amssymb, amsmath, algorithm, algpseudocode, wrapfig, setspace, xspace, soul, multirow, booktabs, arydshln,soul,bm, mathrsfs, wrapfig, caption, adjustbox}
\usepackage{pifont}
\usepackage[dvipsnames]{xcolor}
\usepackage{comment}
\usepackage{float}

\algrenewcommand\alglinenumber[1]{\scriptsize #1:}

\def\ie{\emph{i.e.}\xspace} 
\def\eg{\emph{e.g.}\xspace} 
\def\wrt{w.r.t.\xspace}

\def\vs{\emph{vs.}\xspace}
\newcommand{\quotes}[1]{``#1''} 

\definecolor{tablegrey}{rgb}{.61, .61, .61}
\definecolor{mygreen}{RGB}{77,175,74}
\definecolor{myblue}{RGB}{55,126,184}
\definecolor{skyblue}{RGB}{117,187,253}
\definecolor{myred}{RGB}{228,26,28}
\definecolor{blue_palette}{HTML}{008374}
\definecolor{yellow_palette}{HTML}{f7b32b} %
\newcommand{\sgdbox}[1]{\colorbox{blue_palette!30}{#1}}
\newcommand{\sambox}[1]{\colorbox{yellow_palette!65}{#1}}

\newcommand{\cifarten}{\textsc{Cifar10}\xspace}
\newcommand{\cifar}{\textsc{Cifar100}\xspace}
\newcommand{\gld}{\textsc{Landmarks-User-160k}\xspace}

\newcommand{\ours}{\textsc{FedGloSS}\xspace}
\newcommand{\ourslong}{Federated Global Server-side Sharpness}
\newcommand{\ourslongbold}{\textbf{Fed}erated \textbf{Glo}bal \textbf{S}erver-side \textbf{S}harpness}
\newcommand{\naiveours}{\textsc{NaiveFedGloSS}\xspace}

\newcommand{\fedsmoo}{\textsc{FedSmoo}\xspace}
\newcommand{\scaffold}{\textsc{Scaffold}\xspace}
\newcommand{\fedgamma}{\textsc{FedGamma}\xspace}
\newcommand{\fedspeed}{\textsc{FedSpeed}\xspace}
\newcommand{\fedsam}{\textsc{FedSam}\xspace}
\newcommand{\feddyn}{\textsc{FedDyn}\xspace}
\newcommand{\fedavg}{\textsc{FedAvg}\xspace}
\newcommand{\fedopt}{\textsc{FedOpt}\xspace}
\newcommand{\fedavgm}{\textsc{FedAvgM}\xspace}
\newcommand{\fedprox}{\textsc{FedProx}\xspace}

\newcommand{\sam}{\textsc{Sam}\xspace}
\newcommand{\sgd}{\textsc{Sgd}\xspace}
\newcommand{\adabest}{\textsc{AdaBest}\xspace}

\newcommand{\cmark}{\ding{51}}%
\newcommand{\xmark}{\ding{55}}%

\DeclareMathOperator{\heps}{\hat{\pmb{\epsilon}}}
\DeclareMathOperator{\teps}{\Tilde{\pmb{\epsilon}}}
\DeclareMathOperator{\beps}{\pmb{\epsilon}}
\DeclareMathOperator{\w}{\pmb{w}}
\DeclareMathOperator{\W}{\pmb{W}}
\DeclareMathOperator{\tw}{\Tilde{\pmb{w}}}
\DeclareMathOperator{\X}{\mathcal{X}}
\DeclareMathOperator{\Y}{\mathcal{Y}}
\DeclareMathOperator{\C}{\mathcal{C}}
\DeclareMathOperator{\D}{\mathcal{D}}
\DeclareMathOperator{\F}{\mathcal{F}}

\DeclareMathOperator*{\argmax}{argmax}
\DeclareMathOperator*{\argmin}{argmin}

\definecolor{cvprblue}{rgb}{0.21,0.49,0.74}
\usepackage[pagebackref,breaklinks,colorlinks,allcolors=cvprblue]{hyperref}

\def\paperID{4978} %
\def\confName{CVPR}
\def\confYear{2025}

\title{Beyond Local Sharpness: Communication-Efficient Global Sharpness-aware Minimization for Federated Learning}

\author{Debora Caldarola$^{1,}$\thanks{Corresponding author: \texttt{debora.caldarola@polito.it}}, Pietro Cagnasso$^1$, Barbara Caputo$^1$, Marco Ciccone$^{2,}$\thanks{Work mainly carried out while at Politecnico di Torino}\\
$^1$Politecnico di Torino, $^2$ Vector Institute}

\begin{document}
\maketitle
\begin{abstract}
Federated learning (FL) enables collaborative model training with privacy preservation. Data heterogeneity across edge devices (clients) can cause models to converge to sharp minima, negatively impacting generalization and robustness. Recent approaches use client-side sharpness-aware minimization (SAM) to encourage flatter minima, but the discrepancy between local and global loss landscapes often undermines their effectiveness, as optimizing for local sharpness does not ensure global flatness.  
This work introduces \ours (\ourslongbold), a novel FL approach that prioritizes the optimization of global %
sharpness on the server, using SAM. To reduce communication overhead, \ours cleverly approximates sharpness using the previous global gradient, eliminating the need for additional client communication. Our extensive evaluations demonstrate that \ours consistently achieves flatter minima and better performance compared to state-of-the-art FL methods in various federated vision benchmarks. Code available at \href{https://github.com/pietrocagnasso/fedgloss/}{github.com/pietrocagnasso/fedgloss}.
\end{abstract}
    
\section{Introduction}
\label{sec:intro}

Federated Learning (FL) \citep{mcmahan2017communication} provides a powerful framework to collaboratively train machine learning models on private data distributed across multiple endpoints. Unlike traditional methods, FL enables edge devices (\emph{clients}), like smartphones or IoT (Internet of Things) hardware, to train a shared model without compromising their sensitive information. This is achieved through communication rounds, where clients independently train on their local data and then exchange updated model parameters with a central server, preserving data privacy. The optimization on the server side relies on \textit{pseudo-gradients} \citep{reddi2020adaptive}, \ie, the average of the difference between the global model and the client's update, which serve as an estimate of the true global gradient on the overall dataset. This approach holds immense potential for privacy-sensitive applications, proving its value in areas like healthcare \citep{liu2021feddg,antunes2022federated,rauniyar2023federated,nevrataki2023survey}, finance \citep{nevrataki2023survey}, autonomous driving \citep{fantauzzo2022feddrive,shenaj2023learning,miao2023fedseg}, IoT \citep{zhang2022federated}, and more \citep{li2020review,wen2023survey}. However, the real-world deployment of FL presents unique challenges stemming from data heterogeneity and communication costs \citep{li2020federated}. Clients gather their data influenced by various factors such as personal habits or geographical locations, leading to inherent differences across devices \citep{kairouz2021advances,hsu2020federated,shenaj2023learning}. This results in the global model suffering from degraded performance and slower convergence \citep{li2019convergence,karimireddy2020mime,karimireddy2020scaffold,caldarola2022improving}, with instability emerging as client-specific optimization paths diverge from the global one.
This phenomenon, known as \emph{client drift} \citep{karimireddy2020scaffold}, limits the model's ability to generalize to the overall underlying distribution. 

\begin{figure}[t]
    \centering
    \vspace{-7pt}
    \captionsetup{font=scriptsize}
    \captionsetup[sub]{font=scriptsize}
    \subfloat[][\cifarten \\$\alpha=0$]{\includegraphics[width=.25\linewidth]{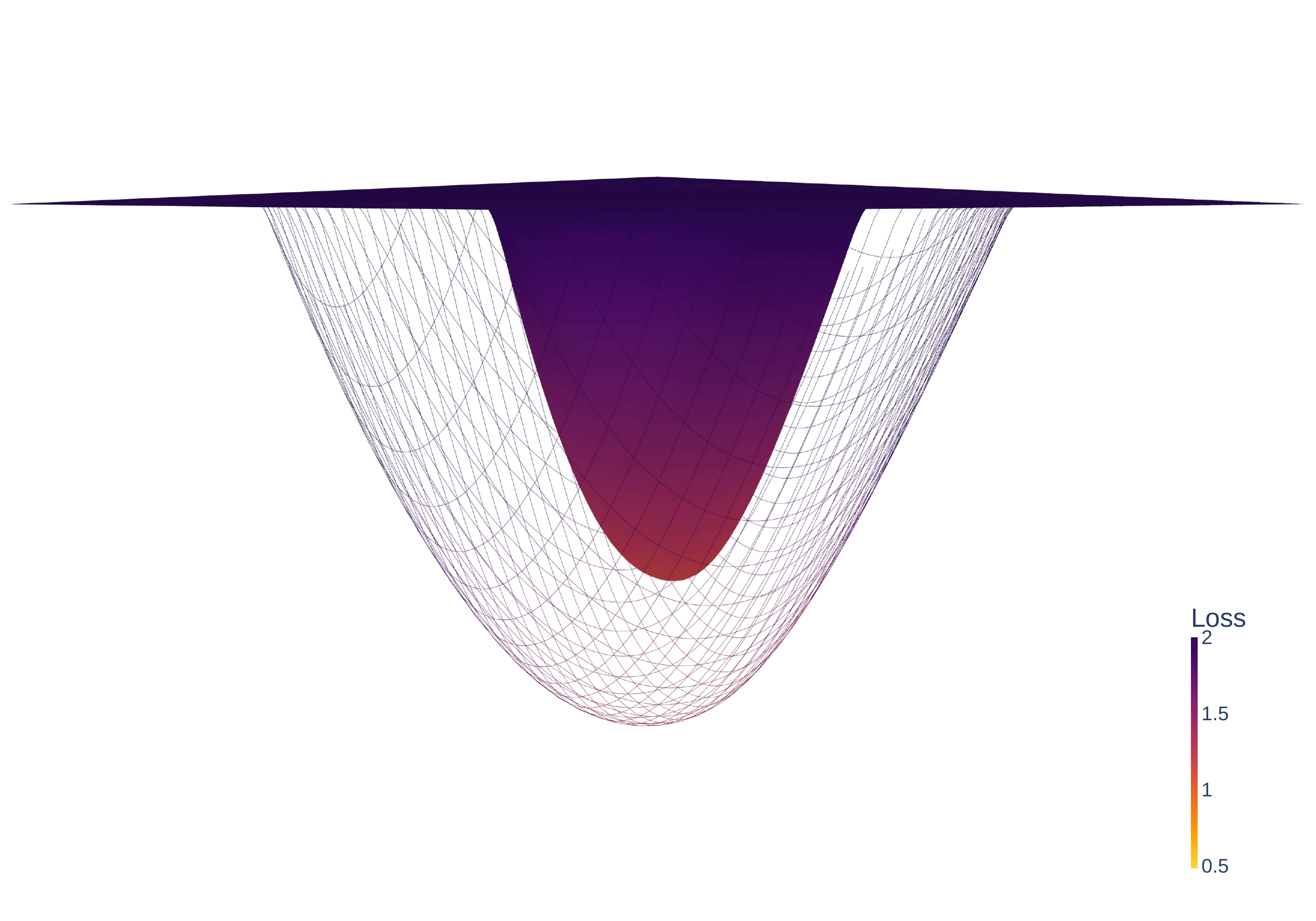}} \hfill
    \subfloat[][\cifarten \\$\alpha=0.05$]{\includegraphics[width=.25\linewidth]{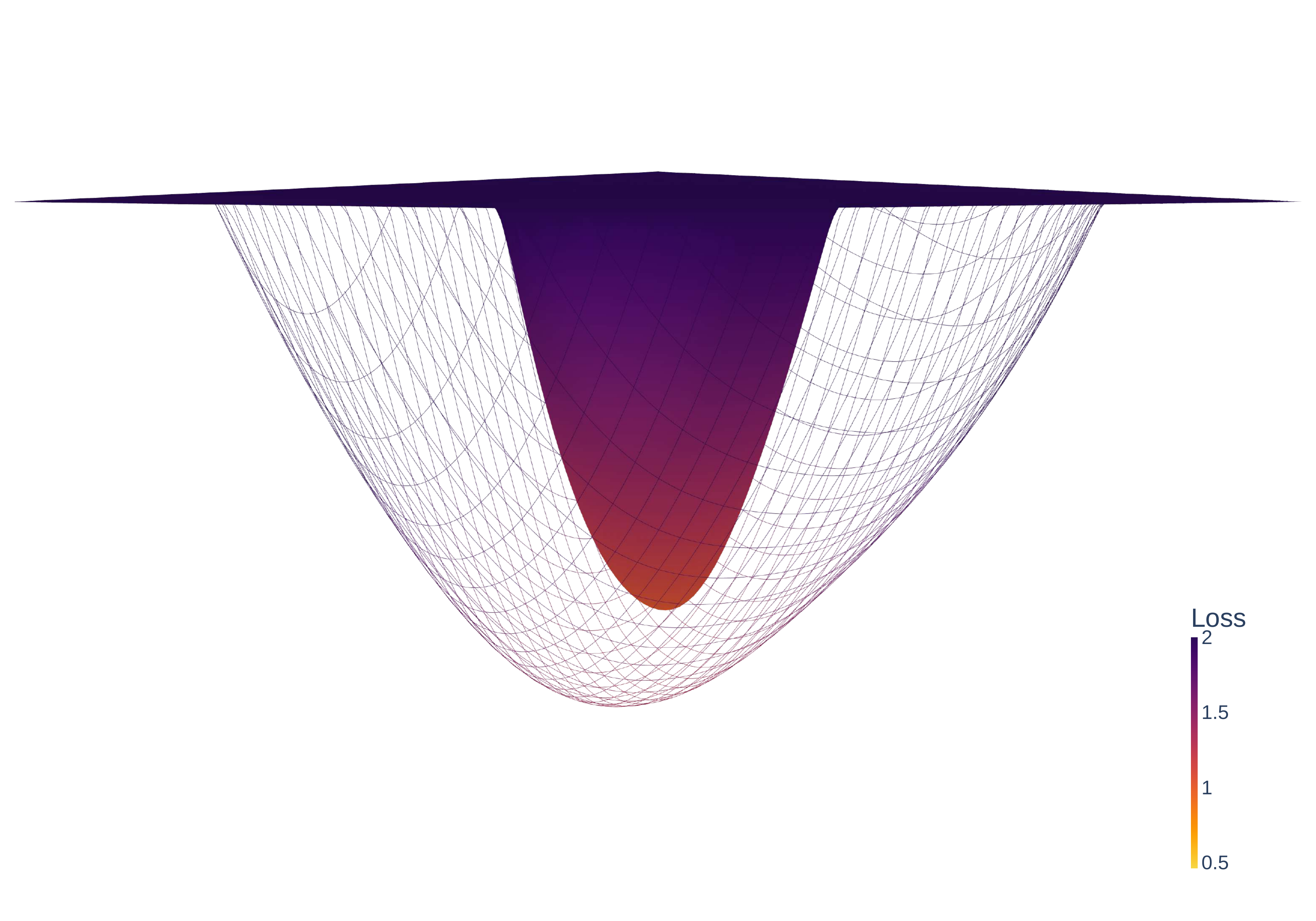}} \hfill
    \subfloat[][\cifar \\$\alpha=0$]{\includegraphics[width=.25\linewidth]{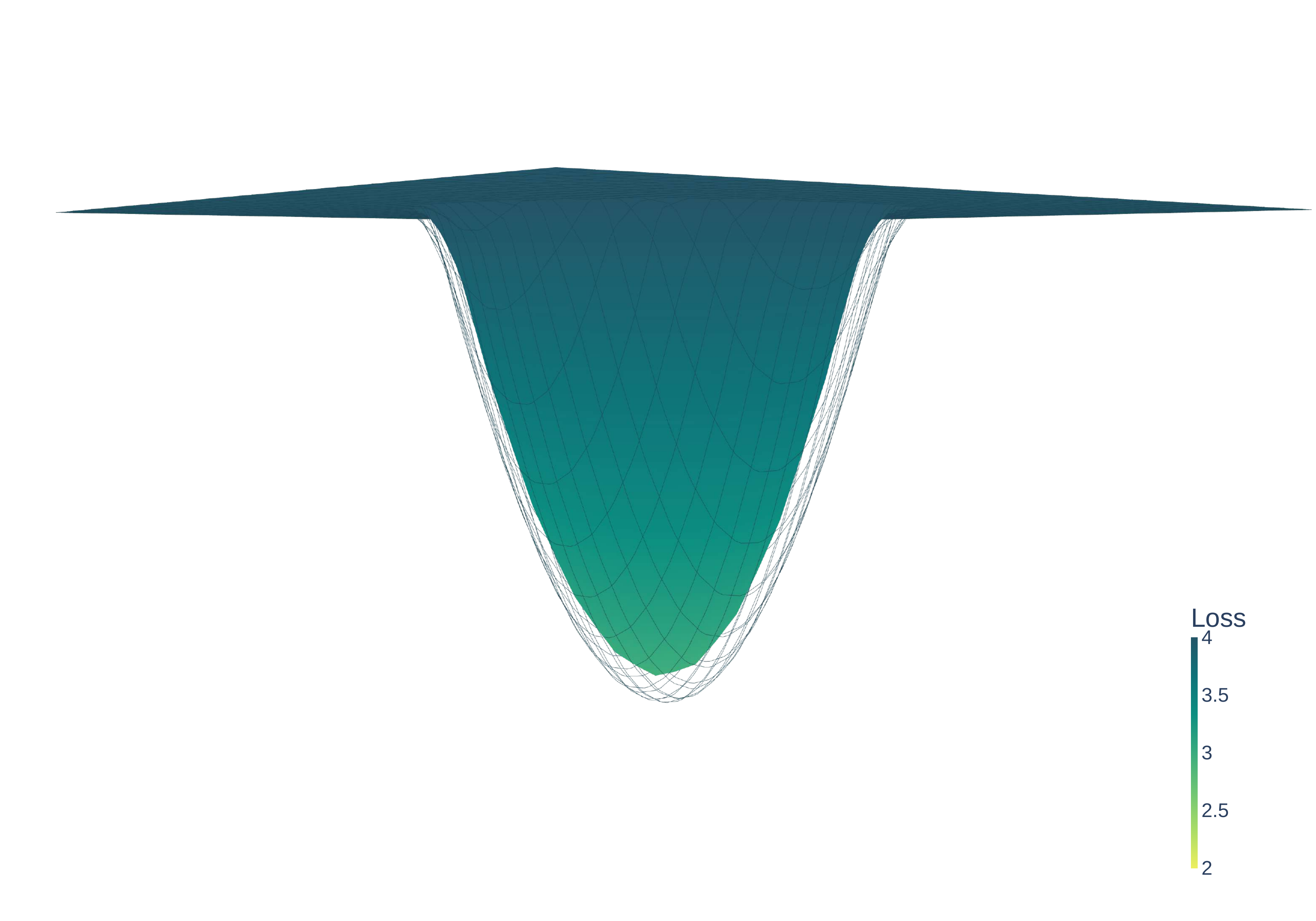}} \hfill
    \subfloat[][\cifar \\$\alpha=0.5$]{\includegraphics[width=.25\linewidth]{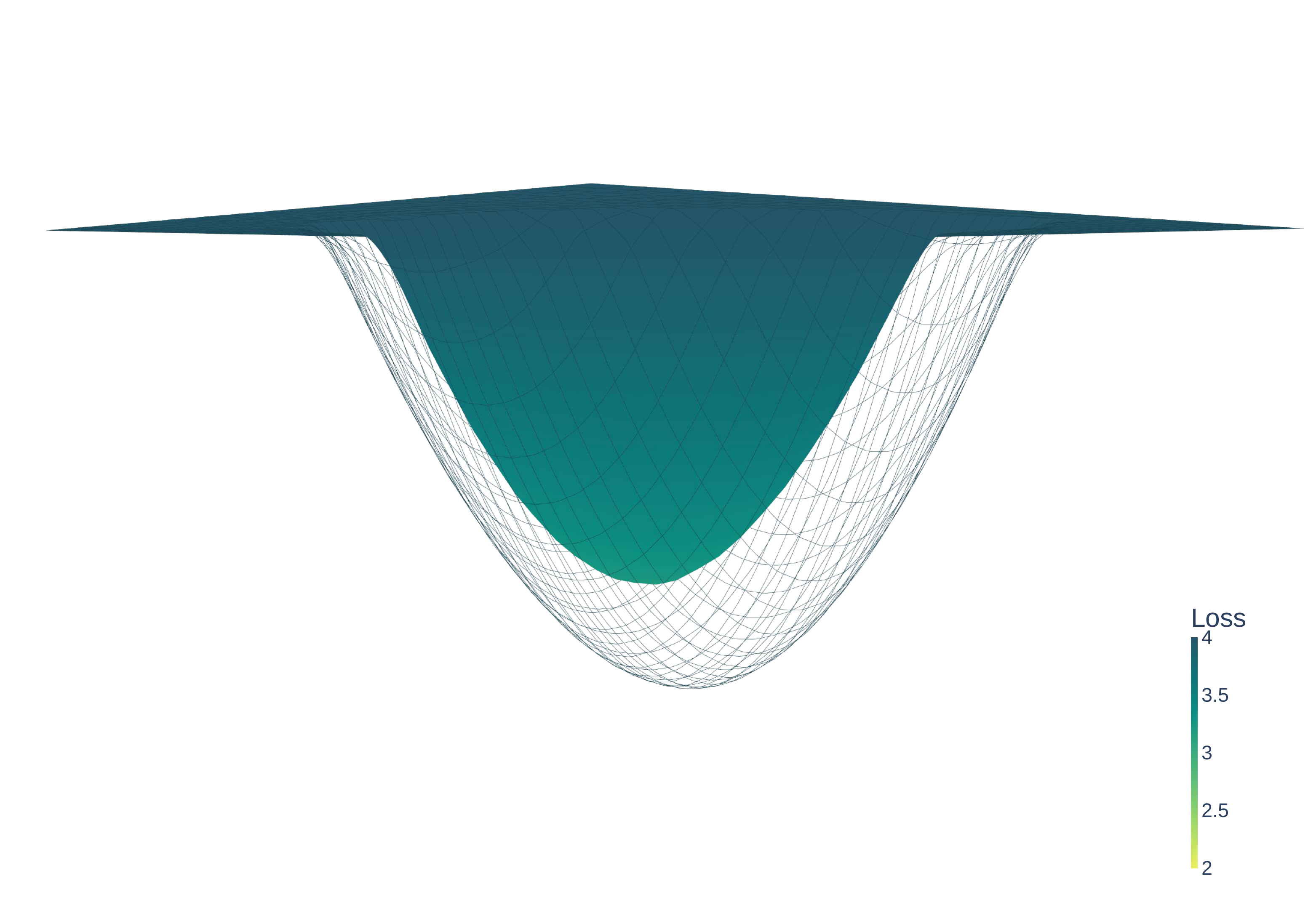}}
    \vspace{-5pt}
    \caption{{Comparison of \textbf{\fedavg} (\textit{solid}) and \textbf{\fedsam} (\textit{net}) loss landscapes with varying degrees of data heterogeneity ($\alpha$) on the \textsc{Cifar} datasets. \textbf{\fedsam's effectiveness in converging to \textit{global} flat minima is highly influenced by the data heterogeneity}, where higher heterogeneity ($\alpha \rightarrow 0$) leads to sharper minima, %
    and the complexity of the task, \eg, higher sharpness for the more complex \cifar. This highlights the importance of optimizing global sharpness. Model: CNN.%
    }}
    \label{fig:landscapes_heterogeneity}
    \vspace{-20pt}
\end{figure}

While many FL approaches focus on mitigating client drift through client-side regularization~\citep{li2020federated_fedprox,acar2021federated,varno2022adabest}, a recent trend leverages the geometry of the loss landscape to improve generalization~\citep{caldarola2022improving,qu2022generalized,sun2023fedspeed,dai2023fedgamma,sun2023dynamic}. These methods build upon the notion that convergence to sharp minima correlates with poor generalization \citep{hochreiter1997flat,keskar2016large, Jiang2020Fantastic}.  
\fedsam \citep{caldarola2022improving,qu2022generalized} employs Sharpness-aware Minimization (\sam) \citep{foret2020sharpness} in local training to guide clients toward flatter loss regions, enhancing  %
 global performance. 
{This comes at the cost of increased client-side computation, since \sam requires two forward/backward passes for each local optimization step: 
a gradient ascent step to compute the maximum sharpness and a descent step for sharpness and loss value minimization. %
} 
Although \fedsam and its variants~\citep{sun2023fedspeed, dai2023fedgamma} demonstrated their effectiveness in various settings, they rely solely on local flatness, assuming that minimizing sharpness locally leads to a globally flat minimum\footnote{We use the term \quotes{global flat minima} to refer to local minima within the global (\ie, server-side) loss landscape.}. 
However, in real-world scenarios with significant data heterogeneity, there can be {substantial discrepancies between local and global loss landscapes} and {optimizing for local sharpness does not guarantee the global model will reside in a flat region} (\cref{fig:landscapes_heterogeneity}). 
Addressing these limitations, \fedsmoo \citep{sun2023dynamic} uses the alternating direction method of multipliers (ADMM) \citep{boyd2011distributed} to include global sharpness information in \sam's local training. While this approach reduces the inconsistency between local and global geometries, it increases communication costs by {requiring double the bandwidth} in each round. This hinders its real-world applicability, as FL relies on minimizing communication overhead (\ie,  message size and exchange frequency) to avoid network congestion and account for potential connection failures.

Given the limitations of existing methods, achieving global flat minima while ensuring communication efficiency in heterogeneous FL remains a critical challenge. To address this, 
we propose \ours (\ourslongbold) that
directly \textbf{optimizes global sharpness by using \sam on the server side}, avoiding additional exchanges over the network. {Such adaptation is not straightforward, as \sam would require dual exchanges with each client set per round to solve its optimization problem. Instead, \ours approximates the sharpness measure using available 
{previous pseudo-gradients}}. %
As a result, it facilitates faster training and keeps communication efficiency. To summarize, our core contributions are the following:
\begin{itemize}
    \item {Empirical proof of local-global discrepancies}: we provide the first empirical evidence showing the limitations of approaches that focus solely on local sharpness. Our analysis highlights the {inconsistency between local and global loss geometries} even when using sharpness-aware approaches like \fedsam, demonstrating that local flatness does not necessarily ensure a flat global minimum. While reaching flat global solutions in simpler problems, we show that their effectiveness diminishes as data complexity and heterogeneity increase (\cref{fig:landscapes_heterogeneity}). 
    \item To bridge this gap and motivated by {communication efficiency}, our \ours algorithm directly optimizes for {global sharpness} on the server using \sam, reducing the communication overhead and the clients' computational costs compared to previous works. %
    \ours consistently achieves flatter minima and outperforms state-of-the-art methods across various vision benchmarks. %
    \item We show the importance of aligning global and local solutions and illustrate how \sam, especially on the server side, enables {effective ADMM use in FL}. 
    While typically ADMM-based methods suffer from parameter explosion \cite{varno2022adabest}, we show that by targeting flat minima, \sam encourages smaller gradient steps and minimal weight updates, leading to a significantly more stable algorithm.
\end{itemize}

\section{Related works}
\label{sec:related}
\paragraph{Federated framework.} In the last few years, Federated Learning (FL) \citep{mcmahan2017communication}  garnered significant attention from both the machine learning and computer vision communities. While the former has primarily focused on optimizing FL algorithms and guaranteeing their convergence \citep{li2019convergence,acar2021federated,reddi2020adaptive}, the latter has explored its applications in real-world settings, spanning diverse domains like autonomous driving \citep{fantauzzo2022feddrive,shenaj2023learning,miao2023fedseg} and healthcare \citep{liu2021feddg}. The key appeal of FL lies in its ability to efficiently learn from privacy-protected, distributed data while complying with regulations and leveraging edge resources. %
Real-world deployments of FL range across both \emph{cross-silo} and \emph{cross-device} settings \citep{kairouz2021advances}. This work focuses on the latter, with up to millions of individual devices at the network edge, with typically limited data and computational power, and potential unavailability due to battery life or network connectivity issues. 
User-specific factors like geographical location, capturing devices and daily habits introduce inherent \textit{bias} and \textit{statistical heterogeneity} into the local datasets. 
In this setting, \ours aims to learn a global model that {generalizes} to the overall data distribution under statistical heterogeneity {without increasing communication complexity, unlike other algorithms for local-global consistency in heterogeneous FL}. %

\noindent \textbf{Flatness search in FL.} \xspace Recent research has explored the connection between loss landscape geometry and generalization in heterogeneous FL. Studies suggest that convergence to sharp minima might hinder generalization performance \citep{hochreiter1997flat,keskar2016large, petzka2021relative}. \sam (Sharpness-Aware Minimization) \citep{foret2020sharpness} tackles this issue by guiding the optimization toward flatter regions, seeking minima that exhibit both low loss and low sharpness.  %
\fedsam \citep{caldarola2022improving,qu2022generalized} deploys \sam in local training, marking the first step toward leveraging loss surface geometry in FL to reduce discrepancies between local and global objectives, ultimately improving the global model's generalization ability. 
Following its success, %
\fedspeed \citep{sun2023fedspeed} uses perturbed gradients as \sam to reduce local overfitting, \fedgamma \citep{dai2023fedgamma} combines the stochastic variance reduction of \scaffold with \sam and \citet{shi2023make} show \fedsam's effectiveness in mitigating the negative effects of differential privacy. However, these approaches  rely on \textit{local} sharpness information, assuming its minimization %
directly translates to a globally flat minimum. This may not always be true, as we hypothesize discrepancies may exist between the geometries of local and global losses. Optimizing local sharpness alone does not guarantee a server model residing in a flat region of the \textit{global} loss landscape (\cref{fig:landscapes_heterogeneity}). Addressing these limitations, \fedsmoo \citep{sun2023dynamic} applies ADMM \citep{boyd2011distributed} to the sharpness measure to enforce global and local consistency. This adds communication overhead, doubling the message size in each round and hindering its real-world practicality. In contrast, our work focuses on {minimizing global sharpness} while maintaining {communication efficiency}. Lastly, building on Stochastic Weight Averaging \citep{izmailov2018averaging}, other works \citep{caldarola2022improving,caldarola2023window} use a window-based average of global models across rounds to reach wider minima. Being agnostic to the underlying optimization algorithm, they remain orthogonal to our approach. %

\noindent \textbf{Heterogeneity in FL.} \xspace The de-facto standard algorithm for FL is \fedavg \citep{mcmahan2017communication}, which updates the global model with a weighted average of the clients' parameters. However, \fedavg struggles when faced with heterogeneous data distributions, leading to performance degradation and slow convergence due to the local optimization paths diverging from the global one \citep{karimireddy2020scaffold}. 
\citet{reddi2020adaptive} shows \fedavg is equivalent to applying Stochastic Gradient Descent (\sgd) \cite{ruder2016overview} with a unitary learning rate on the server side, using the difference between the initial global model parameters and the clients' updates as \textit{pseudo-gradient}, opening the door to alternative optimizers beyond \sgd to improve performance and convergence speed. %
Building on this intuition, this work proposes {\sam} \citep{foret2020sharpness} {as a server-side optimizer} to enhance generalization by converging toward \emph{global} flat minima. Since \sam requires two optimization steps per iteration, a direct adaptation to the FL setting would double communication exchanges between clients and server; \ours overcomes this limitation and maintains communication efficiency through the use of the latest pseudo-gradient as sharpness approximation.

\noindent Several approaches address client drift by adding regularization during local training. FedProx \citep{li2020federated_fedprox}  introduces a term to keep local parameters close to the global model, \feddyn \citep{acar2021federated} employs ADMM to align local and global convergence points,  \adabest \citep{varno2022adabest} adjusts local updates with an adaptive bias estimate, and \scaffold \citep{karimireddy2020scaffold} applies stochastic variance reduction. 
Momentum-based techniques \citep{sutskever2013importance} are also employed to maintain a consistent global trajectory, either on the server side (\eg, \fedavgm \citep{hsu2019measuring}) or by incorporating global information into local training \citep{karimireddy2020mime,kim2022communication,gao2022feddc,zaccone2023communication}. 
Unlike \feddyn, where ADMM can lead to parameter explosion \citep{varno2022adabest}, our \ours successfully leverages {ADMM to align global and local solutions}, even under extreme heterogeneity, aided by \sam on server. %

\noindent \textbf{Centralized SAM.} \xspace %
To avoid doubling client-server exchanges caused by \sam's two-step process, \ours draws on insights from the literature on \sam in centralized settings. 
Several strategies have been proposed to minimize computational overhead, including reducing the number of parameters needed to compute the sharpness-aware components \citep{du2021efficient}, 
or approximating them \citep{liu2022towards,du2022sharpness,park2023differentially}. %
DP-SAT \citep{park2023differentially} approximates the ascent step with the gradient from the previous iteration, and SAF \citep{du2022sharpness} replaces \sam's sharpness approximation with the trajectory of weights learned during training. Aiming to the same goal, {\ours approximates the sharpness measure with the pseudo-gradient from the previous round on the server side}, without incurring in unnecessary exchanges with the clients and effectively guiding the optimization toward globally flat minima. %
\vspace{-5pt}

\vspace{-10pt}
\section{Background}
\label{sec:background}
\vspace{-5pt}
This section introduces the FL problem setting and preliminary notations on \sam \citep{foret2020sharpness} and \fedsam \citep{caldarola2022improving,qu2022generalized}.

\vspace{-5pt}
\subsection{Problem setting}
In FL, a central server communicates with a set of clients $\C$ for $T$ rounds. The goal is to learn a global model $f(\w): \X \rightarrow \Y$ parametrized by $\w\in\mathbb{R}^d$, where $\X$ and $\Y$ are the input and the output spaces respectively. In image classification, $\X$ contains the images and $\Y$ their corresponding labels. Each client $k\in\C$ has access to a local dataset $\D_k$ of $N_k$ pairs $\{(x_i, y_i), x_i\in\X,  y_i\in\Y\}_{i=1}^{N_k}$. In realistic heterogeneous settings, clients usually hold different data distributions and quantity, \ie, $\mathit{D}_i \neq \mathit{D}_j$ and $N_i \neq N_j \, \forall i \neq j \in\C$. The global FL objective is:
\vspace{-5pt}
\begin{equation}
    \min_{\w} \left\{f(\w) = \frac{1}{C} \sum_{k\in\C} f_k(\w)\right\}, f_k(\w) \triangleq \mathbb{E}%
    f_k(\w, \xi_k),
    \label{math:fl_goal}
    \vspace{-5pt}
\end{equation}
where $C \triangleq |\C|$ is the total number of clients, $f_k$ is the empirical loss on the $k$-th client (\eg, cross-entropy loss) and $\xi_k$ is the data sample randomly drawn from the local data distribution $\mathit{D}_k$. The training process is a two-phase optimization approach within each round $t\in[T]$. First, due to potential client unavailability, a subset of selected clients $\C^t\subset\C$  trains the received global model using their local optimizer \textsc{ClientOpt} (\eg, \sgd, \sam). Then, the server aggregates their updates with a server optimizer, \textsc{ServerOpt}. %
\fedopt \citep{reddi2020adaptive} solves \cref{math:fl_goal} as
\vspace{-5pt}
\begin{align}
    \Delta^t_{\w} &\triangleq \sum_{k\in\C^t} \frac{N_k}{N} (\w^t - \w^{t}_k) \, \text{ and }\\
    \w^{t+1} &\leftarrow \w^t - \textsc{ServerOpt}(\w^t, \Delta^t_{\w}, \eta_s),
    \label{math:fedavg}
    \vspace{-5pt}
\end{align}
where $\Delta^t_{\w}$ is the \textbf{global pseudo-gradient} at round $t$, $N = \sum_{k\in\C^t} N_k$ the total number of images seen during the current round, $\eta_s$ the server learning rate, $\w^t$ the global model and $\w^t_k$ the local update resulting from training on client $k$'s data with \textsc{ClientOpt} for $E$ epochs. \fedavg \citep{mcmahan2017communication} computes $\w^{t+1}$ as $\sum_{k\in\C^t} \nicefrac{N_k}{N}\w^{t}_k$, corresponding to one SGD step on the pseudo-gradient $\Delta^t_{\w}$ with $\eta_s=1$ \citep{reddi2020adaptive}. %

\subsection{Sharpness-aware Minimization}
\sam \citep{foret2020sharpness} jointly minimizes the loss value and the sharpness of the loss landscape by solving the  min-max problem
\begin{equation}
    \min_{\w} \left\{F(\w) \triangleq \max_{\|\beps\|\leq \rho} f(\w + \beps)\right\},
    \label{math:sam}
\end{equation}
where $\beps$ is the perturbation to estimate the sharpness, $f$ the loss function, $\rho$ the neighborhood size and $\|\cdot\|$ the $\ell_2$ norm. Using the first-order Taylor expansion of $f$, \sam efficiently solves the inner maximization as 
\vspace{-5pt}
\begin{equation}
     \argmax_{\|\beps\|\leq \rho} f(\w) + \beps^\top \nabla_{\w} f(\w) = \rho \frac{\nabla_{\w} f(\w)}{\|\nabla_{\w} f(\w)\|} \triangleq \heps(\w).
    \label{math:eps}
    \vspace{-5pt}
\end{equation}
$\heps$ is the scaled gradient of the loss \wrt the current parameters $\w$. %
The \emph{sharpness-aware gradient} is $\nabla_{\w} f(\w)|_{\w+\heps(\w)}$. \cref{math:sam} is solved with a first gradient ascent step to compute $\heps$ and a descent step with the sharpness-aware gradient, updating the model as $\w \leftarrow \w - \eta \nabla_{\w} f(\w)|_{\w+\heps(\w)}$. 

\subsection{SAM in Federated Learning}
\fedsam \citep{caldarola2022improving,qu2022generalized} aims to improve the clients' models generalization through convergence to flatter regions by using \sam in the local training. From \cref{math:fl_goal,math:sam}, the global objective becomes $\min_{\w} \left\{f^{\sam}(\w) = \nicefrac{1}{C} \sum_{k\in\C} f_k^{\sam}(\w)\right\}$, with $f_k^{\sam}(\w) \triangleq \max_{\|\beps_k\|\leq \rho} f_k(\w + \beps_k)$  
with local perturbation $\beps_k$. The intuition behind this approach is that the improved local models' generalization positively reflects on the global model performance. However, by independently applying \cref{math:sam} in the local optimization, \fedsam does not explicitly address global flatness, potentially leading to discrepancies between local and global loss geometries.  %
\vspace{-8pt}

\section{Local-Global Sharpness Inconsistency}
\label{sec:inconsistency}
\vspace{-5pt}
This section empirically investigates the hypothesis that discrepancies between local and global loss landscapes impact \fedsam's performance, using a CNN model on \cifarten and \cifar datasets — further details in \cref{sec:exp}. 

\cref{fig:landscapes_heterogeneity} compares the loss surfaces of CNNs trained with \fedavg and \fedsam. 
On the easier \cifarten, \fedsam exhibits flatter minima \wrt \fedavg, %
effectively navigating simpler landscapes. However, their  difference diminishes with increasing dataset complexity (\cifar) and heterogeneity ($\alpha\rightarrow0$). This suggests {larger discrepancies between local and global geometries arise as tasks become more complex and data distributions more diverse}. %

\begin{figure}[t]
    \vspace{-17pt}
    \centering
    \captionsetup{font=scriptsize}
    \subfloat[Trained on class \texttt{sea}][Trained on class \texttt{sea}\label{fig:local_global_full:c0_sam}]{\includegraphics[width=.3\linewidth]{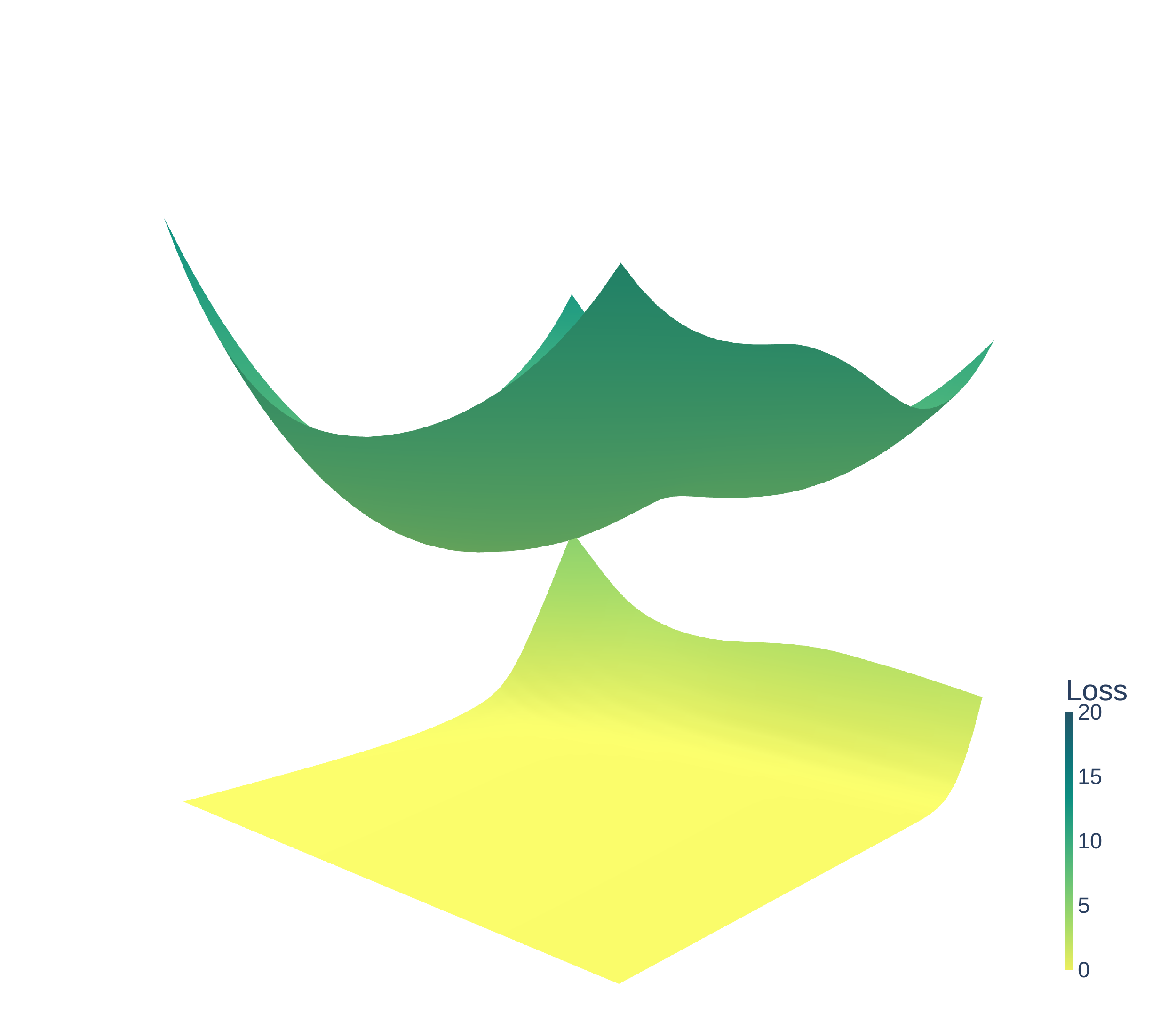}}
    \hfill
    \subfloat[Trained on class \texttt{snail}][Trained on class \texttt{snail}\label{fig:local_global_full:c1_sam}]{\includegraphics[width=.3\linewidth]{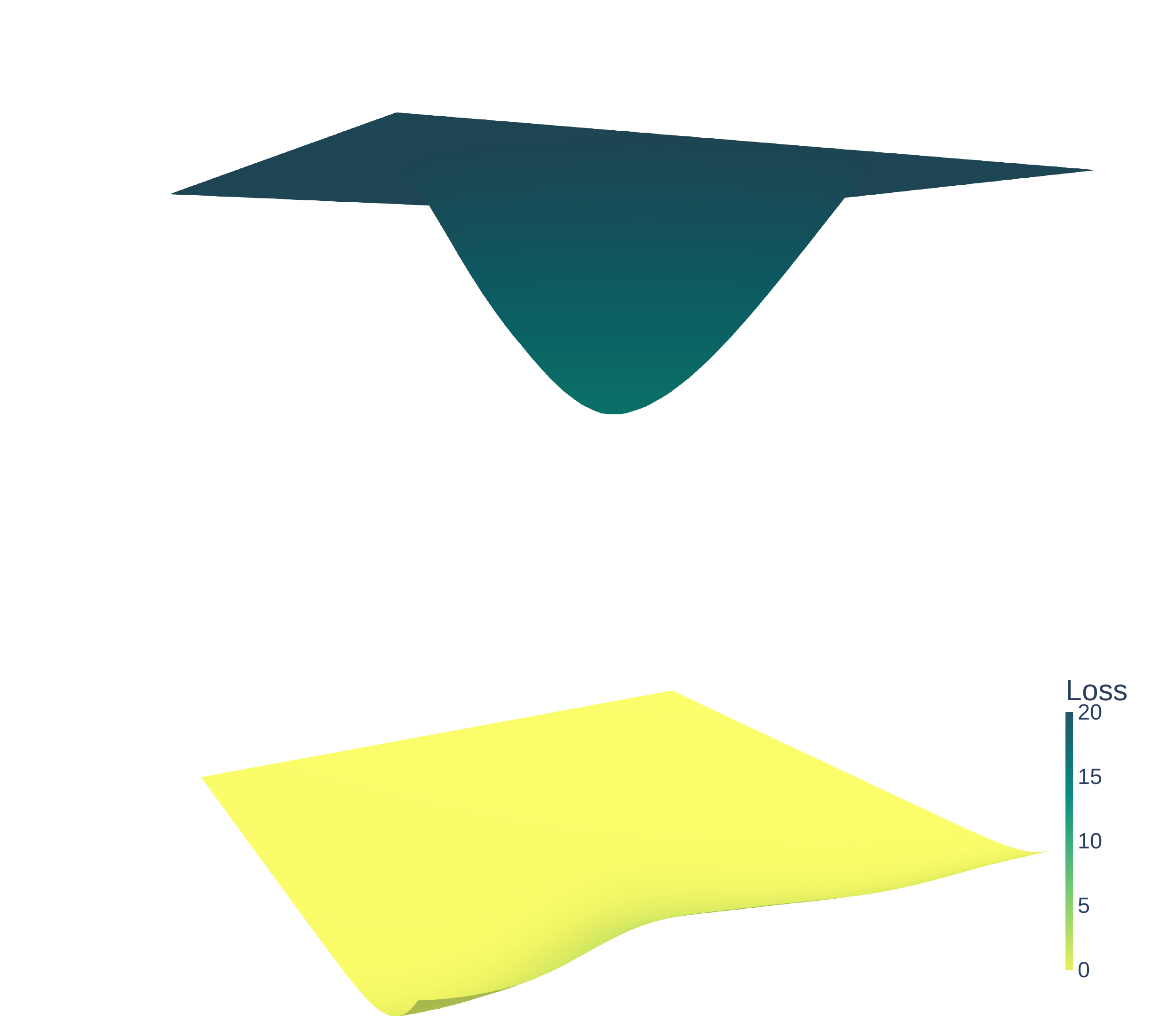}}
    \hfill
    \subfloat[Trained on class \texttt{skyscraper}][Trained on class \texttt{skyscraper}\label{fig:local_global_full:c3_sam}]{\includegraphics[width=.3\linewidth]{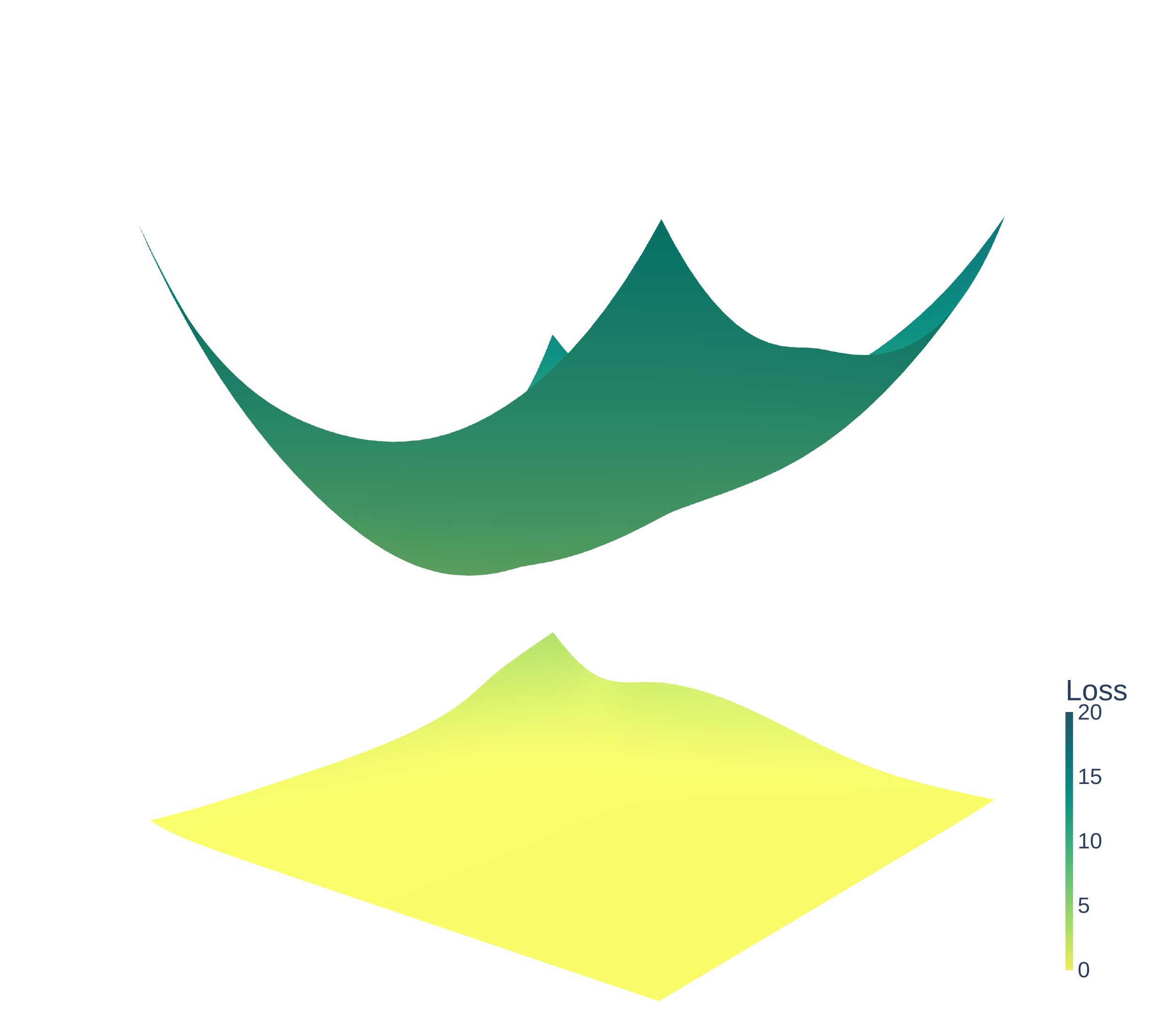}}
    \vspace{-5pt}
    \caption{\textbf{Global \vs local perspective on \fedsam}. \cifar $\alpha=0$ @ $20k$ rounds on CNN. Local models trained on one \texttt{class}, tested on the local (\textit{bottom landscape}) or global dataset (\textit{top landscape}). \textbf{Models trained with \fedsam present significant differences between local and global behaviors.}} %
    \label{fig:global_local_full}
    \vspace{-8pt}
\end{figure}

To highlight the existing difference between local and global behavior, \cref{fig:global_local_full} investigates the behavior of client models at the end of local training when tested on their own data $\D_k$ (\textit{bottom} landscape), prior to server-side aggregation, \wrt the overall dataset $\D$ (\textit{top} landscape). %
Each plot shows the behavior of one of the randomly selected clients during the last round with \fedsam, distinguished by the locally seen class (results for all 5 clients in \cref{app:benefits}). %
The inconsistency between local and global behavior can be easily appreciated: locally, each model lands in a flat region; differently, the same model is close to saddle points (\cref{fig:local_global_full:c0_sam}) %
or sharp minima (\cref{fig:local_global_full:c1_sam,fig:local_global_full:c3_sam}) in the global landscape. %
These findings are further corroborated by the maximum Hessian eigenvalues presented in \cref{tab:eigs_fedgloss}, computed using each client's local dataset ($\lambda_{1,l}$) and the overall one ($\lambda_{1,g}$). \fedsam performs well on simple datasets like \cifarten with {global eigenvalues smaller than local ones}. However, $\lambda_{1,l} \ll \lambda_{1,g}$  on the more complex \cifar. 
This suggests that \fedsam effectively achieves \textit{local} convergence to flatter regions of the loss landscape on individual devices, but the higher global eigenvalue indicates limitations in reaching a \textit{globally} flat minimum. The challenge of achieving flat regions under high heterogeneity and the gap between local and global flatness support the introduction of \ours. 

\begin{table}[t]
    \centering
    \setlength{\tabcolsep}{1.5pt}
    \captionsetup{font=scriptsize}
    \caption{\textbf{Maximum Hessian eigenvalues of local models}, computed on global ($\lambda_{1,g}$) and local datasets ($\lambda_{1,l}$). \cifarten and \cifar, $\alpha=0$. Each client is identified via its local class. The lowest $\lambda_{1,g}$ in \textbf{bold}. \feddyn does not converge on \cifar with $\alpha=0$ \citep{caldarola2022improving,varno2022adabest}, hence the lack of results (\textcolor{red}{\xmark}).}
    \vspace{-5pt}
    \scriptsize
    \resizebox{\linewidth}{!}{
    \begin{tabular}{llrrrrrrrrrrrrrrrrr}
        \toprule
            & \textbf{Local} & \multicolumn{2}{c}{\textbf{\fedavg}} && \multicolumn{2}{c}{\textbf{\fedsam}} && \multicolumn{2}{c}{\textbf{\feddyn}} && \multicolumn{2}{c}{\textbf{\feddyn + \sam}} && \multicolumn{2}{c}{\textbf{\fedsmoo}} && \multicolumn{2}{c}{\textbf{\ours}} \\
        \cline{3-4} \cline{6-7} \cline{9-10} \cline{12-13} \cline{15-16} \cline{18-19}
            & \textbf{Class} & $\lambda_{1,l}$ & $\lambda_{1,g}$ && $\lambda_{1,l}$ & $\lambda_{1,g}$ && $\lambda_{1,l}$ & $\lambda_{1,g}$ && $\lambda_{1, l}$ & $\lambda_{1, g}$ && $\lambda_{1,l}$ & $\lambda_{1,g}$ && $\lambda_{1,l}$ & $\lambda_{1,g}$ \\
        \midrule
            \multirow{5}{*}{\rotatebox[origin=c]{90}{\cifarten}} & \texttt{airplane} & 9.1 & 239.1 && 100.6 & 36.4 && 752.5 & 347.8 && 199.6 & 12.0 && 122.1 & 26.5 && 190.1 & \textbf{4.3} \\
            & \texttt{cat} & 424.2 & 273.6 && 28.8 & 16.5 && 59.9 & 242.3 && 122.0 & 11.1 && 82.4 & 26.9 && 106.9 & \textbf{3.9} \\
            & \texttt{bird} & 18.4 & 237.0 && 106.4 & 35.7 && 894.0 & 371.2 && 200.2 & 12.0 && 134.2 & 25.7 && 200.1 & \textbf{4.1} \\
            & \texttt{airplane} & 483.5 & 269.5 && 103.2 & 30.6 && 761.6 & 348.9 && 206.9 & 12.3 && 122.8 & 25.2 && 207.8 & \textbf{4.0} \\
            & \texttt{frog} & 263.2 & 259.6 && 68.1 & 32.9 && 528.9 & 286.0 && 155.6 & 11.7 && 79.3 & 33.5 && 84.8 & \textbf{4.1} \\
        \midrule
            \multirow{5}{*}{\rotatebox[origin=c]{90}{\cifar}} & \texttt{sea} & 251.0 & 224.5 && 0.1 & 238.5 && \multirow{5}{*}{\textcolor{red}{\xmark}} & \multirow{5}{*}{\textcolor{red}{\xmark}} && \multirow{5}{*}{\textcolor{red}{\xmark}} & \multirow{5}{*}{\textcolor{red}{\xmark}} && 33.2 & 31.4 && 28.3 & \textbf{19.6} \\
            & \texttt{snail} & 91.2 & 267.0 && 0.2 & 149.1 &&&&&&&& 331.2 & 102.2 && 260.8 & \textbf{40.7} \\
            & \texttt{bear} & 108.4 & 215.2 && 6.7 & 129.3 &&&&&&&& 428.6 & 121.0 && 220.3 & \textbf{49.6} \\
            & \texttt{skyscraper} & 613.3 & 300.1 && 1.3 & 194.6 &&&&&&&& 143.5 & 40.2 && 269.2 & \textbf{22.2}  \\
            & \texttt{possum} & 37.9 & 259.6 && 15.3 & 142.6 &&&&&&&& 455.5 & 90.9 && 392.4 & \textbf{39.0} \\
        \bottomrule
    \end{tabular}}
    \label{tab:eigs_fedgloss}
    \vspace{-15pt}
\end{table}

\vspace{-10pt}
\section{FL with Global Server-side Sharpness}
\label{sec:method}
\vspace{-5pt}
\ours (\ourslong) overcomes \fedsam's limitations by efficiently optimizing both global flatness and consistency. %
\vspace{-5pt}

\subsection{Rethinking SAM in Federated Learning}
\vspace{-2pt}
Aiming to optimize \sam's objective (\cref{math:sam}) on the global function, \ours solves $\min_{\w} \left\{ \F(\w) = \frac{1}{C} \sum_{k\in\C} \F_k(\w) \right\}$, with $\F_k(\w) \triangleq \max_{\|\beps\|\leq\rho} f_k(\w+\beps)$,  %
where $\beps$ is the global perturbation. Calculating the true $\beps$ value requires the global gradient $\nabla_{\w}f$ (\cref{math:eps}) computed on the entire dataset $\D \triangleq \cup_{k\in\C} \D_k$, which is not available in FL due to data privacy and communication constraints. %
While \fedsmoo \citep{sun2023dynamic} tackles this issue by using ADMM on the sharpness with the constraint $\beps = \beps_k$, it necessitates transmitting $\beps$ alongside the model parameters $\w$ to both clients and server in each round, hindering its practicality in real-world scenarios with limited communication budgets. This observation motivates the question: \textit{how to minimize global sharpness while maintaining communication efficiency}? 
\subsubsection{Challenges of Server-side SAM} 
We address this question by applying \sam on the server side, directly optimizing for global sharpness and eliminating the need to align local sharpness on the clients. 
The global model has to be updated as $\w^{t+1} \leftarrow \w^t - \eta_s \nabla_{\w} \F(\w)|_{\w^t + \heps^t (\w)}$, where $\heps^t$ is the global perturbation at each round $t$. However, a key challenge arises: the computation of both $\heps^t$ and the sharpness-aware gradient necessitates two transmissions with the clients, 
making its direct application in server-side FL non-trivial. A straightforward solution is to emulate SAM's double computation step through two communication exchanges $\forall t\in[T]$. 
\begin{itemize}
    \item {\textbf{Step 1}}: %
    the server sends the global model $\w^t$ to a subset $\C^t$ of clients, which update it using their local data. With the resulting pseudo-gradient %
    $\Delta^t_{\w}$, $\heps^t(\w) = \rho (\nicefrac{\Delta^t_{\w}}{\|\Delta^t_{\w}\|})$ and the perturbed model $\tw^{t} = \w^t + \heps^t(\w)$. 
    \item {\textbf{Step 2}}: the server transmits $\tw^{t}$ to the \textit{same} $\C^t$, which %
    compute their update $\tw_k^t \, \forall k$. The resulting global pseudo-gradient $\Tilde{\Delta}^t_{\w} \triangleq \sum_{k\in\C^t} \nicefrac{N_k}{N}(\tw^t - \tw^t_k)$ is an estimate of $\nabla_{\w}\F(\w)|_{\w^t + \heps^t (\w)}$. 
\end{itemize}
This two-step approach, referred to as \naiveours, is conceptually simple but suffers from communication \emph{in}efficiency, doubling the communication cost \wrt FedAvg,  %
while requiring the same set of clients $\C^t$ to remain active for two consecutive exchanges. This may be unrealistic in real-world settings often characterized by network failures. These limitations highlight the need for an efficient alternative that accounts for practical real-world FL factors.

\setlength\tabcolsep{3pt}
\begin{table}[t]
\vspace{-17pt}
    \centering
    \captionsetup{font=scriptsize}
    \caption{Overview of FL methods using \sam. Differently from previous works, \ours uses \sam as server optimizer and allows any local optimizer.}
    \vspace{-8pt}
    \scriptsize
    \resizebox{\linewidth}{!}{
    \begin{tabular}{lccccc}
    \toprule
         \multirow{2}{*}{\textbf{Method}} & \multirow{2}{*}{\textbf{\textsc{ServerOpt}}}  & \multirow{2}{*}{\textbf{\textsc{ClientOpt}}} & \textbf{Global} & \textbf{Communication} & \textbf{Local Computation}\\
         &&& \textbf{Flatness} & \textbf{Cost} & \textbf{Cost}\\
         \midrule
         \fedsam \cite{caldarola2022improving,qu2022generalized} & \sgd & \sam & \xmark & \textcolor{ForestGreen}{\boldmath$1\times$} & \textcolor{BrickRed}{\boldmath$2\times$}\\
         \feddyn \cite{acar2021federated} + \sam & \sgd & \sam & \xmark & \textcolor{ForestGreen}{\boldmath$1\times$} & \textcolor{BrickRed}{\boldmath$2\times$}\\
         \fedspeed \cite{sun2023fedspeed} & \sgd & Similar to \sam & \xmark &\textcolor{ForestGreen}{\boldmath$1\times$} & \textcolor{BrickRed}{\boldmath$2\times$}\\
         \fedgamma \cite{dai2023fedgamma}& \sgd & \sam & \textcolor{ForestGreen}{\cmark} & \textcolor{BrickRed}{\boldmath$2\times$} & \textcolor{BrickRed}{\boldmath$2\times$}\\
         \fedsmoo \cite{sun2023dynamic} & \sgd & \sam & \textcolor{ForestGreen}{\cmark} & \textcolor{BrickRed}{\boldmath$2\times$} & \textcolor{BrickRed}{\boldmath$2\times$}\\
         \textbf{\ours} & \textbf{\sam} & \textbf{Any optimizer} & \textcolor{ForestGreen}{\cmark} & \textcolor{ForestGreen}{\boldmath$1\times$} & \textcolor{ForestGreen}{\boldmath$1\times$} or \textcolor{BrickRed}{\boldmath$2\times$} \\
        \bottomrule
    \end{tabular}}
    \label{tab:algs}
    \vspace{-15pt}
\end{table}

\subsection{FedGloSS}
\label{subsec:method:fedgloss}

\begin{figure}[t]
  \centering
  \vspace{-20pt}
  \captionsetup{font=scriptsize}
  \includegraphics[width=\linewidth]{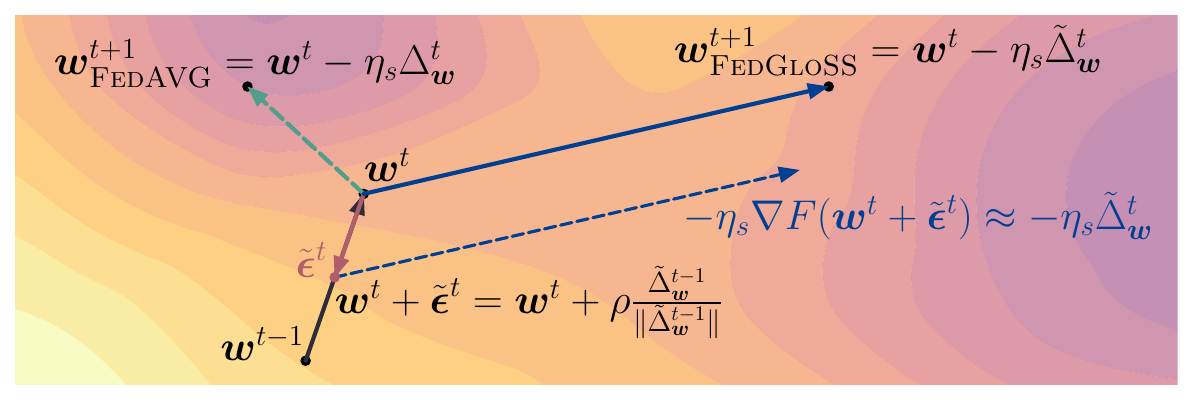}
  \vspace{-18pt}
  \caption{Illustration of \ours. The model $\w^t$ is \textcolor{DarkOrchid}{perturbed} using $\Tilde{\Delta}_{\w}^{t-1}$. The \textcolor{RoyalBlue}{sharpness-aware direction} (\textit{dashed}) is used to compute $\w^{t+1}$ (\textit{solid}), which lands in a flat region. Compared to \textcolor{Aquamarine}{\fedavg}.%
  }%
  \vspace{-15pt}
  \label{fig:method}
\end{figure}

To overcome the challenges posed by \naiveours, following \citep{park2023differentially}, \ours estimates $\heps^t$ using the {perturbed global pseudo-gradient from the previous round} $\Tilde{\Delta}^{t-1}_{\w}$ at each round $t$. This approach leverages available information \textit{without incurring extra communications} and avoids unnecessary computations. Intuitively, %
the use of the previous pseudo-gradient to minimize the sharpness allows \ours to access information on the \textit{global} loss landscape geometry, thus {guiding the \textit{global} optimization towards flatter minima}. From \cref{math:eps,math:fedavg}, \ours updates the global model $\w^t$ as 
\vspace{-8pt}
\begin{align*}
\vspace{-5pt}
    \textcircled{{\footnotesize 1}}&\, \textcolor{DarkOrchid}{\teps^t(\w)} \triangleq \rho \frac{\Tilde{\Delta}^{t-1}_{\w}}{\|\Tilde{\Delta}^{t-1}_{\w}\|} \quad \textcircled{{\footnotesize 2}}\, \tw^t \leftarrow \w^t \textcolor{DarkOrchid}{+ \teps^t(\w)}\\
    \textcircled{{\footnotesize 3}}&\, \text{Obtain $\tw^t_k$ from clients and } \Tilde{\Delta}^t_{\w} = \sum_{k\in\C^t} \frac{N_k}{N} (\tw^t - \tw^t_k) \\
    \textcircled{{\footnotesize 4}}& \w^{t+1} \leftarrow \w^t \textcolor{RoyalBlue}{- {\small \ours}(\w^t, \Tilde{\Delta}^t_{\w}, \eta_s)} = \w^t \textcolor{RoyalBlue}{- \eta_s \Tilde{\Delta}^t_{\w}},
    \vspace{-5pt}
\end{align*}
where with a slight abuse of notation \textsc{ServerOpt} from \cref{math:fedavg} is substituted with the server-side strategy proposed by \ours. The notation follows the colors of \cref{fig:method}, which depicts our approach. Notably, as summarized in \cref{tab:algs}, \ours enables \sam on the server side while {allowing any \textsc{ClientOpt} for local training}, with computational costs varying based on the chosen optimizer. This differs from previous methods constrained to the more computationally expensive \sam. In addition, differently from \fedsmoo, \ours maintains \fedavg's communication complexity while optimizing for global flatness.

\begin{figure*}[t]
    \centering
    \captionsetup{font=scriptsize}
    \captionsetup[sub]{font=scriptsize}
    \vspace{-25pt}
    \subfloat[Trained on class \texttt{sea}][Local model on class \texttt{sea}\\ w/ \textbf{\ours}\label{fig:consistency:c0_fedgloss}]{\includegraphics[width=.22\linewidth]{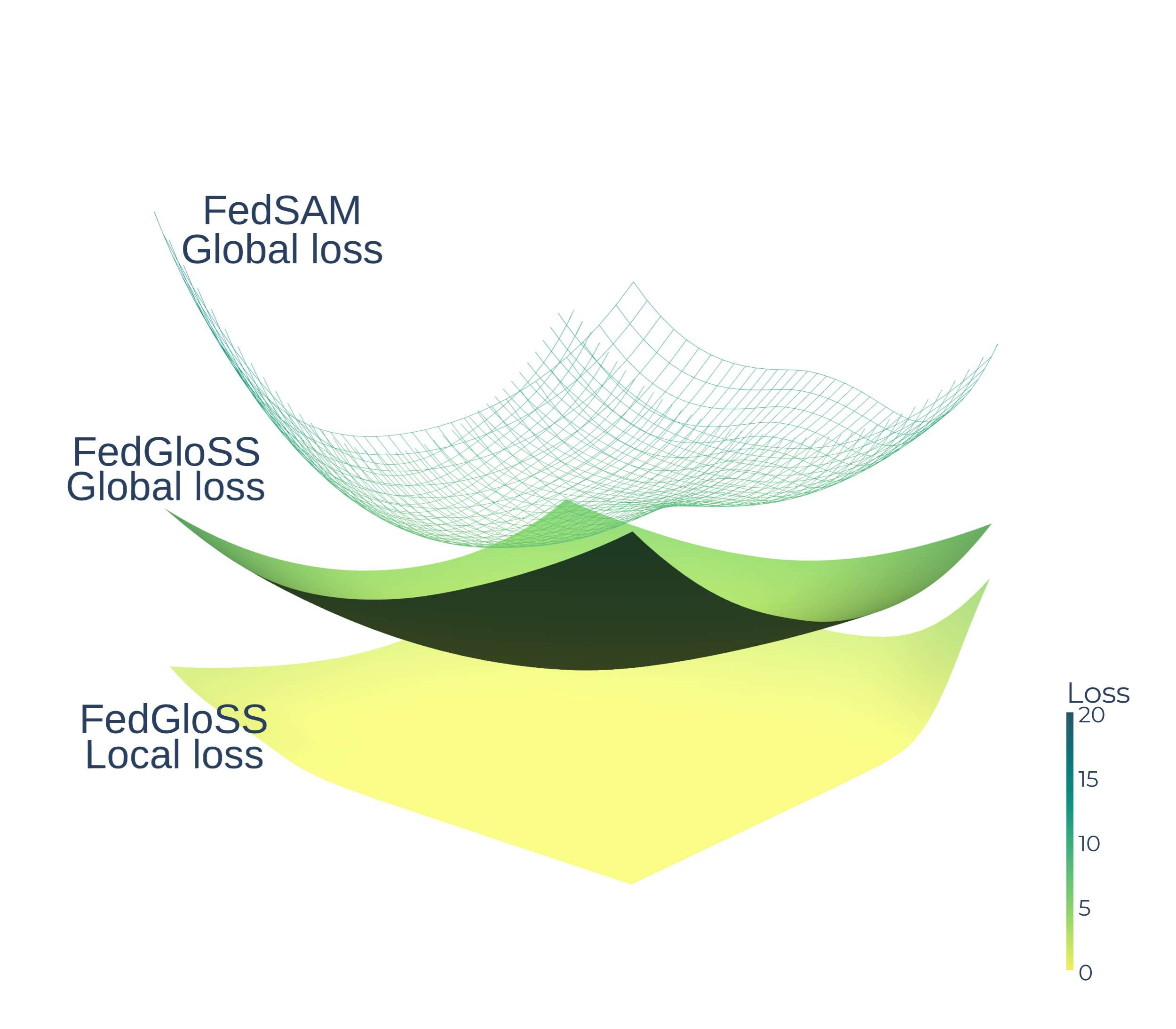}}
    \hfill
    \subfloat[Local model on class \texttt{snail}][Local model on class \texttt{snail}\\ w/ \textbf{\ours}\label{fig:consistency:c1_fedgloss}]{\includegraphics[width=.22\linewidth]{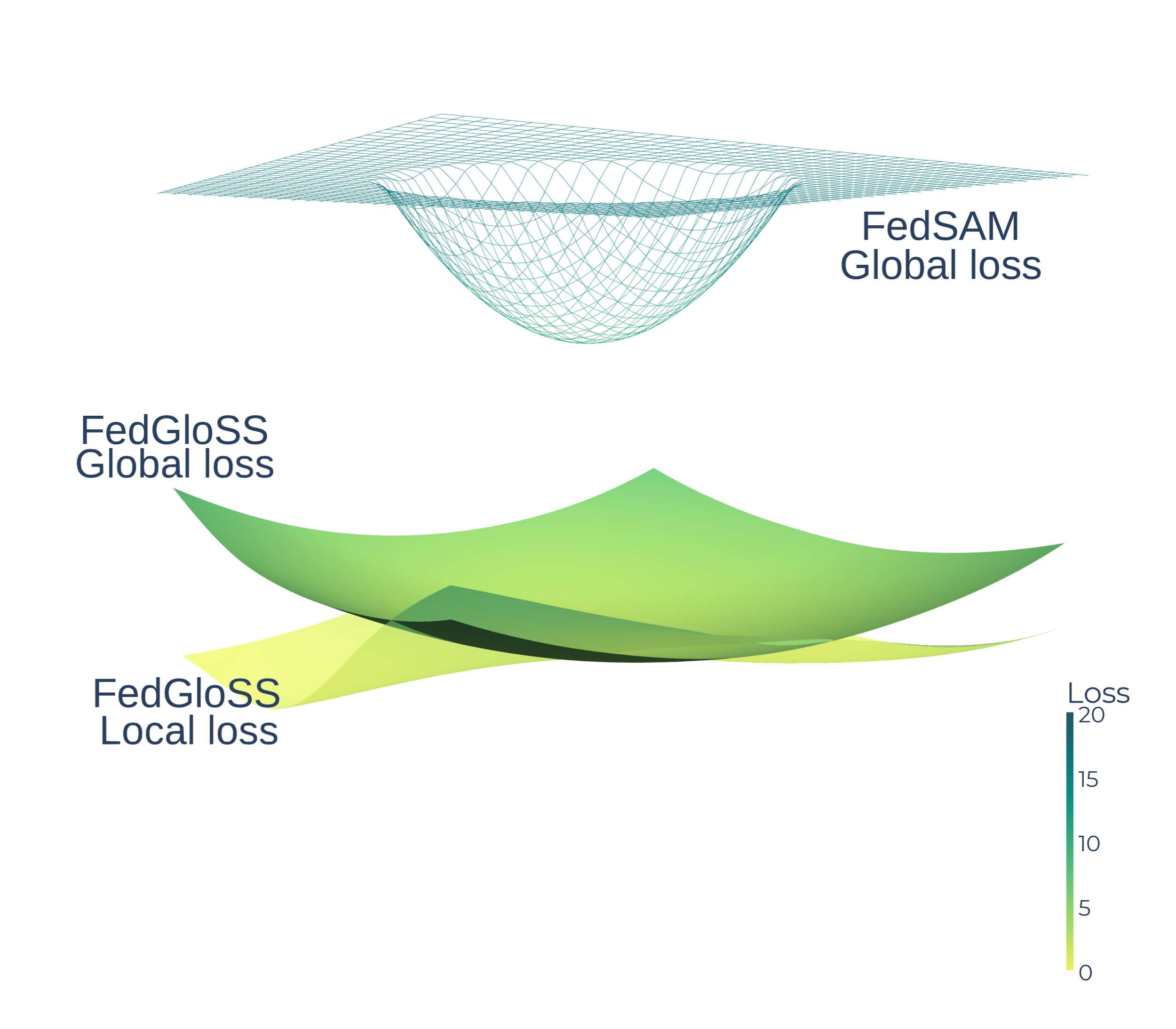}}
    \hfill
    \subfloat[Local model on class \texttt{sea}][Local model on class \texttt{sea}\\ w/ \textbf{\fedsmoo}\label{fig:consistency:c0_fedsmoo}]{\includegraphics[width=.22\linewidth]{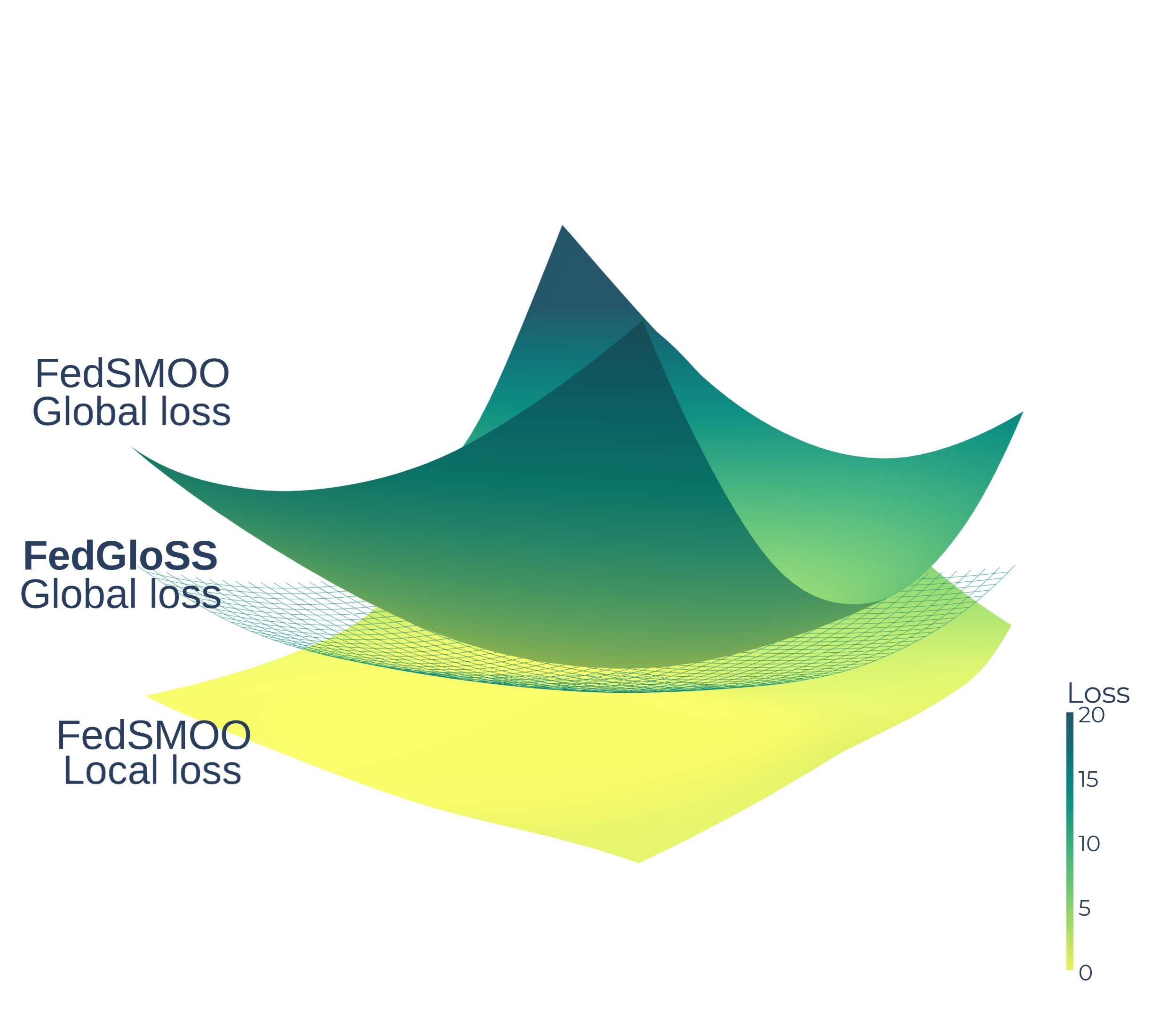}}
    \hfill
    \subfloat[Trained on class \texttt{snail}][Local model on class \texttt{snail}\\ w/ \textbf{\fedsmoo}\label{fig:consistency:c1_fedsmoo}]{\includegraphics[width=.22\linewidth]{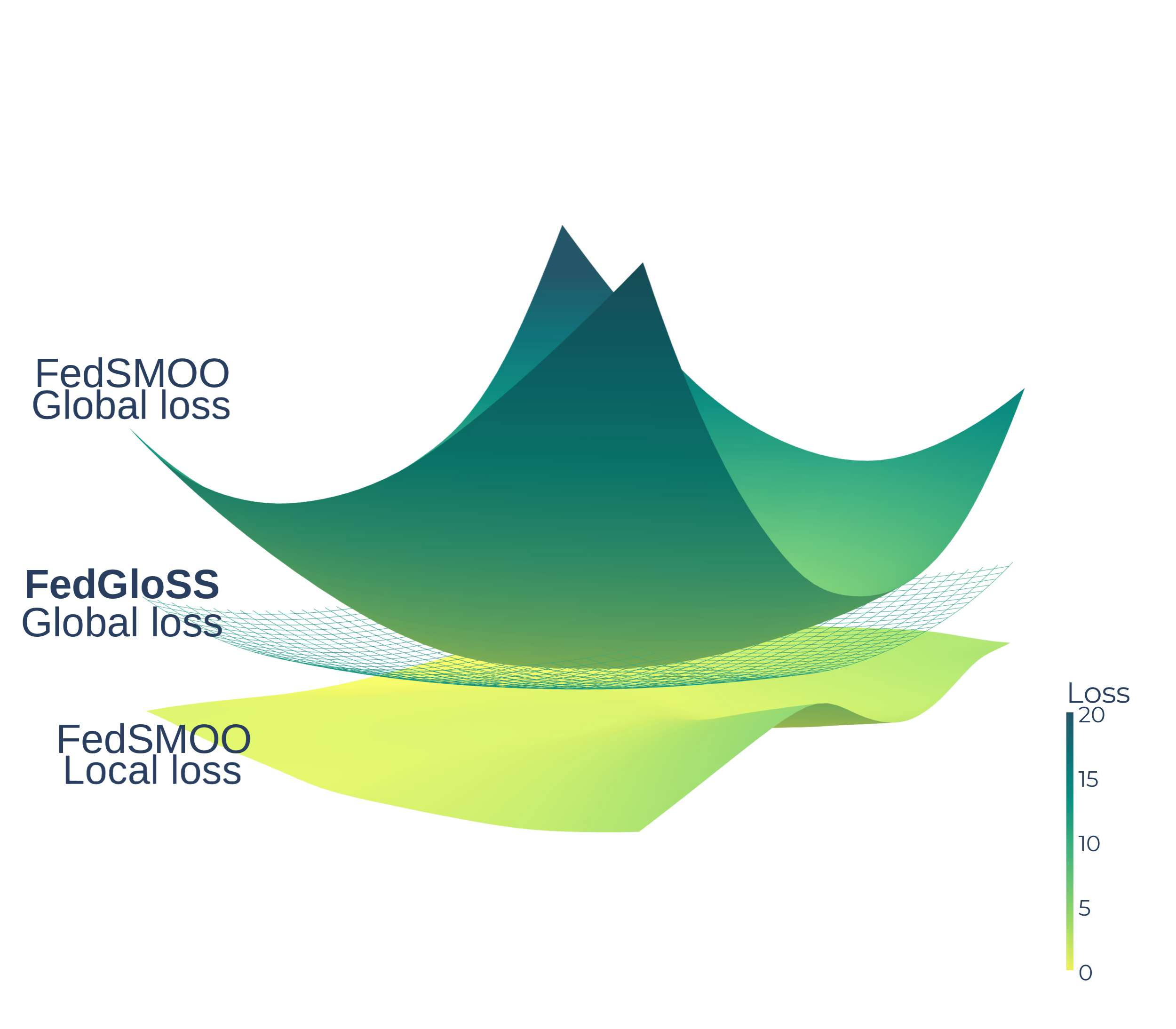}}
    \vspace{-8pt}
    \caption{\textbf{Global \vs local perspective of \ours and \fedsmoo.} Loss landscapes of clients models trained on one \texttt{class}, tested on the local (``\textit{Local loss}'') or global dataset (``\textit{Global loss}''). \cifar $\alpha=0$ with \sam as local optimizer @ $t=20k$, CNN. \textbf{(a)-(b):} Models trained with \ours. Global loss of \underline{\fedsam}'s local model  (\textit{net})  as reference. %
    \textbf{(c)-(d)}: Models trained with \fedsmoo. Global loss of \underline{\ours}' local model (\textit{net}) as reference. \textbf{\ours achieves better consistency \wrt \fedsmoo.}}%
    \label{fig:local_global_fedgloss}
    \vspace{-15pt}
\end{figure*}

\subsubsection{Promoting Global Consistency with ADMM}
The difference in using the approximation $\teps^t$ (\ours) and the true $\heps^t$ (\naiveours) is
\begin{equation}
\vspace{-5pt}
    \delta_{\epsilon}^t \triangleq \|\teps^t(\w) - \heps^t(\w)\| = \rho \left\|  \frac{\Tilde{\Delta}^{t-1}_{\w}}{\|\Tilde{\Delta}^{t-1}_{\w}\|} - \frac{{\Delta}^{t}_{\w}}{\|{\Delta}^{t}_{\w}\|} \right\|,
    \label{math:eps_difference}
\end{equation}
where $\Tilde{\Delta}^{t-1}_{\w}$ is computed using the updates of the clients in $\C^{t-1}$ and $\Tilde{\Delta}^{t}_{\w}$ with $\C^t$. 
\cref{math:eps_difference} suggests $\delta_{\epsilon}^t$ is minimized when $\Tilde{\Delta}^{t-1}_{\w}$ and $\Tilde{\Delta}^{t}_{\w}$ are aligned, which occurs when clients' updates are directionally consistent. %
However, in real-world heterogeneous FL, %
\textit{i}) to due clients' unavailability, only a subset of them participates in training at each round, with $\C^t$ likely differing from $\C^{t-1}$, and \textit{ii}) clients hold different data distributions, \ie, local optimization paths likely converge towards different local minima, leading to unstable global updates \citep{karimireddy2020scaffold}. 
As a consequence, $\delta_{\epsilon}^t \not\to 0$ necessarily. 

\begin{figure}[h]
    \centering
    \vspace{-12pt}
    \captionsetup{font=scriptsize}
    \includegraphics[width=0.9\linewidth]{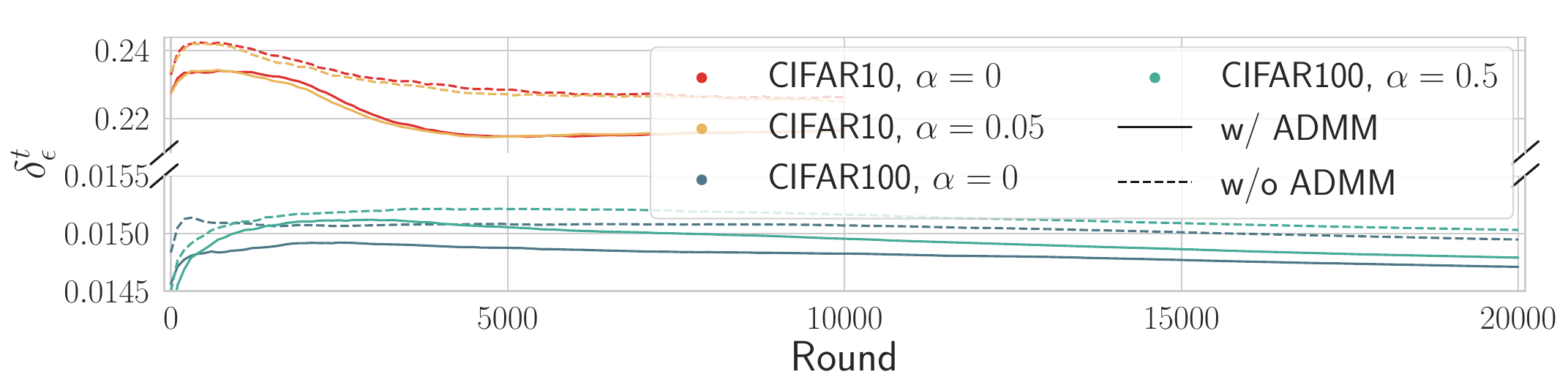}
    \vspace{-10pt}
    \caption{Trend of the difference $\delta_{\beps}^t$ (\cref{math:eps_difference}), which decreases as ADMM is used and over training rounds. \textsc{Cifar} datasets, CNN.}
    \label{fig:delta_eps}
    \vspace{-20pt}
\end{figure}

To align local and global objectives - guiding client and server updates in the same direction and minimizing  \cref{math:eps_difference} - \ours leverages the Alternating Direction Method of Multipliers (ADMM) \citep{boyd2011distributed} on $\w^t$ \citep{acar2021federated,sun2023dynamic,sun2023fedspeed}. While alternative approaches could be used, they either lack full immunity to data heterogeneity or have shown poor performance on realistic scenarios (\eg, variance reduction \cite{karimireddy2020scaffold,dai2023fedgamma}). In contrast, ADMM has been proved to converge under arbitrary heterogeneity \cite{acar2021federated} and can thus be leveraged as a base algorithm for \ours, as shown in  \cref{alg:fedgloss} in \cref{app:alg}. ADMM makes use of the augmented Lagrangian function $\mathcal{L}(\w, \W, \sigma) = \sum_{k\in\C} L(\w, \w_k, \sigma_k)$ where $\W=\{\w_1, \cdots, \w_C\}$ and $\sigma$ is the Lagrangian multiplier. 
The problem solved by $\mathcal{L}$ is
\vspace{-5pt}
{\small
\begin{equation}
    \frac{1}{C} \sum_{k\in\C} (f_k + \sigma_k^\top (\w^t - \w^t_k) + \frac{1}{2\beta} \|\w^t - \w_k^t\|^2) \textit{ s.t. } \w=\w_k
    \label{math:admm}
\end{equation}
}
\noindent with $\beta>0$ being an hyperparameter. \cref{math:admm}
is split into $C$ sub-problems of the form $\w_{k,E} = \argmin_{\w_k} \{f_k - \sigma_k^\top (\w^t-\w_k) + \frac{1}{2\beta} \|\w^t - \w_k^t\|^2 \} $. The local dual variable is updated as $\sigma_k \leftarrow \sigma_k - \frac{1}{\beta}(\w_{k, E}^t - \w_{k, 0}^t)$. %
The global one $\sigma$ is updated by adding the averaged $\w_k-\w^t \, \forall k \in\C$.  We note that ADMM introduces a key limitation of our approach, namely reliance on stateful clients. 
\cref{fig:delta_eps} confirms the directional consistency of local and global updates enforced by ADMM reduces $\delta_{\epsilon}^t$ (\cref{math:eps_difference}), \ie, the difference between the true and approximated perturbation. The gap between the directions of $\Tilde{\Delta}^{t-1}_{\w}$ and $\Tilde{\Delta}^{t}_{\w}$ remains constant after an initial phase, suggesting the most challenging loss landscape direction is largely stable over time. 

\section{Experiments}
\label{sec:exp}

\vspace{-3pt}
\subsection{Experimental Setting}
\label{subsec:exp:setting}
\vspace{-2pt}
\cref{app:exp_details} details implementation and hyperparameter settings. %

\noindent\textbf{Federated datasets.}\xspace We leverage established FL benchmarks \citep{caldas2018leaf,hsu2020federated,hsu2019measuring}. \underline{\textit{Small-scale image classification:}} following \citep{hsu2020federated,caldarola2022improving}, the federated versions of \cifarten (10 classes) and \cifar (100 classes) \citep{Krizhevsky09learningmultiple} split the respective $50k$ training images in 100 clients with 500 images each. The data distribution is controlled by Dirichlet's parameter $\alpha\in\{0, 0.05, 1, 5, 10\}$ for \cifarten and $\{0, 0.5\}$ for \cifar \citep{hsu2019measuring}. Lower $\alpha$ signifies increased heterogeneity, with $\alpha=0$ being the most challenging scenario (each client holds samples from one class).  \underline{\textit{Large-scale image classification:}} \gld ($2,028$ classes) \citep{hsu2020federated} is the federated  Google Landmarks v2 \citep{weyand2020google} with $164,172$ pictures of worldwide locations, split among $1,262$ realistic clients. %

\noindent\textbf{Models.} \xspace The effectiveness of \ours is shown using multiple model architectures. As in \citep{hsu2020federated,caldarola2022improving}, we use a Convolutional Neural Network (CNN) similar to LeNet5 \citep{lecun1998gradient} on both \cifarten ($T=10k$) and \cifar ($T=20k$). Experiments with ResNet18 \citep{he2015deep} run for $10k$ rounds. For \gld, we train MobileNetv2 \citep{sandler2018mobilenetv2,hsu2020federated} ($T=1.3k$), considering the limited resources at the edge. %

\noindent\textbf{Baselines.}\xspace To study real-world settings with varying participation, a small fraction of clients is sampled at each round, %
with participation rate set to $5\%$ with the CNN and $10\%$ with ResNet18  on both \textsc{Cifar}s, %
and to 50 clients per round in \gld ($\approx 4\%$).  %
\ours is compatible with any local optimizer (\cref{sec:method}). We choose SGD and SAM to comply with previous works and compare it with state-of-the-art (SOTA) methods for statistical heterogeneity in FL, distinguishing the results by optimizer type to highlight performance differences. %
\sgd-based approaches are FedAvg \citep{mcmahan2017communication}, FedProx \citep{li2020federated_fedprox}, %
FedDyn \citep{acar2021federated} and %
Scaffold \citep{karimireddy2020scaffold}, while use \sam %
 FedSAM \citep{caldarola2022improving,qu2022generalized}, FedDyn + SAM, FedSpeed \citep{sun2023fedspeed}, %
FedGamma \citep{dai2023fedgamma} %
and FedSmoo \citep{sun2023dynamic}. %

\vspace{-5pt}
\subsection{Achieving Local-Global Sharpness Consistency}
\label{subsec:landscape_consistency}
\begin{figure*}[th]
    \centering
    \vspace{-25pt}
    \captionsetup{font=scriptsize}
    \subfloat[][\textsc{C10} $\alpha=0$ \underline{CNN}\\ \ours \vs \fedsam \label{subfig:ours-fedsam-c10-a0}]{\includegraphics[width=.15\linewidth]{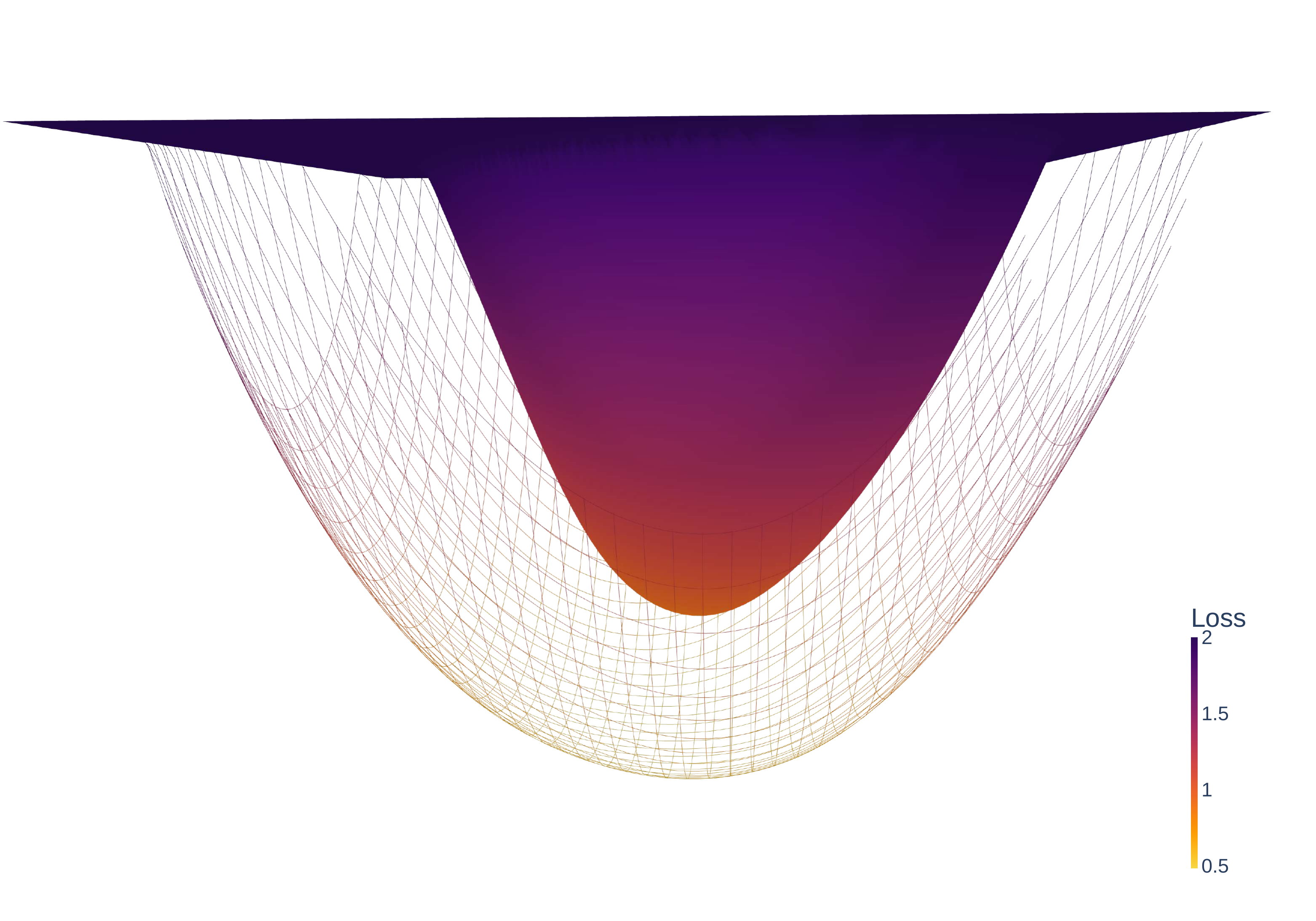}} \hfill
    \subfloat[][\textsc{C10} $\alpha=0$ \underline{CNN}\\ \ours \vs \fedsmoo  \label{subfig:ours-fedsmoo-c10-a0}]{\includegraphics[width=.15\linewidth]{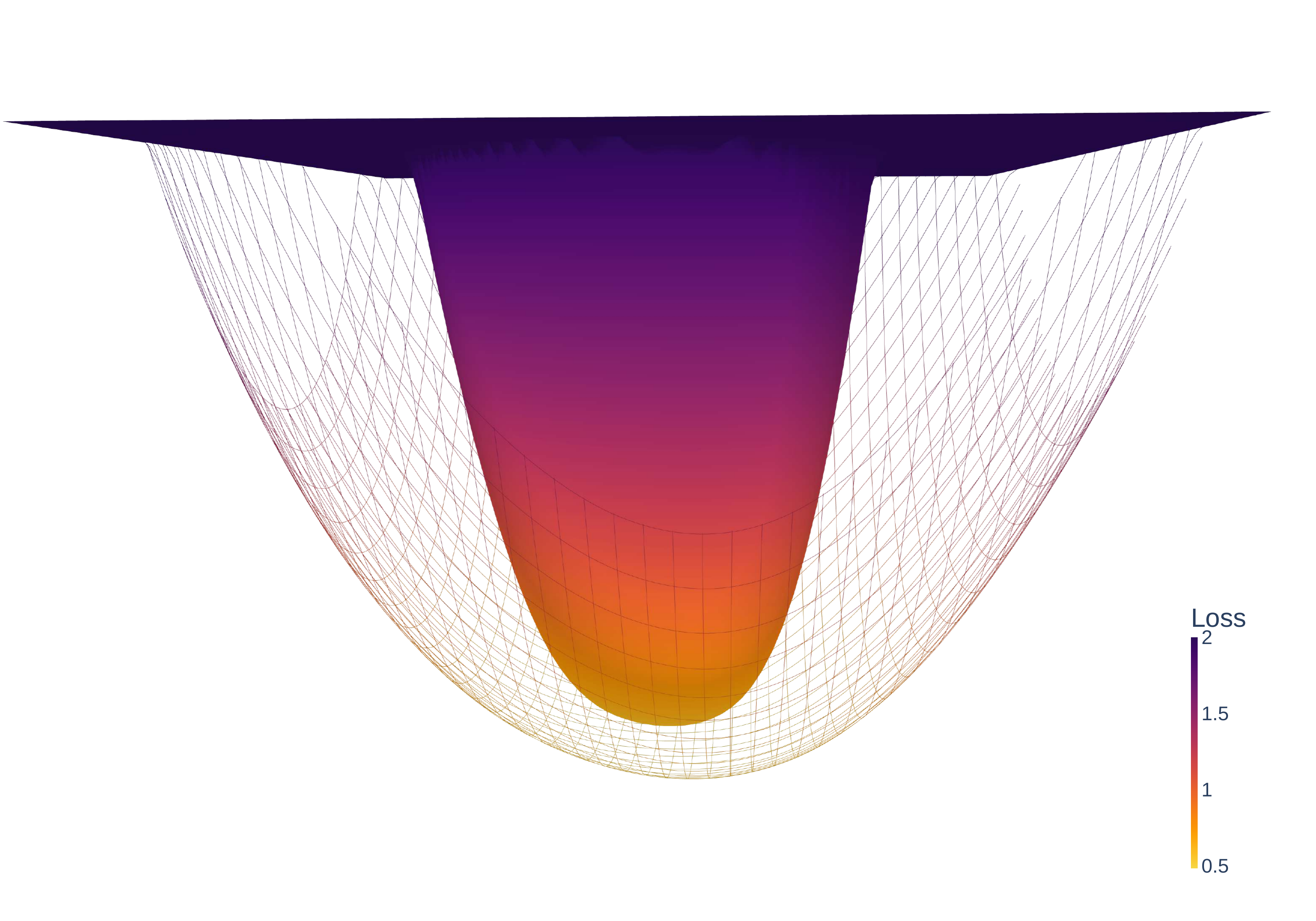}} \hfill
    \subfloat[][\textsc{C100} $\alpha=0$ \underline{CNN}\\ \ours \vs \fedsam  \label{subfig:ours-fedsam-c100-a0}]{\includegraphics[width=.15\linewidth]{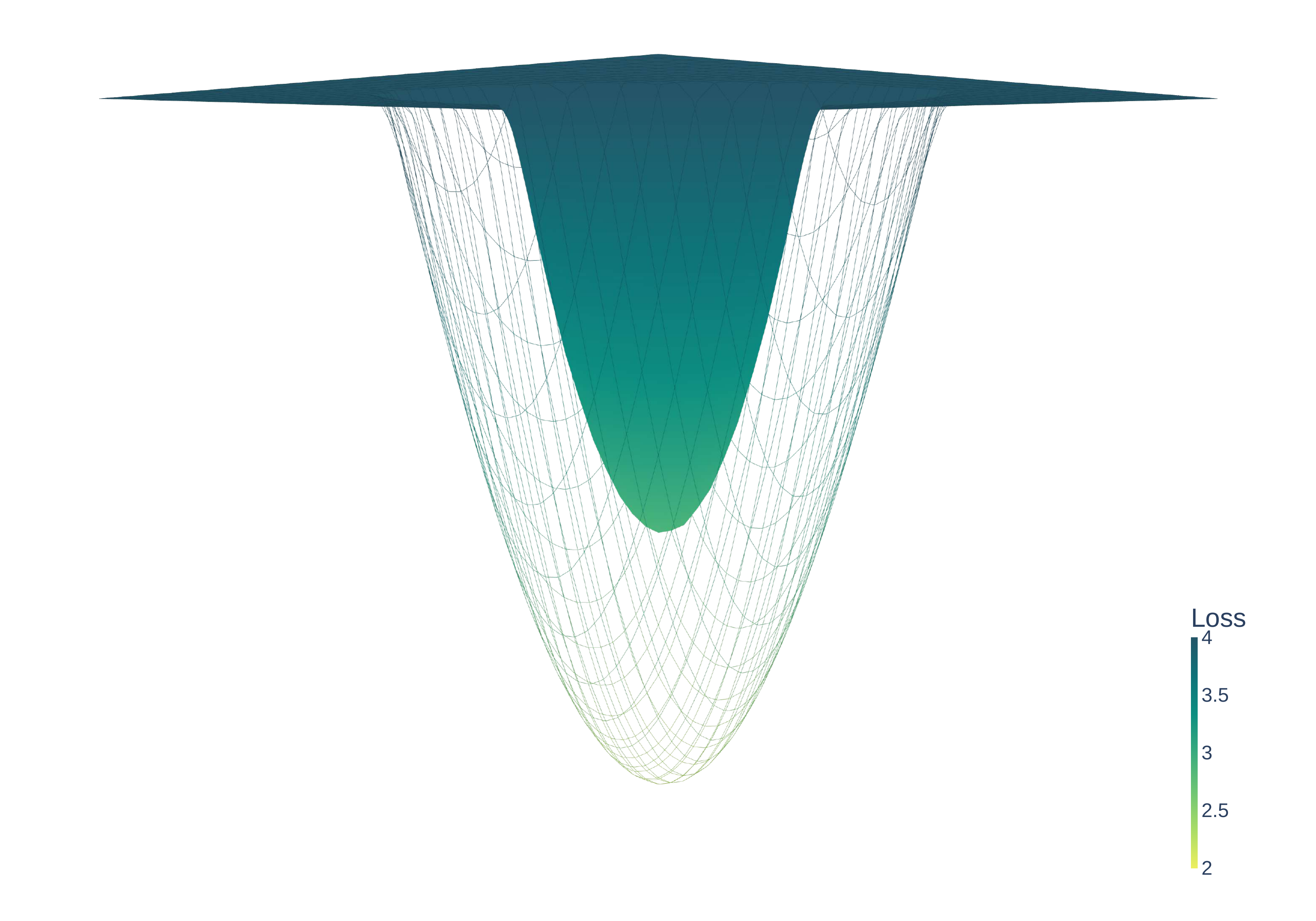}} \hfill
    \subfloat[][\textsc{C100} $\alpha=0$ \underline{CNN}\\ \ours \vs \fedsmoo \label{subfig:ours-fedsmoo-c100-a0}]{\includegraphics[width=.15\linewidth]{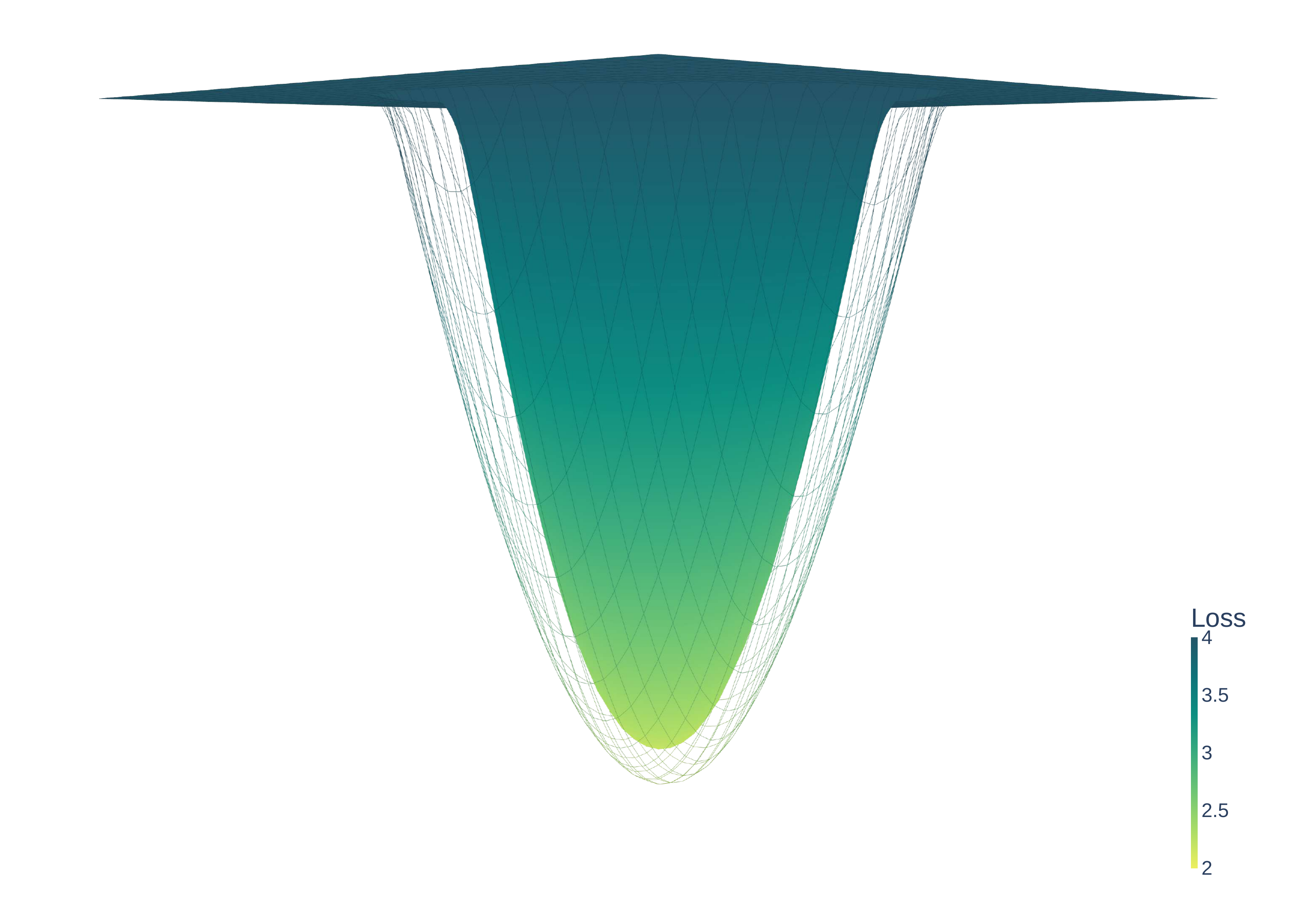}}
    \hfill
    \subfloat[][\textsc{C10} $\alpha=0.05$ \underline{ResNet18}\\ \ours \vs \fedsam  \label{subfig:ours-fedsam-c10-r18}]{\includegraphics[width=.15\linewidth]{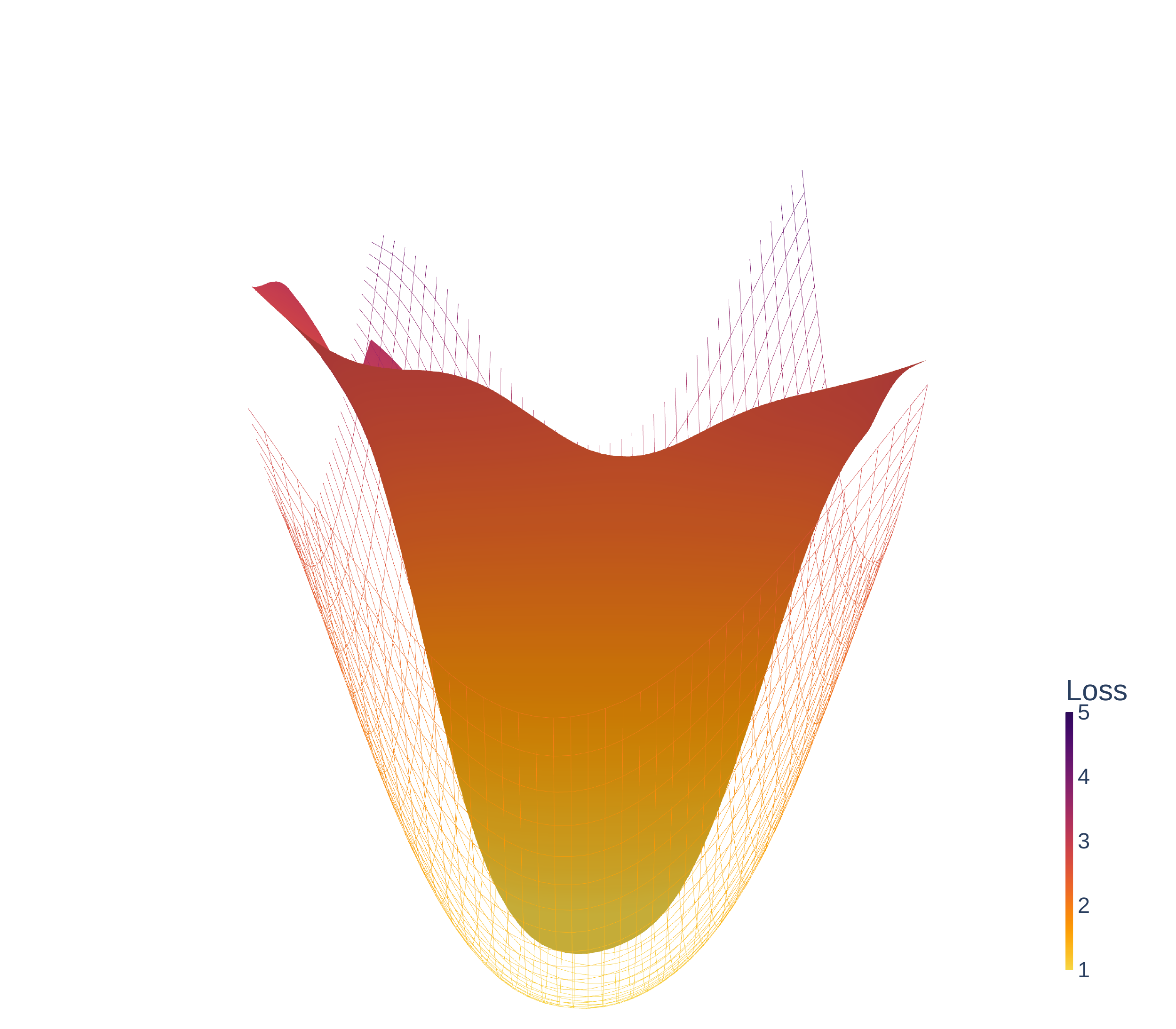}}
    \hfill
    \subfloat[][\textsc{C10} $\alpha=0.05$ \underline{ResNet18}\\ \ours \vs \fedsmoo  \label{subfig:ours-fedsmoo-c10-r18}]{\includegraphics[width=.15\linewidth]{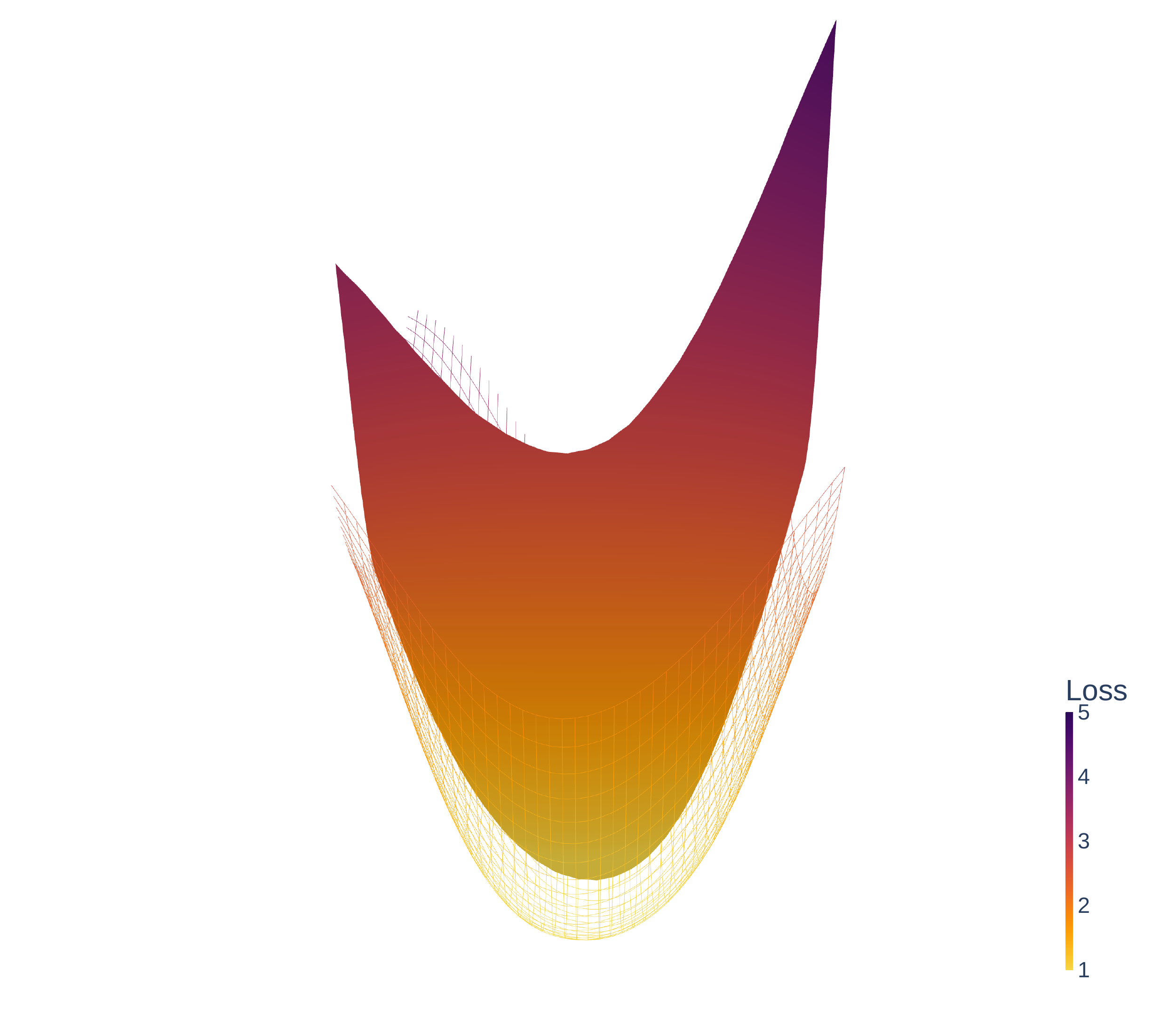}} %
    \vspace{-8pt}
    \caption{\textbf{Loss landscapes of models trained with \textbf{\ours} (\textit{net}) \vs \fedsam and \fedsmoo (\textit{solid})} on \cifarten/\textsc{100}. %
    \textbf{(a) - (c) - (e):} The flatter regions reached by \ours \wrt \fedsam prove the effectiveness of optimizing for global flatness. \textbf{(b) - (d) - (f):} \ours achieves flatter minima and lower loss values \wrt \fedsmoo. %
    }
    \label{fig:landscape_fedgloss_fedsam}
    \vspace{-10pt}
\end{figure*}

\vspace{-5pt}
To assess the effectiveness of \ours in promoting consistency between local and global loss landscapes, \cref{fig:local_global_fedgloss} replicates the analysis previously conducted on \fedsam (\cref{fig:global_local_full}). The behavior of local models is shown from both local (``\textit{Local loss}'') and global perspectives (``\textit{Global loss}''). \cref{app:ablations:local-global} offers visualizations for the remaining clients. %
Compared to \fedsam, the gap between local and global loss landscapes in \ours is significantly smaller, and both global and local loss surfaces are found in \textit{flat and low-loss regions} (\cref{fig:consistency:c0_fedgloss,fig:consistency:c1_fedgloss}). This suggests our method effectively promotes convergence toward {aligned low-loss flat regions}, minimizing the discrepancy between local and global geometries. This results in a global model residing in a flat minimum in the global landscape (\cref{subfig:ours-fedsam-c10-a0,subfig:ours-fedsam-c100-a0}). \cref{fig:consistency:c0_fedsmoo,fig:consistency:c1_fedsmoo} instead compare \ours with the best-performing SOTA  \fedsmoo, where the position in the global landscape of \ours' local models is added for reference. While \fedsmoo improves consistency between local and global sharpness compared to \fedsam, it falls short of \ours in reaching a flatter global minimum. %

\cref{tab:eigs_fedgloss} confirms these claims. By combining ADMM for consistency and server-side \sam for global flatness, \ours prioritizes achieving a flatter \textit{global} region during training, as proven by the {lowest global maximum eigenvalue} $\lambda_{1,g}$ and larger $\lambda_{1,l}$, across all clients and methods. %

\subsection{Benchmarking FedGloSS against SOTA}
\label{subsec:sota}

\begin{figure}[t]
\vspace{-5pt}
   \centering
   \captionsetup{font=scriptsize}
   \captionsetup[sub]{font=scriptsize}
    \subfloat[][\cifarten]{\includegraphics[width=.48\linewidth]{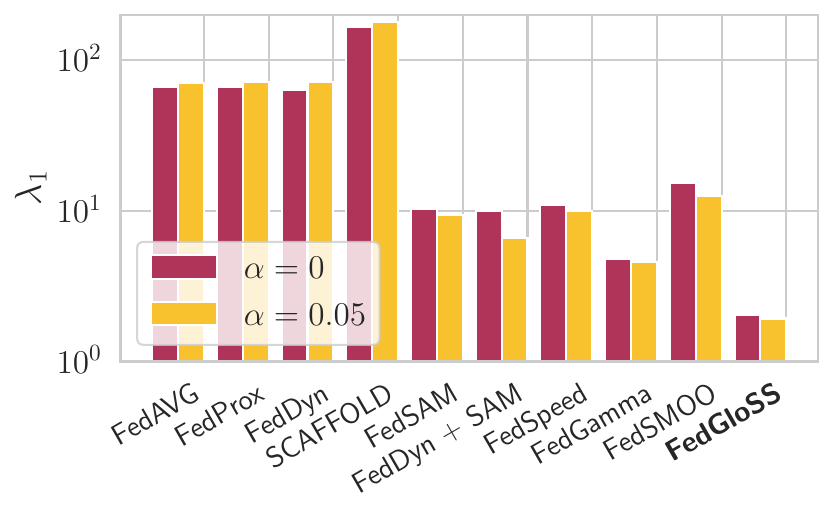}}
    \hfill
    \subfloat[][\cifar]{\raisebox{2.5pt}{\includegraphics[width=.48\linewidth]{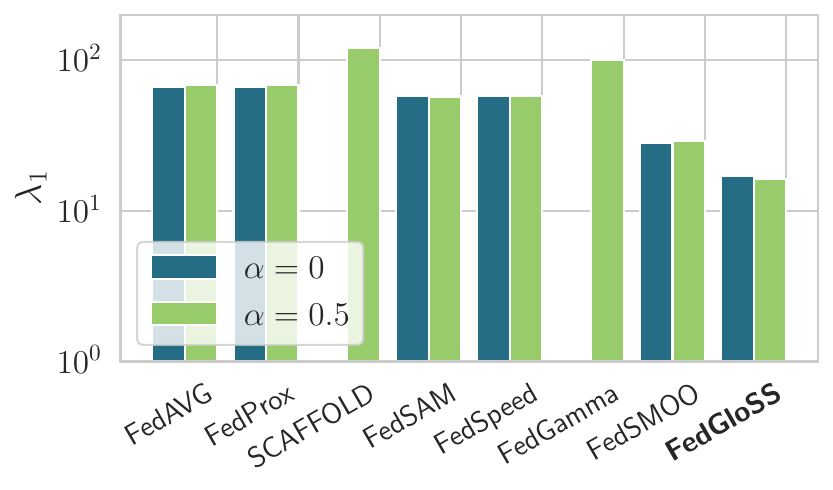}}}
    \vspace{-9pt}
    \caption{\textbf{Maximum Hessian eigenvalues} ($\lambda_1$), CNN. Values shown only if algorithm converged. \textbf{\ours reaches the 
    flattest global minima}. %
    }
   \label{fig:eigs}
   \vspace{-13pt}
\end{figure}

\setlength\tabcolsep{1.5pt}
\begin{table}
\centering
\captionsetup{font=scriptsize}
    \caption{\textbf{\ours \vs the state of the art} on \textsc{Cifar} datasets, distinguished by local optimizer, \textcolor{NavyBlue}{\sgd} (\textit{top}) and \textcolor{Bittersweet}{\sam} (\textit{bottom}), in terms of communication cost and accuracy (\%). Best results in \textbf{bold}. Model: {CNN}.}%
    \vspace{-8pt}
    \centering    
    \tiny
    \resizebox{\linewidth}{!}{
    \begin{tabular}{clccccccccc}
        \toprule
         &\multicolumn{1}{c}{\multirow{2}{*}{\textbf{Method}}} & \textbf{Comm.} & \multicolumn{3}{c}{\textbf{\cifarten}} && \multicolumn{3}{c}{\textbf{\cifar}} \\ 
        \cline{4-6} \cline{8-10}
         && \textbf{Cost} & {$\alpha=0$} && {$\alpha=0.05$} && {$\alpha=0$} && {$\alpha=0.5$} \\
        \midrule
        \multirow{5}{*}{\rotatebox[origin=c]{90}{\textcolor{NavyBlue}{Client \textbf{\sgd}}}} & \fedavg & \textcolor{ForestGreen}{\boldmath$1\times$} & $59.9\scriptscriptstyle{\pm 0.4}$ && $65.7\scriptscriptstyle{\pm 1.0}$ && $28.6\scriptscriptstyle{\pm 0.7}$ && $38.5\scriptscriptstyle{\pm 0.5}$ \\
        & \fedprox & \textcolor{ForestGreen}{\boldmath$1\times$} & $59.8\scriptscriptstyle{\pm 0.5}$ && $65.6\scriptscriptstyle{\pm 1.0}$ && $28.8\scriptscriptstyle{\pm 0.7}$ && $38.7\scriptscriptstyle{\pm} 0.4$\\
        &\feddyn & \textcolor{ForestGreen}{\boldmath$1\times$} & $65.5\scriptscriptstyle{\pm 0.3}$ && $70.1\scriptscriptstyle{\pm 1.2}$ &&  \textcolor{red}{\xmark} &&  \textcolor{red}{\xmark} \\
        &\scaffold & \textcolor{BrickRed}{\boldmath$2\times$} & $25.1\scriptscriptstyle{\pm 3.7}$ && $54.0\scriptscriptstyle{\pm 2.6}$ &&  \textcolor{red}{\xmark} && $30.0\scriptscriptstyle{\pm 1.1}$\\    \cdashline{2-11}[1pt/3pt] 
        &\textbf{\ours} & \textcolor{ForestGreen}{\boldmath$1\times$} & \boldmath$69.5\scriptscriptstyle{\pm 0.4}$ && \boldmath$75.5\scriptscriptstyle{\pm 0.3}$ && \boldmath$42.5\scriptscriptstyle{\pm 0.6}$ && \boldmath$47.9\scriptscriptstyle{\pm 0.5}$\\
        \midrule
        \multirow{6}{*}{\rotatebox[origin=c]{90}{\textcolor{Bittersweet}{Client \textbf{\sam}}}} &\fedsam & \textcolor{ForestGreen}{\boldmath$1\times$} & $70.2\scriptscriptstyle{\pm 0.9}$ && $71.5\scriptscriptstyle{\pm 1.08}$ && $28.7\scriptscriptstyle{\pm 0.5}$ && $39.6\scriptscriptstyle{\pm 0.5}$ \\
        &\feddyn & \textcolor{ForestGreen}{\boldmath$1\times$} & $79.3\scriptscriptstyle{\pm 3.1}$ && $81.5\scriptscriptstyle{\pm 0.6}$ &&  \textcolor{red}{\xmark} &&  \textcolor{red}{\xmark}\\
        &\fedspeed & \textcolor{ForestGreen}{\boldmath$1\times$} & $70.9\scriptscriptstyle{\pm 0.4}$ && $72.3\scriptscriptstyle{\pm 1.1}$ && $28.9\scriptscriptstyle{\pm 0.5}$ && $39.7\scriptscriptstyle{\pm 0.5}$ \\
        &\fedgamma & \textcolor{BrickRed}{\boldmath$2\times$} & $58.9\scriptscriptstyle{\pm 1.8}$ && $61.9\scriptscriptstyle{\pm 1.8}$ &&  \textcolor{red}{\xmark} && $29.4\scriptscriptstyle{\pm 1.4}$\\
        &\fedsmoo & \textcolor{BrickRed}{\boldmath$2\times$} & $81.3\scriptscriptstyle{\pm 0.5}$ && $82.8\scriptscriptstyle{\pm 0.6}$ && $47.8\scriptscriptstyle{\pm 0.5}$ && $51.7\scriptscriptstyle{\pm 0.46}$ \\
        \cdashline{2-11}[1pt/3pt]
        &\textbf{\ours} & \textcolor{ForestGreen}{\boldmath$1\times$} & \boldmath$83.9\scriptscriptstyle{\pm 0.4}$ && \boldmath$84.4\scriptscriptstyle{\pm 0.5}$ && \boldmath$50.6\scriptscriptstyle{\pm 0.6}$ && \boldmath$53.4\scriptscriptstyle{\pm 0.5}$ \\
        \bottomrule
    \end{tabular}}
    \label{tab:sota_cnn}
\vspace{-20pt}
\end{table}

This section compares \ours with SOTA methods on vision tasks in heterogeneous FL. \cref{app:settings:hom} presents results for $\alpha\in\{1,5,10\}$ on \cifarten and homogeneous settings. %

\subsubsection{FedGloSS on Standard Federated Benchmarks}
\label{subsec:fl_cifar}
\cref{tab:sota_cnn} presents results on \cifar and \cifarten with varying levels of heterogeneity on the CNN model.  %
Several observations highlight the advantages of \ours.  %
{\ours achieves the {best results} among both \sgd and \sam-based approaches while maintaining {communication efficiency}}. {\ours} with local \sam consistently outperforms the best-performing SOTA \fedsmoo by  $\approx 2.5$ percentage points in accuracy across all configurations \textit{with half the communication cost}. Our method reaches the {flattest global minima} (\eg, $\lambda_1^{\ours}=2.03$ \vs $\lambda_1^{\fedsmoo}=15.37$ on \cifarten $\alpha=0$), as shown in \cref{fig:eigs,fig:landscape_fedgloss_fedsam}, achieving the {best overall performance}. \ours with local \sgd overcomes by $\approx 4$ percentage points \textit{all} \sgd-based approaches. %
\feddyn suffers from parameter explosion in highly heterogeneous settings \citep{varno2022adabest}, failing to converge on \cifar. Differently, \ours successfully employs ADMM to align global and local solutions, with the best results under extreme heterogeneity. %
Studies showed \scaffold performs poorly in complex heterogeneous environments \citep{li2022federated,caldarola2022improving}, resulting in its inability to converge on \cifar alongside \fedgamma.   \cref{tab:sota_r18} confirms \ours' effectiveness, consistently outperforming
SOTA methods with the more complex {ResNet18} architecture, with $\approx 8$ points higher accuracy \wrt \fedavg with both \sgd and \sam, and $+5$ \wrt \fedsmoo, with the flattest solutions (\cref{subfig:ours-fedsam-c10-r18,subfig:ours-fedsmoo-c10-r18}).
\setlength\tabcolsep{1pt}
\begin{table}[t]
\vspace{-5pt}
\begin{minipage}{.49\linewidth}
\setlength\tabcolsep{1pt}
    \centering
    \captionsetup{font=scriptsize}
    \caption{\textbf{ResNet18} 
     on \cifar\\$\alpha=0.5$ and \cifarten $\alpha=0.05$.}
     \vspace{-5pt}
    \tiny
    \resizebox{\linewidth}{!}{
    \begin{tabular}{clccc}
    \toprule
         & \multicolumn{1}{l}{\textbf{Method}} & \textbf{Comm. cost} & \textsc{C100} \textbf{Acc.} & \textsc{C10} \textbf{Acc.}  \\
         \midrule
         \multirow{5}{*}{\rotatebox[origin=c]{90}{\textcolor{NavyBlue}{Client \sgd}}} & \fedavg & \textcolor{ForestGreen}{\boldmath$1\times$} & $37.4\scriptscriptstyle{\pm 0.2}$ & $72.6\scriptscriptstyle{\pm 0.1}$\\
         & \fedprox & \textcolor{ForestGreen}{\boldmath$1\times$} & $37.6\scriptscriptstyle{\pm 0.1}$ & $72.2\scriptscriptstyle{\pm 0.2}$\\
         & \feddyn & \textcolor{ForestGreen}{\boldmath$1\times$} & $38.8\scriptscriptstyle{\pm 0.6}$ & $70.2\scriptscriptstyle{\pm 0.6}$\\
         & \scaffold & \textcolor{BrickRed} {\boldmath$2\times$} & $38.6\scriptscriptstyle{\pm 0.1}$ & $70.8\scriptscriptstyle{\pm 0.6}$\\
         \cdashline{2-5}
         & \textbf{\ours} & \textcolor{ForestGreen}{\boldmath$1\times$} & \boldmath$46.7\scriptscriptstyle{\pm 0.6}$ & \boldmath$79.1\scriptscriptstyle{\pm 0.5}$\\
         \midrule
         \multirow{6}{*}{\rotatebox[origin=c]{90}{\textcolor{Bittersweet}{Client \sam}}} & \fedsam & \textcolor{ForestGreen}{\boldmath$1\times$} & $38.5\scriptscriptstyle{\pm 0.1}$ &$72.8\scriptscriptstyle{\pm 0.1}$ \\
         & \feddyn & \textcolor{ForestGreen}{\boldmath$1\times$} & $39.6\scriptscriptstyle{\pm 0.8}$ & $72.6\scriptscriptstyle{\pm 0.2}$\\
         & \fedspeed & \textcolor{ForestGreen}{\boldmath$1\times$} & $38.7\scriptscriptstyle{\pm 0.6}$ & $72.6\scriptscriptstyle{\pm 0.1}$ \\
         & \fedgamma & \textcolor{BrickRed}{\boldmath$2\times$} & $38.8\scriptscriptstyle{\pm 0.3}$ & $72.2\scriptscriptstyle{\pm 0.1}$\\
         & \fedsmoo & \textcolor{BrickRed}{\boldmath$2\times$} & $44.8\scriptscriptstyle{\pm 0.5}$ & $75.3\scriptscriptstyle{\pm 0.6}$ \\
         \cdashline{2-5}
         & \textbf{\ours} & \textcolor{ForestGreen}{\boldmath$1\times$} & \boldmath$47.2\scriptscriptstyle{\pm 0.2}$ & \boldmath$80.0\scriptscriptstyle{\pm 0.3}$ \\
         \bottomrule
    \end{tabular}}
    \label{tab:sota_r18}
\end{minipage}
\hfill
\begin{minipage}{.49\linewidth}
    \captionsetup{font=scriptsize}
    \centering
    \caption{\textbf{MobileNetv2} on \gld.}
    \vspace{-5pt}
    \resizebox{\linewidth}{!}{
    \tiny
    \begin{tabular}{lcc}
    \toprule
         \multicolumn{1}{l}{\textbf{Method}} & \textbf{Comm. cost} & \textbf{Accuracy} \\
         \midrule
         \fedavg & \textcolor{ForestGreen}{\boldmath$1\times$} & $56.3\scriptscriptstyle{\pm 0.2}$\\
         \fedprox & \textcolor{ForestGreen}{\boldmath$1\times$} & $55.0\scriptscriptstyle{\pm 0.2}$\\
         \feddyn & \textcolor{ForestGreen}{\boldmath$1\times$} & $55.2\scriptscriptstyle{\pm 0.6}$ \\
         \scaffold & \textcolor{BrickRed}{\boldmath$2\times$} & \textcolor{red}{\xmark} \\
         \fedsam & \textcolor{ForestGreen}{\boldmath$1\times$} & $56.7\scriptscriptstyle{\pm 0.1}$ \\
         \feddyn + \sam & \textcolor{ForestGreen}{\boldmath$1\times$} & $56.0\scriptscriptstyle{\pm 1.3}$\\
         \fedgamma & \textcolor{BrickRed}{\boldmath$2\times$} &  \textcolor{red}{\xmark}\\
         \fedsmoo & \textcolor{BrickRed}{\boldmath$2\times$} & $59.5\scriptscriptstyle{\pm 0.1}$ \\
         \cdashline{1-3}
         \textbf{\ours} & \textcolor{ForestGreen}{\boldmath$1\times$} & \boldmath$59.7\scriptscriptstyle{\pm 1.2}$ \\
         \bottomrule
    \end{tabular}}
    \label{tab:sota_gld}
\end{minipage}
\vspace{-15pt}
\end{table}

\vspace{-3pt}
\subsubsection{FedGloSS on Real-World Large-Scale Datasets}
\label{subsec:large_scale}
To further highlight \ours's effectiveness, we evaluate it on \textit{large-scale image classification} using the challenging \gld dataset. %
\cref{tab:sota_gld} compares \ours with local \sam against the best-performing baselines. \ours is among the few methods, alongside \fedsam and \fedsmoo, outperforming \fedavg. Similarly to the \cifar results (\cref{subsec:fl_cifar}), both \scaffold and \fedgamma fail to converge.  Importantly, {\ours achieves the best overall performance} ($+3.4\%$ \wrt \fedavg) with reduced communication overhead.

\vspace{-3pt}
\subsubsection{ADMM and SAM Interaction in FedGloSS} 
\begin{figure}[h]
\vspace{-9pt}
    \centering
    \captionsetup{font=scriptsize}
    \includegraphics[width=0.9\linewidth]{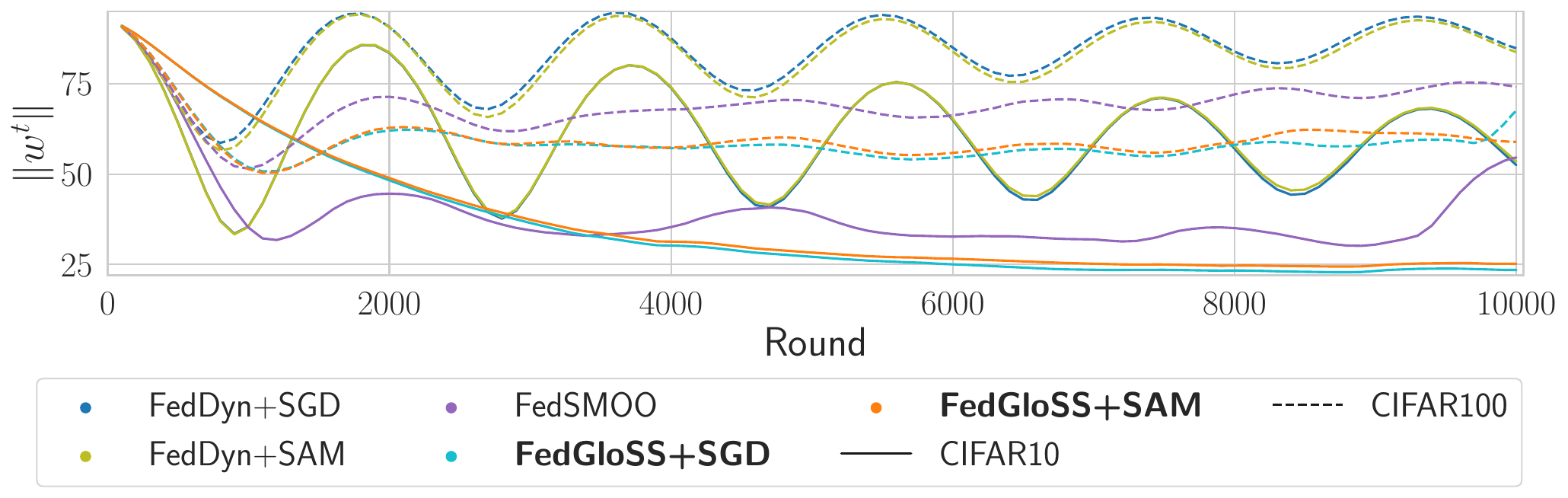}
    \vspace{-8pt}
    \caption{\textbf{Trend of model parameters norm},  $\|\w^t\|_2$, on \sam-based methods with ResNet18 on \textsc{Cifar} datasets. \textbf{\sam reduces the norm and the risk of parameters explosion, successfully enabling ADMM in heterogeneous FL}.}
    \label{fig:sam-admm}
    \vspace{-15pt}
\end{figure}

ADMM-based methods are often prone to parameter explosion in highly heterogeneous FL settings with many clients \cite{varno2022adabest}. This occurs as multiple gradients accumulate in the global dual variable $\sigma$ (\cref{sec:method}), causing the parameter norms to grow uncontrollably. However, empirical results indicate this issue is mitigated with \sam (\eg, see \feddyn \vs \ours in \cref{tab:sota_cnn}). We attribute this to \sam’s nature: by targeting flat minima, it promotes smaller gradient steps and minimal weight updates, resulting in a more stable algorithm. \cref{fig:sam-admm} confirms our hypothesis by showing \sam's stability effectively lowers parameter norms and the consequent risk of explosion, particularly when \sam is applied directly to the global model, as in \ours. 

\subsubsection{Communication Efficiency with FedGloSS}
\label{subsec:comm}
\begin{figure}
    \centering
    \captionsetup{font=scriptsize}
    \includegraphics[width=0.9\linewidth]{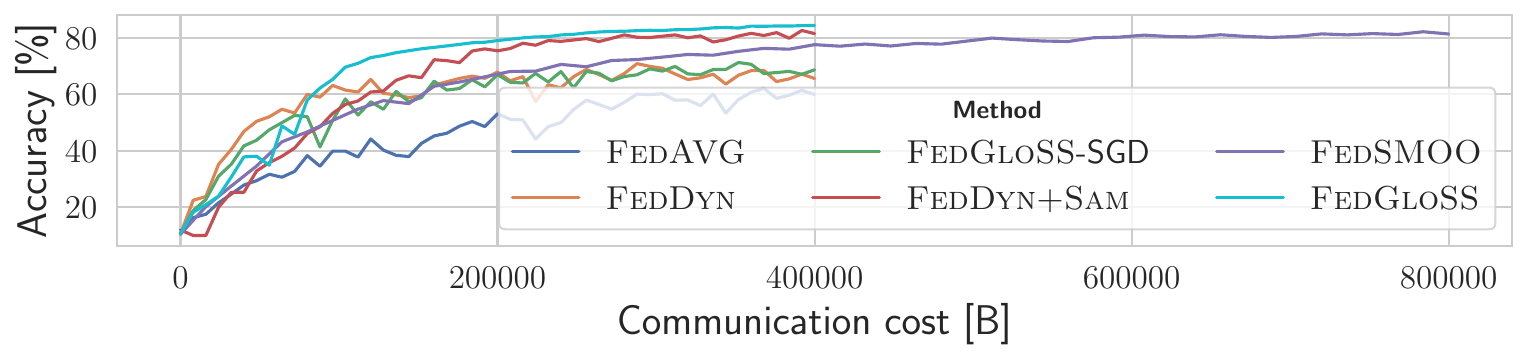}
    \vspace{-12pt}
    \caption{\textbf{Accuracy gain \vs communication cost.} CNN, \cifarten $\alpha=0$.}
    \vspace{-13pt}
    \label{fig:acc-v-comm}
    \vspace{-5pt}
\end{figure}

Communication cost is the main bottleneck in FL \citep{li2020review}, making its optimization a relevant challenge. As already previously highlighted, \ours considers communication efficiency its primacy concern. Defined $B$ the number of bits exchanged by \fedavg in $T$ training rounds, \cref{tab:comm_eff_2} studies \ours's communication cost against the SOTA baselines in terms of   %
rounds necessary to reach \fedavg's performance %
and quantity of exchanged bits. The ADMM-based methods are usually faster, with \ours being the fastest with ResNet18 and MobileNetv2. While \fedsmoo is faster when using the CNN model, the transmitted bits double due to its increased communication cost, making {\ours the most efficient method in all cases}. %
{Analyses on left-out settings in \cref{app:others:comm_eff}.} %
\cref{fig:acc-v-comm} further shows the \textit{total} communication cost of the FL algorithms (\textit{\# model exchanges per round} $\times |\mathcal{C}^t| \times \, T$). The doubled cost of \fedsmoo is clear, while \ours achieves nearly double \fedavg's accuracy at the same cost. 

\setlength\tabcolsep{.5pt}
\begin{table}[t]
\vspace{-17pt}
\captionsetup{font=scriptsize}
\caption{\textbf{Communication costs} comparison %
\wrt \textsc{FedAvg}. ``-'' for not reached accuracy, \quotes{ \textcolor{red}{\xmark}} for non-convergence. \textsc{Gldv2} is \textsc{Landmarks-User-160k}.}
\vspace{-8pt}
    \centering
    \tiny
    \resizebox{\linewidth}{!}{
    \begin{tabular}{clcccccc|cccccc|ccc}
    \toprule
        & \multirow{4}{*}{\textbf{Method}} && \multicolumn{5}{c}{\textbf{CNN}} && \multicolumn{5}{c}{\textbf{ResNet18}} && \multicolumn{2}{c}{\textbf{MobileNetv2}} \\
    \cline{4-8} \cline{10-14} \cline{16-17}
        &&& \multicolumn{2}{c}{{\textsc{Cifar10} $\alpha=0$}} && \multicolumn{2}{c}{{\textsc{Cifar100} $\alpha=0$}} && \multicolumn{2}{c}{{\textsc{Cifar10} $\alpha=0.05$}} && \multicolumn{2}{c}{{\textsc{Cifar100} $\alpha=0.5$}} && \multicolumn{2}{c}{{{\textsc{Gldv2}}}} \\
   \cline{4-5} \cline{7-8} \cline{10-11} \cline{13-14} \cline{16-17} %
        &&&  \textit{Rounds} & \textit{Cost} &&  \textit{Rounds} & \textit{Cost} && \textit{Rounds} & \textit{Cost} && \textit{Rounds} & \textit{Cost} && \textit{Rounds} & \textit{Cost} \\
    \midrule
        \multirow{5}{*}{\rotatebox[origin=c]{90}{\textcolor{NavyBlue}{Client \textbf{\textsc{Sgd}}}}} & \textsc{FedAvg} &&   $10k$ & $1B$ &&  $20k$ & $1B$ && $10k$ & $1B$ && $10k$ & $1B$ && $1.3k$ & $1B$ \\
        & \textsc{FedProx} &&  $7.6k$ & $0.8B$ &&  $18.7k$ & $0.9B$ && $8.8k$ & $0.9B$ && $8.3k$ & $0.8B$ && - & - \\
        & \textsc{FedDyn} &&   $2k$ & \boldmath$0.2B$ && \multicolumn{2}{c|}{ \textcolor{red}{\xmark}} && - & - && $3.5k$ & $0.4B$ && - & - \\
        & \textsc{Scaffold} &&   - & - &&  - & - && - & - && $8.9k$ & $1.8B$ && \multicolumn{2}{c}{ \textcolor{red}{\xmark}} \\
    \cdashline{2-17}[1pt/3pt] 
        & \textbf{\textsc{FedGloSS}} && $3.4k$ & $0.3B$ &&   $5k$ & \boldmath$0.3B$ && $2.4k$ & \boldmath$0.2B$ && $1.9k$ & \boldmath$0.2B$ && $1.3k$ & $1B$ \\
    \midrule
        \multirow{6}{*}{\rotatebox[origin=c]{90}{\textcolor{Bittersweet}{Client \textbf{\textsc{Sam}}}}} & \textsc{FedSam} &&  $6.3k$ & $0.6B$ &&  $18.3k$ & $0.9B$ && $9.2k$ & $0.9B$ && $7.8k$ & $0.8B$ && $1.3k$ & $1B$ \\
        & \textsc{FedDyn} &&   $3k$ & $0.3B$ &&  \multicolumn{2}{c|}{ \textcolor{red}{\xmark}} && $4.1k$ & $0.4B$ && $3.5k$ & $0.4B$ && - & - \\
        & \textsc{FedSpeed} &&   $6.3k$ & $0.6B$ &&   $18.3k$ & $$0.9B$$ && $8.3k$ & $0.8B$ && $8.3k$ & $0.8B$ && $1.3k$ & $1B$ \\
        & \textsc{FedGamma} &&   - & - &&  - & - && $9.3k$ & $1.9B$ && $8.1k$ & $1.6B$ && \multicolumn{2}{c}{ \textcolor{red}{\xmark}} \\
        & \textsc{FedSmoo} &&   $1.9k$ & $0.4B$   && $4.5k$ & $0.5B$ && $2.4k$ & $0.5B$ && $2.3k$ & $0.5B$ && $200$ & $0.4B$ \\
    \cdashline{2-17}[1pt/3pt] 
        & \textbf{\textsc{FedGloSS}} &&  $2.2k$ & \boldmath$0.2B$ &&  $6.3k$ & \boldmath$0.3B$ && $2.4k$ & \boldmath$0.2B$ && $1.9k$ & \boldmath$0.2B$ && $200$ & \boldmath$0.2B$ \\
    \bottomrule
    \end{tabular}}
    \label{tab:comm_eff_2}
    \vspace{-10pt}
\end{table}

\vspace{-5pt}
\subsection{Ablation Studies}
\label{subsec:ablation}
\vspace{-3pt}
\subsubsection{Communication-efficient Sharpness}
\label{subsec:naive_vs_real}
This section evaluates using the pseudo-gradient from the previous round $\Tilde{\Delta}_{\w}^{t-1}$ (\cref{math:eps_difference}) to estimate sharpness. \ours leverages past gradients as a reliable indication on the global loss landscape and aligns global and local paths via ADMM for consistent trajectories across rounds. 

\cref{tab:naive_comparison} compares \ours with its baseline, \naiveours (\cref{sec:method}), which computes the exact perturbation $\heps^t$ at the expense of doubled communication costs. %
\ours matches \naiveours in accuracy while maintaining communication efficiency, with minimal or negligible gap in performance, consistent with similar sharpness
($\lambda_1$). %
To fairly assess communication costs, 
\cref{tab:naive_comparison} also reports \ours' final accuracy \vs \naiveours' performance at 50\% training progress, showing \ours'  higher accuracy given the same number of exchanges. %
Overall, given $B$ bits transmitted by \ours, the cost of \naiveours is $1.96B$ on \textsc{C10} $\alpha=0$, $2B$ for $\alpha=0.05$, $1.94B$ on \textsc{C100} $\alpha=0$ and $1.97B$ for $\alpha=0.5$. Thus, \ours reduces communication costs by up to 50\%, with no trade-off in model performance. If \naiveours were preferred to maintain exact gradients, it would still achieve the \textit{best results} with costs \textit{lower or comparable} to \fedsmoo. 
In addition, our strategy in centralized settings lowers the performance \wrt \sam, thus reducing our centralized upper bound \wrt \naiveours. %
At equal performance, {\ours narrows the gap to the upper bound}: $-2.4\%$ on \cifarten with $\alpha=0$ and $-1.9\%$ with $\alpha=0.05$ \vs respectively $-3.2\%$ and $-2.7\%$ of \naiveours \wrt \sam. In \cifar instead, $-7\%$ on $\alpha=0$ and $-4.2\%$ on $\alpha=0.5$ of \ours \vs $-8.1\%$ and $-5.2\%$ of its baseline. %
Aiming to reduce communication bottlenecks and improve performance, these results validate choosing \ours over \naiveours. Further analyses in \cref{app:others:comm_eff}.

\setlength\tabcolsep{1pt}
\begin{table}[t]
\vspace{-17pt}
\captionsetup{font=scriptsize}
\caption{\textbf{\ours \vs \naiveours} in terms of communication cost, accuracy (50\% and 100\% of training) and maximum Hessian eigenvalue $\lambda_1$. \sam as \textsc{ClientOpt}. \textsc{Cifar} datasets with $\alpha=0$ (\textit{top}) and $\alpha = 0.05/0.5$ (\textit{bottom}). $\widetilde{\sam}$ is \sam with the sharpness approximation of \ours, using the previous gradient. %
}
\vspace{-8pt}
    \centering
    \resizebox{\linewidth}{!}{
    \begin{tabular}{lcccccccc}
    \toprule
          \multicolumn{1}{c}{\multirow{2}{*}{\textbf{Method}}}  & \textbf{Comm.} & \multicolumn{3}{c}{\textbf{\cifarten}} & & \multicolumn{3}{c}{\textbf{\cifar}}  \\ \cline{3-5} \cline{7-9}
          & \textbf{Cost}  & {Acc@50\%} & {Acc@100\%}  & $\lambda_{1}$ ($\downarrow$) &&  {Acc@50\%} & {Acc@100\%} & $\lambda_{1}$ ($\downarrow$) \\
         \midrule
           \naiveours & \textcolor{BrickRed}{\boldmath$2\times$} & $77.6\scriptscriptstyle{\pm 0.3}$ & \boldmath$83.9\scriptscriptstyle{\pm 0.2}$ & $2.78\scriptscriptstyle{\pm 0.13}$ %
         && \boldmath$42.6\scriptscriptstyle{\pm 0.8}$ & \boldmath$50.8\scriptscriptstyle{\pm 0.1}$ & \boldmath$16.93\scriptscriptstyle{\pm 0.27}$ \\
         \ours & \textcolor{ForestGreen}{\boldmath$1\times$} & \boldmath$78.9\scriptscriptstyle{\pm 0.5}$ %
         & \boldmath$83.9\scriptscriptstyle{\pm 0.4}$ & \boldmath$2.03\scriptscriptstyle{\pm 0.05}$ %
         && $39.5\scriptscriptstyle{\pm 0.9}$ %
         & $50.6\scriptscriptstyle{\pm 0.6}$ %
         & $17.18\scriptscriptstyle{\pm 0.97}$ %
         \\
         \midrule
           \naiveours & \textcolor{BrickRed}{\boldmath$2\times$} & $78.7\scriptscriptstyle{\pm 0.1}$ & \boldmath$84.4\scriptscriptstyle{\pm 0.2}$ & $2.75\scriptscriptstyle{\pm 0.09}$ %
         && \boldmath$49.4\scriptscriptstyle{\pm 0.5}$ & \boldmath$53.7\scriptscriptstyle{\pm 0.3}$ & \boldmath$15.84\scriptscriptstyle{\pm 0.52}$ \\
          \ours & \textcolor{ForestGreen}{\boldmath$1\times$} & \boldmath$79.7\scriptscriptstyle{\pm 0.4}$ %
          & \boldmath$84.4\scriptscriptstyle{\pm 0.5}$ & \boldmath$1.93\scriptscriptstyle{\pm 0.03}$ %
         && $47.2\scriptscriptstyle{\pm 1.1}$ %
         & $53.4\scriptscriptstyle{\pm 0.5}$ %
         & $16.22\scriptscriptstyle{\pm 0.35}$ %
         \\
     \midrule
     \textbf{Centralized} && \multicolumn{3}{l}{{$87.1$ \sam \quad  $86.3  \, \widetilde{\sam}$}} && \multicolumn{3}{l}{$58.4$ \sam \quad  $57.6 \, \widetilde{\sam}$}\\
     \bottomrule
    \end{tabular}}
    \label{tab:naive_comparison}
    \vspace{-15pt}
\end{table}

\vspace{-3pt}
\subsubsection{The Role of Global Consistency and Flatness}
\label{subsec:serversam}
\cref{tab:ablation_global_sharpness} isolates the impact of global consistency and global sharpness minimization in \ours. %
We recall \fedavg with client-side \sam is \fedsam and using ADMM only for aligning local and global convergence points is \feddyn. Both components significantly impact performance, with their combination leading {\ours to the best overall results}. %
\ours is not prone to parameter explosion, achieving the best results even where \feddyn fails to converge (\textcolor{red}{\xmark}). The flatness of \ours' solutions \wrt \fedsam in \cref{fig:landscape_fedgloss_fedsam} confirms the efficacy of its strategy. %

\setlength{\tabcolsep}{1pt}
\begin{table}[h]
\vspace{-8pt}
\captionsetup{font=scriptsize}
\caption{Efficacy of global sharpness minimization in \ours: ADMM for global consistency and server-side \sam for global sharpness minimization lead to the best performance. \textsc{Cifars}, CNN with $\alpha=0$ and ResNet18 with $\alpha\in\{0.05, 0.5\}$.} %
\vspace{-8pt}
    \centering
    \tiny
    \resizebox{\linewidth}{!}{
    \begin{tabular}{ccccccccccccc}
    \toprule
         \textbf{\textsc{Client}} & \multirow{2}{*}{\textbf{Method}}  & \textbf{Global} & \textbf{Global} & \multicolumn{2}{c}{\textbf{CNN}} && \multicolumn{2}{c}{\textbf{ResNet18}} \\ \cline{5-6} \cline{8-9}
         \textbf{\textsc{Opt}} & & \textbf{Consistency}& \textbf{Flatness} & \cifarten & \cifar && \cifarten & \cifar\\
        \midrule
         \multirow{3}{*}{SAM} & \fedsam & \xmark & \xmark & $70.2\scriptscriptstyle{\pm 0.9}$  & $28.7\scriptscriptstyle{\pm 0.5}$ && $72.8\scriptscriptstyle{\pm 0.1}$ & $38.5\scriptscriptstyle{\pm 0.1}$ \\ 
         & \feddyn  & \textcolor{ForestGreen}{\cmark} & \xmark & $79.3\scriptscriptstyle{\pm 3.1}$ & \textcolor{red}{\xmark} && $72.6\scriptscriptstyle{\pm 0.2}$ & $39.6\scriptscriptstyle{\pm 0.8}$\\
         & \ours & \textcolor{ForestGreen}{\cmark} & \textcolor{ForestGreen}{\cmark} & \boldmath$83.9\scriptscriptstyle{\pm 0.4}$  & \boldmath$50.6\scriptscriptstyle{\pm 0.6}$ && \boldmath$80.0\scriptscriptstyle{\pm 0.3}$ & \boldmath$47.2\scriptscriptstyle{\pm 0.2}$\\ 
         \midrule
         \multirow{3}{*}{SGD} & \fedavg & \xmark & \xmark & $59.9\scriptscriptstyle{\pm 0.4}$  & $28.6\scriptscriptstyle{\pm 0.7}$ && $72.6\scriptscriptstyle{\pm 0.1}$ & $37.4\scriptscriptstyle{\pm 0.2}$\\
        & \feddyn & \textcolor{ForestGreen}{\cmark} & \xmark & $65.5\scriptscriptstyle{\pm 0.3}$  &  \textcolor{red}{\xmark} && $70.2\scriptscriptstyle{\pm 0.6}$ & $38.8\scriptscriptstyle{\pm 0.6}$\\
         & \ours & \textcolor{ForestGreen}{\cmark} & \textcolor{ForestGreen}{\cmark} & \boldmath$69.5\scriptscriptstyle{\pm 0.4}$   & \boldmath$42.5\scriptscriptstyle{\pm 0.6}$ && \boldmath$79.1\scriptscriptstyle{\pm 0.5}$ & \boldmath$46.7\scriptscriptstyle{\pm 0.6}$\\
    \bottomrule
    \end{tabular}}
    \label{tab:ablation_global_sharpness}
    \vspace{-20pt}
\end{table}

\section{Conclusion}
\vspace{-5pt}

This work tackled the challenge of limited generalization in heterogeneous Federated Learning (FL), prioritizing communication efficiency for real-world use. Building on research linking poor generalization to sharp minima in the loss landscape, we showed data heterogeneity worsens discrepancies between local and global loss surfaces, a problem not resolved by methods focusing only on local sharpness. To address this, our \ourslong\ (FedGloSS) finds flat minima in the \textit{global} loss landscape with server-side Sharpness-Aware Minimization and achieves communication efficiency by approximating SAM's sharpness through past global pseudo-gradients, distinguishing it from prior approaches. This work revealed SAM prevents ADMM-related parameter explosion by guiding optimization along flat directions, enabling stable updates in heterogeneous FL. 
Extensive evaluations showed FedGloSS outperforms SOTA methods in accuracy, flatness and communication efficiency. %

\paragraph{Acknowledgments.} This study was carried out within the FAIR - Future Artificial Intelligence Research and received funding from the European Union Next-GenerationEU (PIANO NAZIONALE DI RIPRESA E RESILIENZA (PNRR) – MISSIONE 4 COMPONENTE 2, INVESTIMENTO 1.3 – D.D. 1555 11/10/2022, PE00000013). This manuscript reflects only the authors’ views and opinions, neither the European Union nor the European Commission can be considered responsible for them.  We acknowledge the CINECA award under the ISCRA initiative for the availability of high-performance computing resources and support.

{
    \small
    \bibliographystyle{ieeenat_fullname}
    \bibliography{main}
}

\newpage

\onecolumn
\appendix
\begin{center}
    \section*{Beyond Local Sharpness: Communication-Efficient \\Global Sharpness-aware
Minimization for Federated Learning\\\textit{Appendix}}
\end{center}

\section{Algorithm}
\label{app:alg}

\cref{alg:fedgloss} summarizes \ours, using as an example \sgd or \sam as local optimizers, differently highlighted. The comment colors replicate those of \cref{fig:method}.

\begin{algorithm}[h]
    \centering
    \small
    \captionsetup{font=small}
    \caption{{\ours with \sambox{\sam} or \sgdbox{\sgd} as local optimizers}}
    \label{alg:fedgloss}
    \begin{algorithmic}[1]  
        \State \textbf{Input:} Global model $\w$, clients $\C$, rounds $T$, local iterations $E$, clients' learning rate $\eta$, clients' \sam neighborhood size $\rho_l$, \ours neighborhood size $\rho$, Lagrangian hyperparameter $\beta$.
        \State \textbf{Initialize:} $\w^{0}$, $\sigma^{0} = \sigma_k = 0$, $\Delta_{\w}^{0} = 0.$
        \For{each round $t \in [1, T]$}
            \State $\teps^t(\w) = \rho \frac{\Delta_{\w}^{t-1}}{\|\Delta_{\w}^{t-1}\|_2}$ \Comment{\textcolor{DarkOrchid}{Global perturbation with previous pseudo-grad}}
            \State 
            $\tilde{\w}^{t} = \w^{t} + \teps^t(\w) $  %
            \hfill\Comment{{Server-side approximated \textbf{\textsc{FedGloSS} ascent step}}}
            \State Randomly select a subset of clients $\C^t\subset \C$ 
            \For{each client $k \in \C^t$ in parallel}
                \State %
                $\w_{k, 0} = \tilde{\w}^{t}$ \hfill\Comment{{Initialize local model with \textit{perturbed} global model $\tw^{t}$}}
                \State Set iteration counter $i=1$  
                \For{each epoch $e \in [1, E]$} 
                    \For{each batch $\mathcal{B} \in \mathcal{D}_k$}
                        \State{\sgdbox{$\boldsymbol{g}_{k, i} = \nabla f_\mathcal{B} (\w_{k, i-1})$}} \hfill\Comment{Local \sgd gradient}
                        \State \sambox{$\hat{\boldsymbol{\epsilon}}_{k, i} = \rho_l \frac{\boldsymbol{g}_{k, i} }{\|\boldsymbol{g}_{k, i}\|_2}$} \hfill\Comment{\sam local perturbation}
                        \State \sambox{$\boldsymbol{g}_{k, i} = \nabla f_\mathcal{B} (\w_{k, i-1} + \hat{\boldsymbol{\epsilon}}_{k, i})$} \hfill\Comment{Local sharpness-aware gradient}
                        \State $\w_{k, i} \leftarrow \w_{k, i-1} - \eta [\boldsymbol{g}_{k, i} - \sigma_k + (\w_{k, i-1} - \w_{k, 0}) / \beta]$ \hfill\Comment{Local update with ADMM}
                        \State $i \leftarrow i + 1$
                    \EndFor
                \EndFor
                \State $\sigma_k \leftarrow \sigma_k - (\w_{k, E} - \tilde{\w}^{t}) / \beta$ \hfill\Comment{Update local dual variable}
                \State Send back to the server the local updated model $\w_k^t = \w_{k, E}$
            \EndFor
            \State $\sigma^{t+1} = \sigma^{t} - \frac{1}{\beta |\mathcal{C}|} \sum_{k \in \mathcal{C}}  (\w_k^{t} - \tilde{\w}^{t})$ \Comment{Update global dual variable}
            \State  $\Tilde{\Delta}_{\w}^{t} =  \sum_{k\in\C^t} \frac{N_k}{N} (\tilde{\w}^t - \w_k^t)$\Comment{\textbf{Global pseudo-gradient}}
            \State  %
            $\w^{t+1} = \w^{t} - \Tilde{\Delta}_{\w}^{t} - \beta \sigma^{t+1}$ \Comment{\textcolor{RoyalBlue}{\textsc{FedGloSS} descent step w/ pseudo-grad using ADMM}}%
        \EndFor
    \end{algorithmic}
\end{algorithm}

\section{The Benefits on \ours}
\label{app:benefits}
\subsection{Achieving Consistency on Local and Global Sharpness with FedGloSS}
\label{app:ablations:local-global}

\begin{figure}[h]
    \centering
    \captionsetup{font=small}
    \captionsetup[sub]{font=small}
    \subfloat[Local model trained on class \texttt{bear} with \ours][Local model trained\\on class \texttt{bear} with \ours]{\includegraphics[width=.28\linewidth]{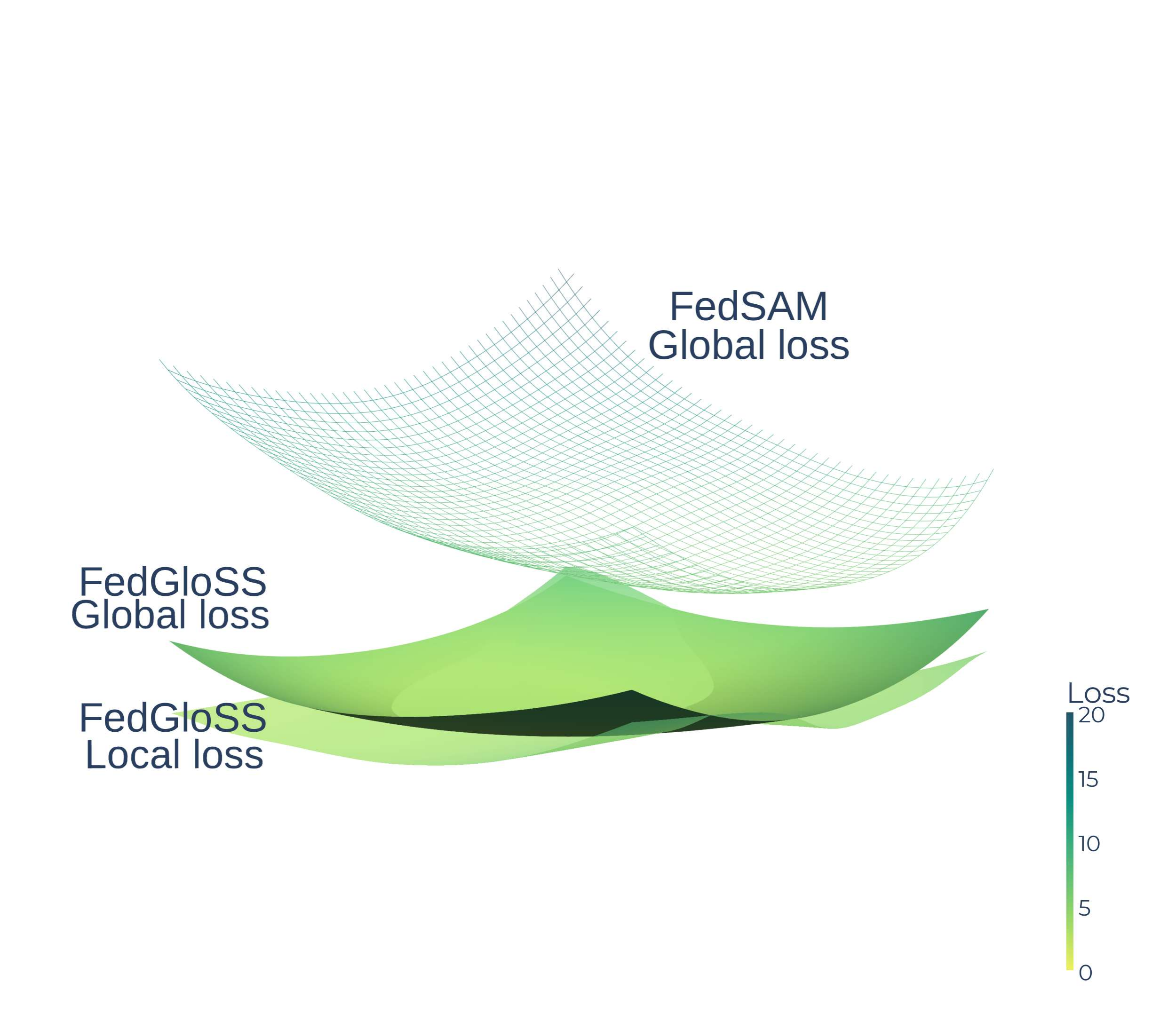}}
    \hfill
    \subfloat[Local model trained on class \texttt{skyscraper} with \ours][Local model trained on\\class \texttt{skyscraper} with \ours]{\includegraphics[width=.28\linewidth]{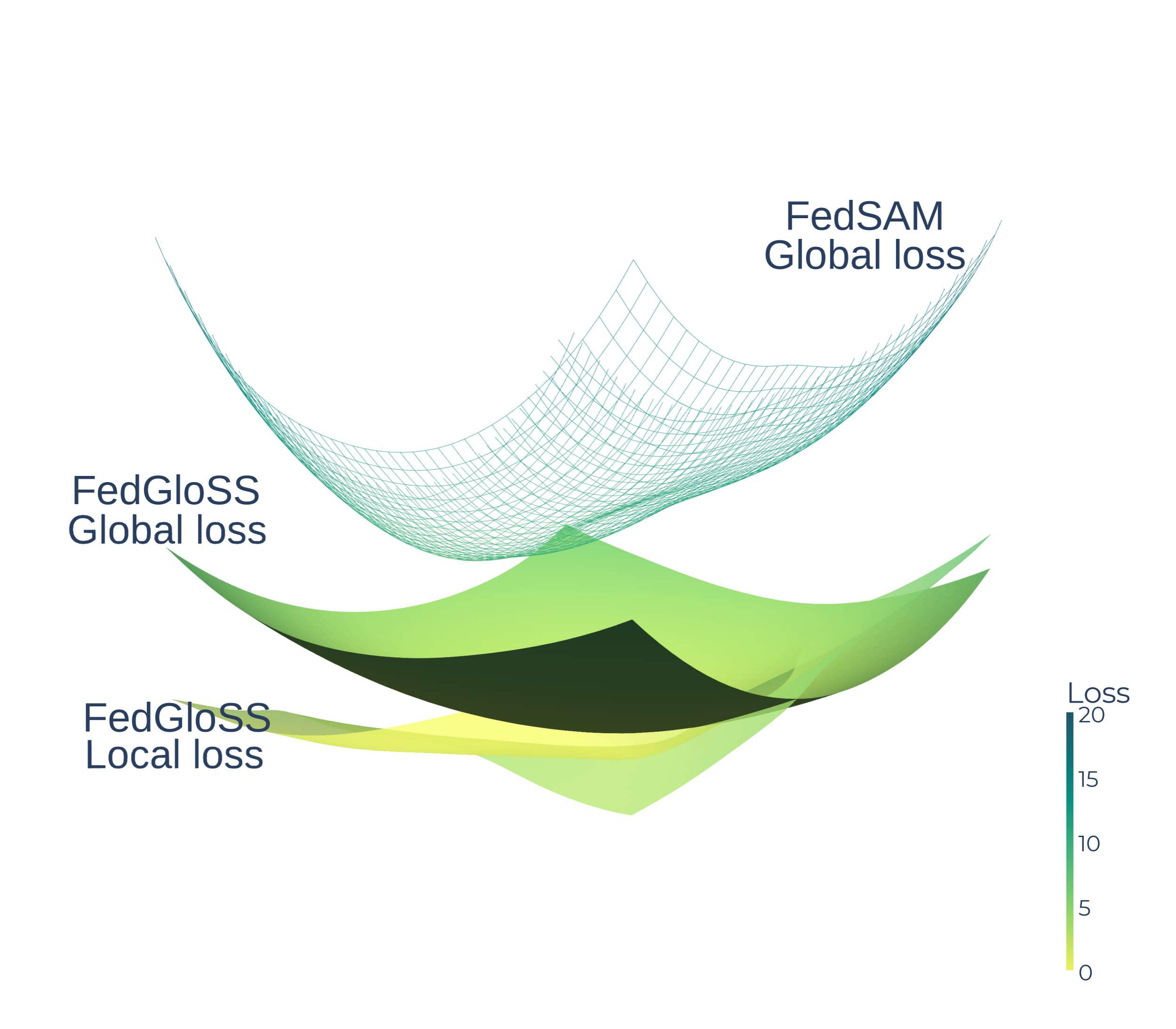}}
    \hfill
    \subfloat[Local model trained on class \texttt{possum} with \ours][Local model trained\\on class \texttt{possum} with \ours]{\includegraphics[width=.28\linewidth]{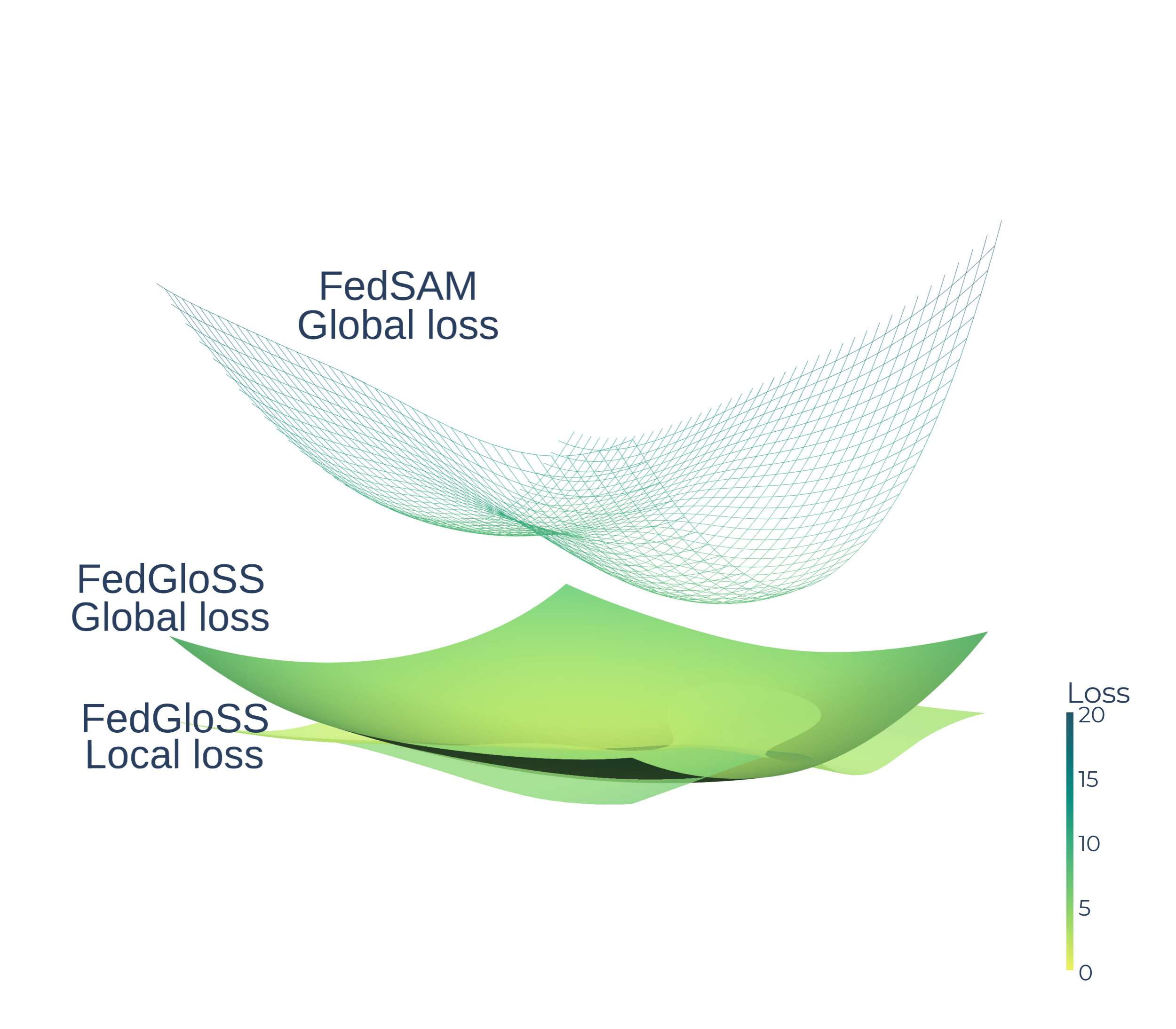}}
    \caption{\textbf{Global \vs local perspective on \ours}. \cifar $\alpha=0$ with \sam as local optimizer @ $20k$ rounds on CNN.  \textbf{(a) - (c):} Local models trained on one \texttt{class}, tested on the local (``\textit{Local loss}'') or global dataset (``\textit{Global loss}''). Corresponding global perspective of local model trained with \fedsam (\textit{net}) added as reference. %
    }
    \label{fig:local_global_fedgloss_appendix}
    \vspace{-15pt}
\end{figure}

This section completes the analyses presented in \cref{sec:inconsistency,subsec:landscape_consistency}, showing how \ours guides towards consistent local and global flat loss landscapes. \cref{fig:local_global_fedgloss_appendix} extends \cref{fig:consistency:c0_fedgloss,fig:consistency:c1_fedgloss} from the main paper with the 3 remaining  clients (out of 5) selected in the last round, comparing the behavior of clients' models trained with \ours from the local and global perspectives.  These results highlight the effectiveness of \ours in achieving local models with consistent local-global behavior. Indeed, these models end up in flat regions both in local and global loss landscapes.  The comparison with \fedsam (\textit{net} surface) further demonstrates the effectiveness of using  our global flatness-aware approach.

\begin{figure}[]
    \centering
    \captionsetup{font=small}
    \captionsetup[sub]{font=small}
    \subfloat[Local model trained on class \texttt{bear} with \ours][Local model trained\\on class \texttt{bear} with \ours]{\includegraphics[width=.28\linewidth]{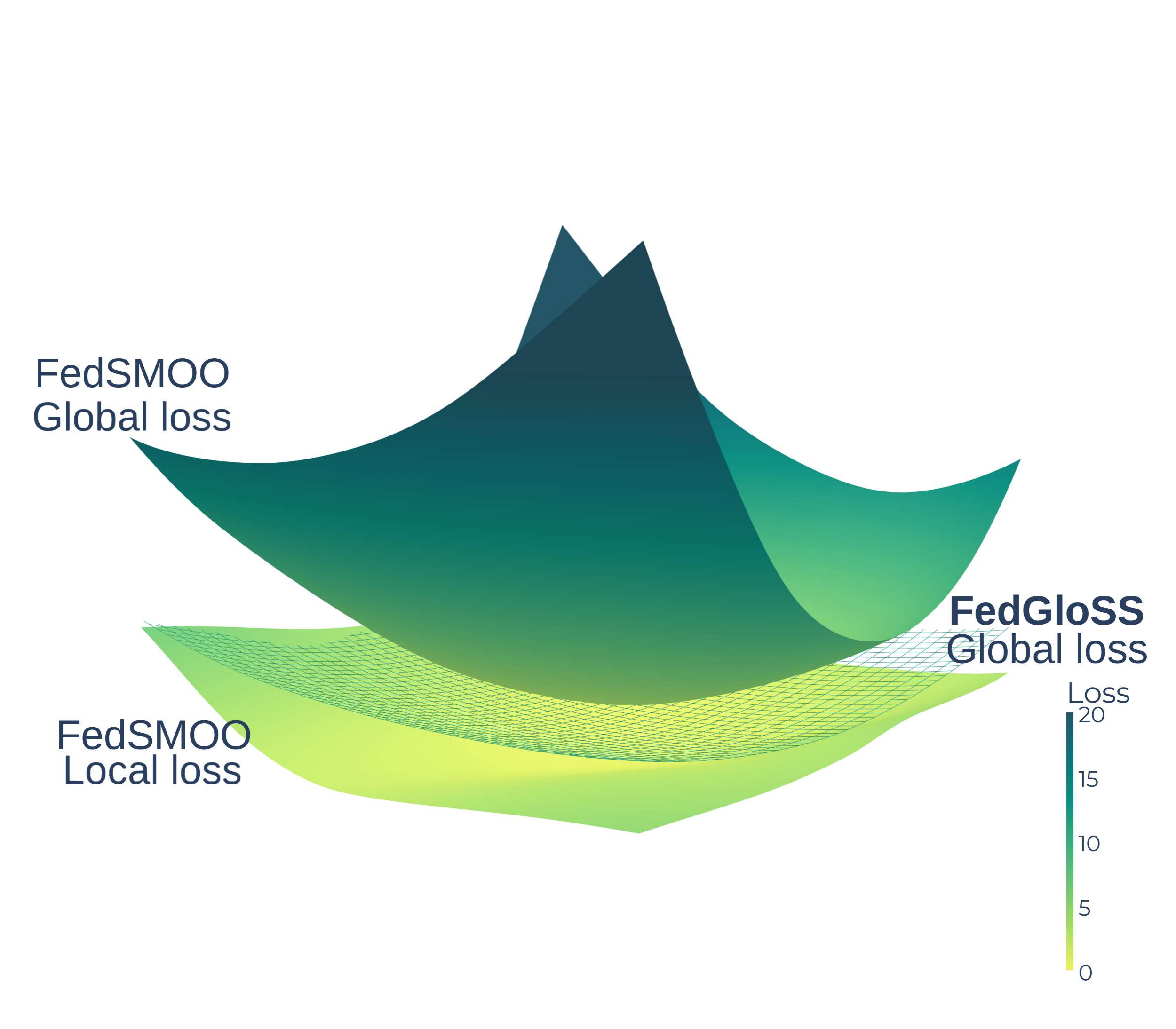}}
    \hfill
    \subfloat[Local model trained on class \texttt{skyscraper} with \ours][Local model trained on\\class \texttt{skyscraper} with \ours]{\includegraphics[width=.28\linewidth]{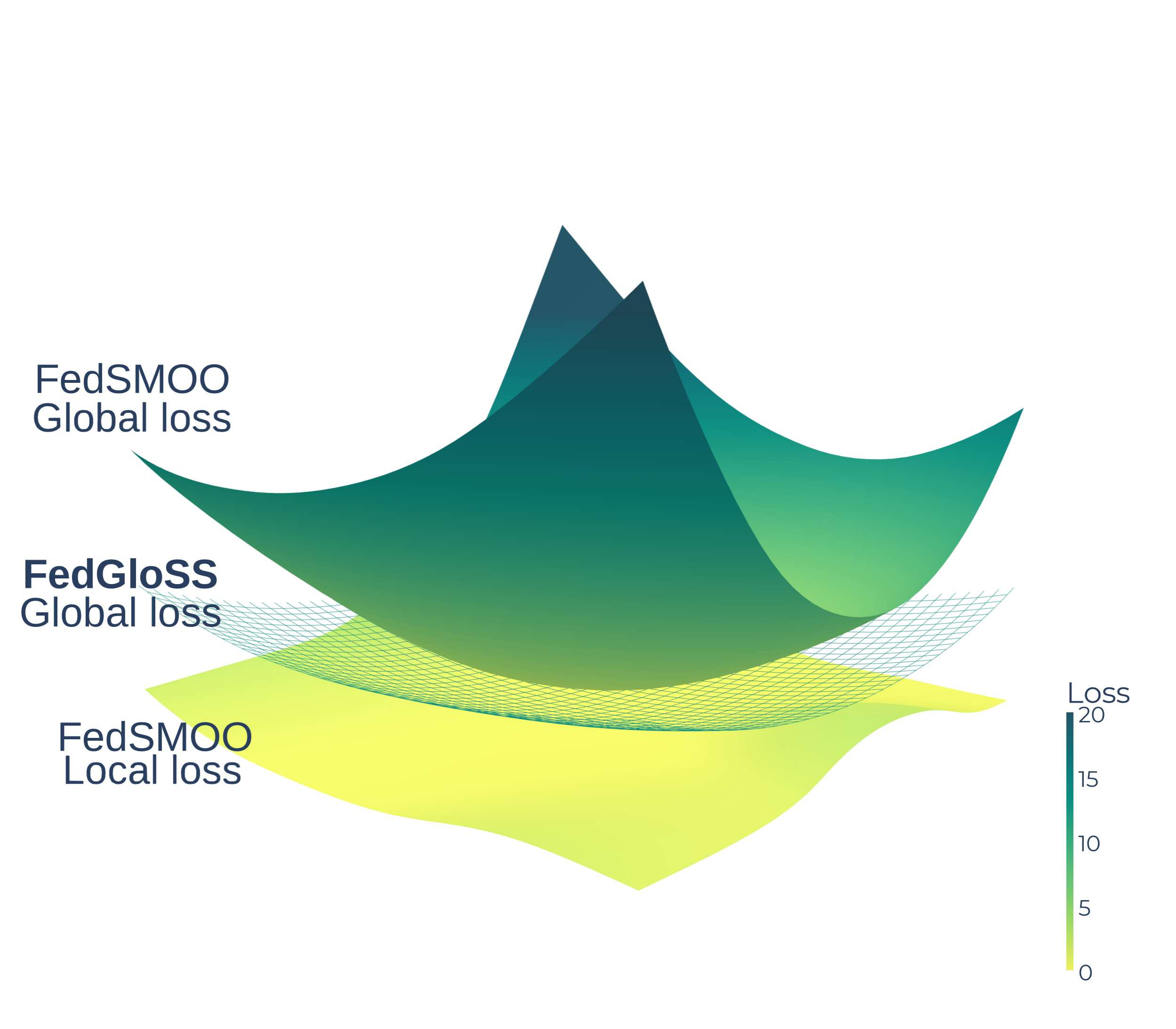}}
    \hfill
    \subfloat[Local model trained on class \texttt{possum} with \ours][Local model trained\\on class \texttt{possum} with \ours]{\includegraphics[width=.28\linewidth]{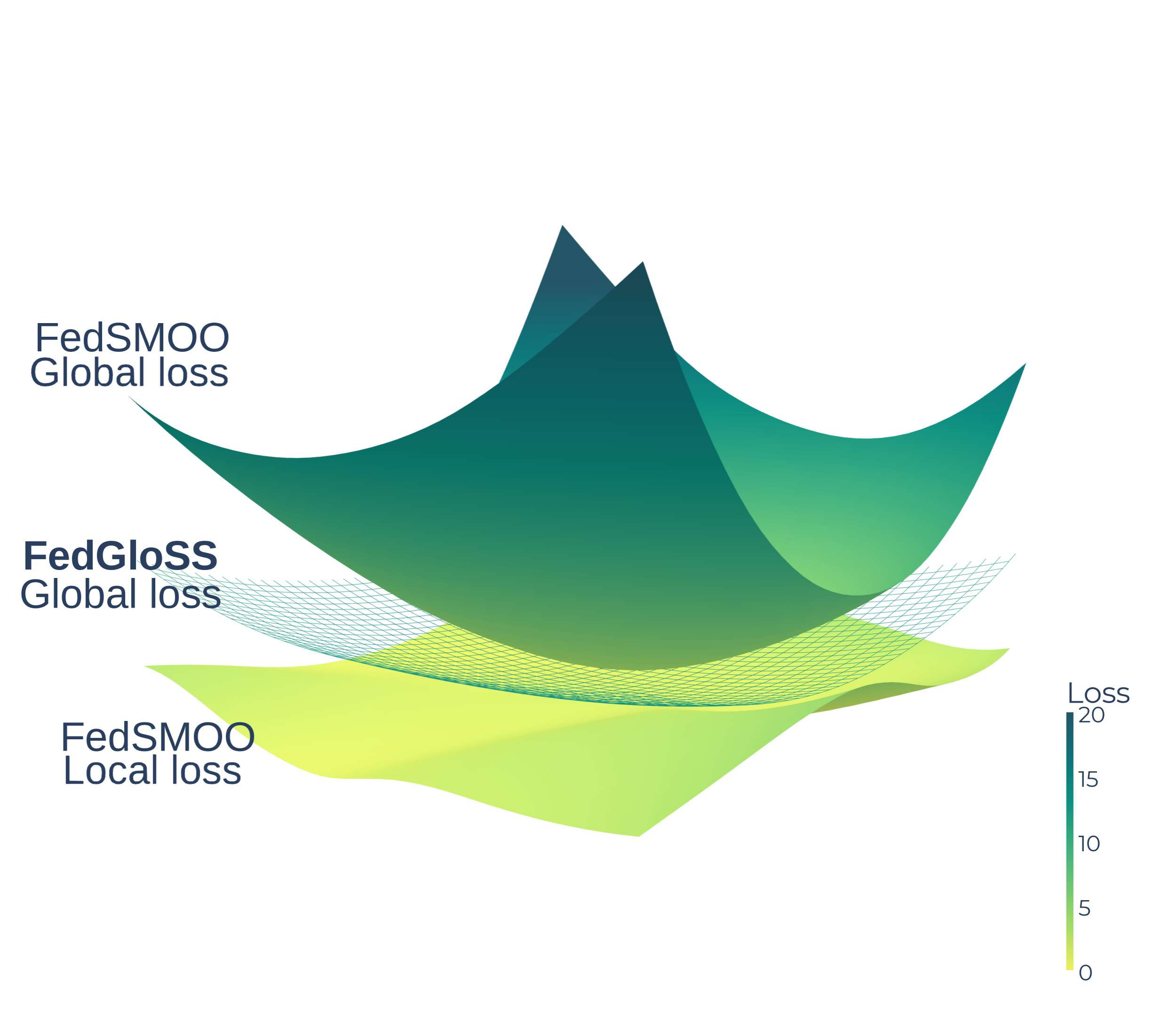}}
    \hfill
    \caption{\textbf{Global \vs local perspective on \fedsmoo}. \cifar $\alpha=0$ with \sam as local optimizer @ $20k$ rounds on CNN.  \textbf{(a) - (c):} Local models trained on one \texttt{class}, tested on the local (``\textit{Local loss}'') or global dataset (``\textit{Global loss}''). Corresponding global perspective of local model trained with \ours (\textit{net}) added as reference. %
    }
    \label{fig:local_global_fedsmoo_appendix}
\end{figure}

\cref{fig:local_global_fedsmoo_appendix} instead extends \cref{fig:consistency:c0_fedsmoo,fig:consistency:c1_fedsmoo}, offering a comparative analysis of \fedsmoo's clients' models \wrt \ours (\textit{net} landscape). The local solutions found by \ours achieve flatter, better (\ie, lower loss) and more consistent convergence points in the \textit{global} loss landscape \wrt our main competitor \fedsmoo. %

\subsection{Achieving Flatter Global Minima}
\label{app:plots:loss_landscapes}
\begin{figure}[t]
    \centering
    \captionsetup{font=small}
    \captionsetup[sub]{font=small}
    \subfloat[][\cifarten $\alpha=0$]{\includegraphics[width=.20\linewidth]{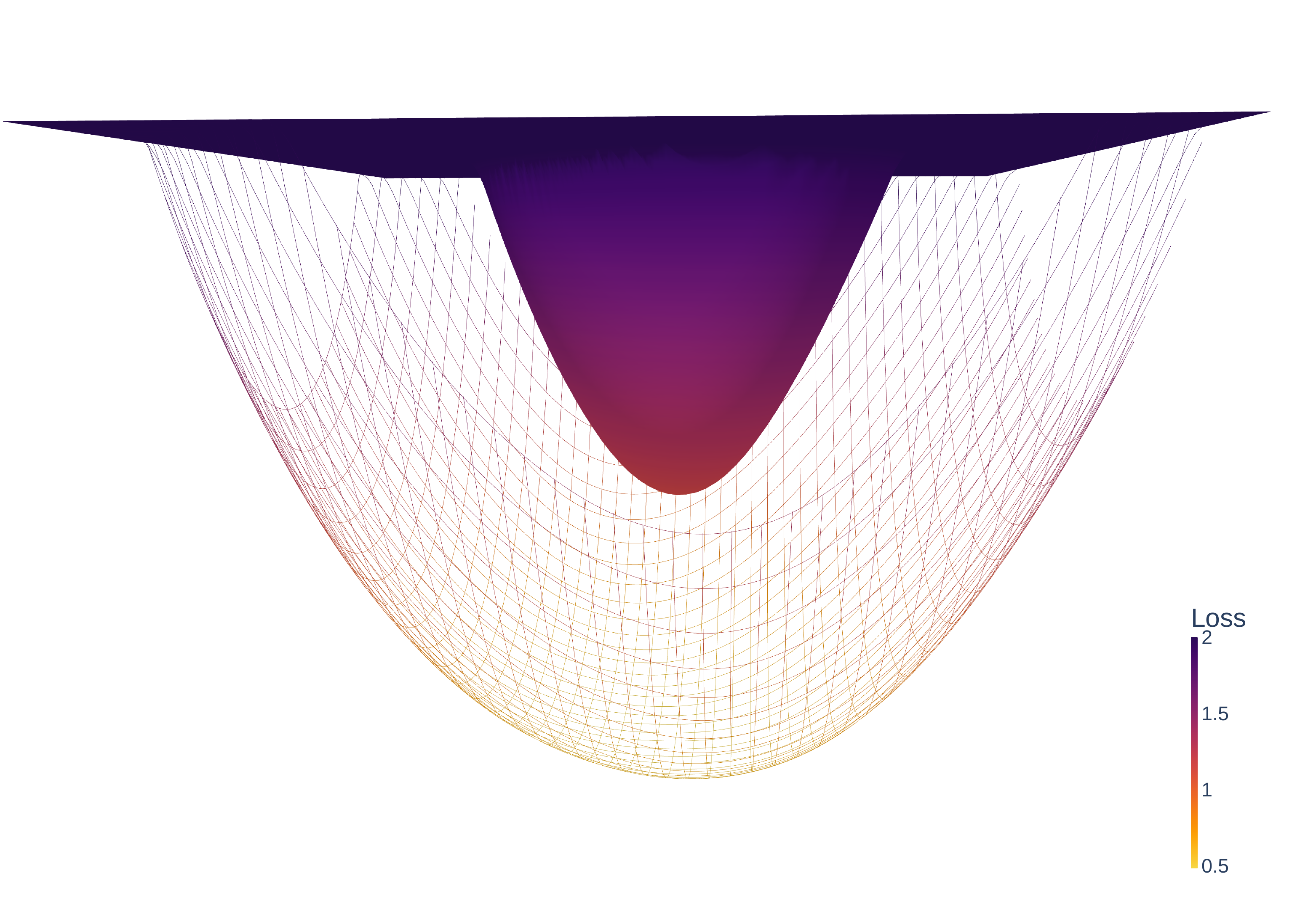}}
    \hfill
    \subfloat[][\cifarten $\alpha=0.05$]{\includegraphics[width=.20\linewidth]{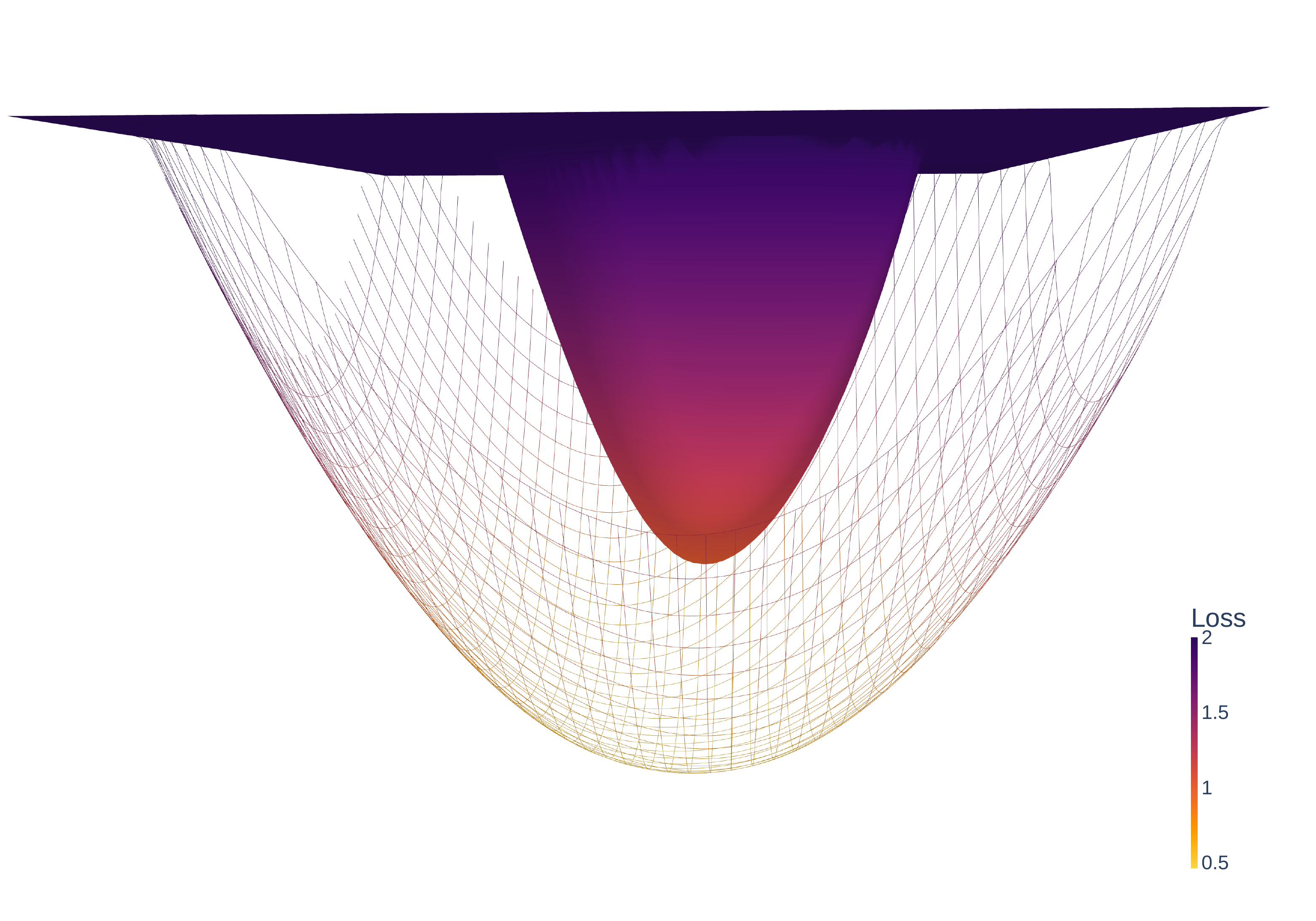}}
    \hfill
    \subfloat[][\cifar $\alpha=0$]{\includegraphics[width=.20\linewidth]{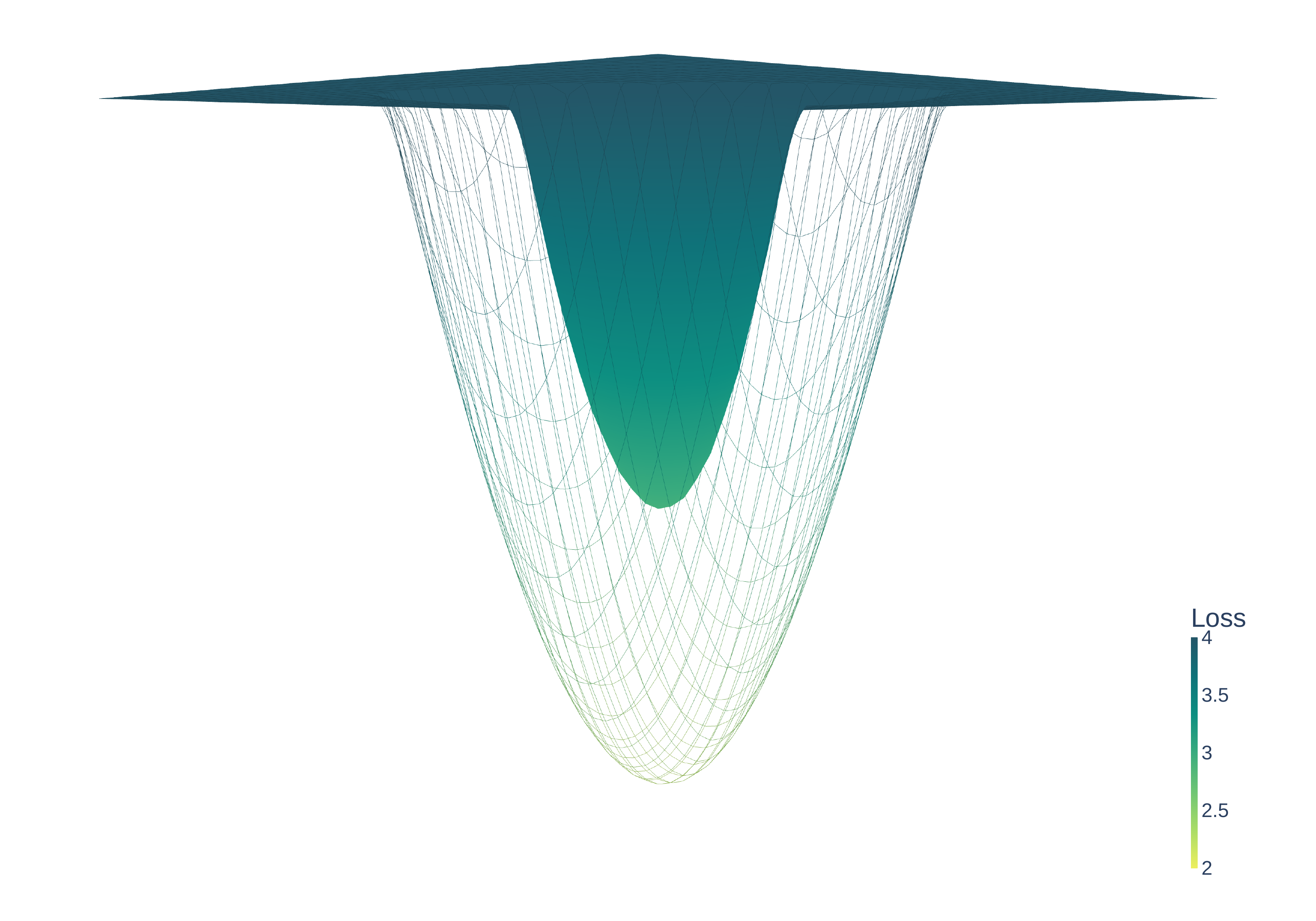}}
    \hfill
    \subfloat[][\cifar $\alpha=0.5$]{\includegraphics[width=.20\linewidth]{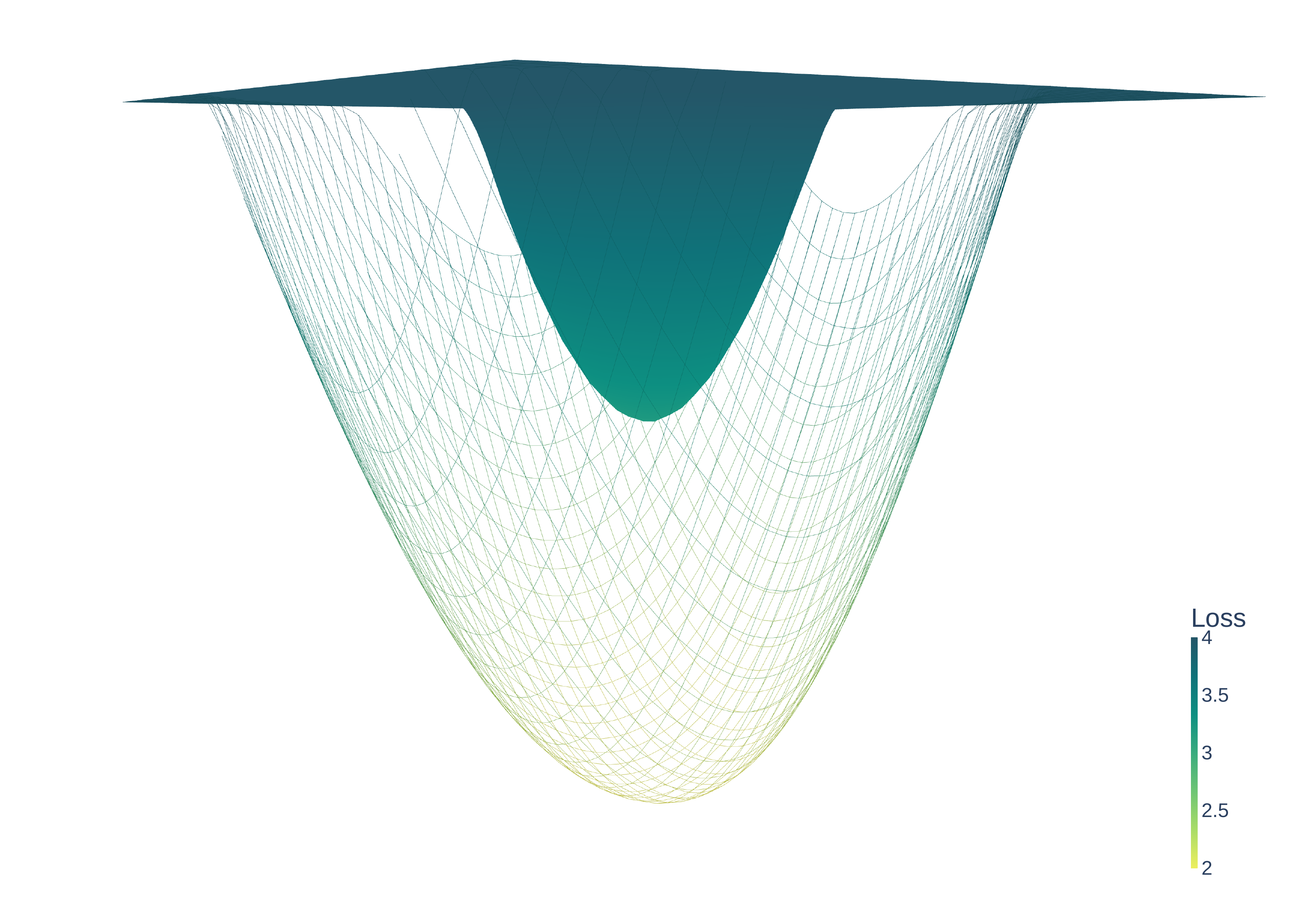}}
    \caption{{Visualization of the loss landscapes of the CNN trained with \textbf{\ours} (\textit{net}) and the de-facto standard optimization algorithm in FL \textbf{\fedavg} (\textit{solid}). Comparison with varying degrees of heterogeneity on \cifarten (left) and \cifar (right). \textbf{\ours consistently achieves flatter minima and lower loss values in the \textit{global} loss landscape.}}}
    \label{fig:landscapes_fedavg_fedgloss}
\end{figure}

\begin{figure}[t]
    \centering
    \captionsetup{font=small}
    \captionsetup[sub]{font=small}
    \subfloat[][\cifarten $\alpha=0.05$ \\ \ours \vs \fedsam]{\includegraphics[width=.20\linewidth]{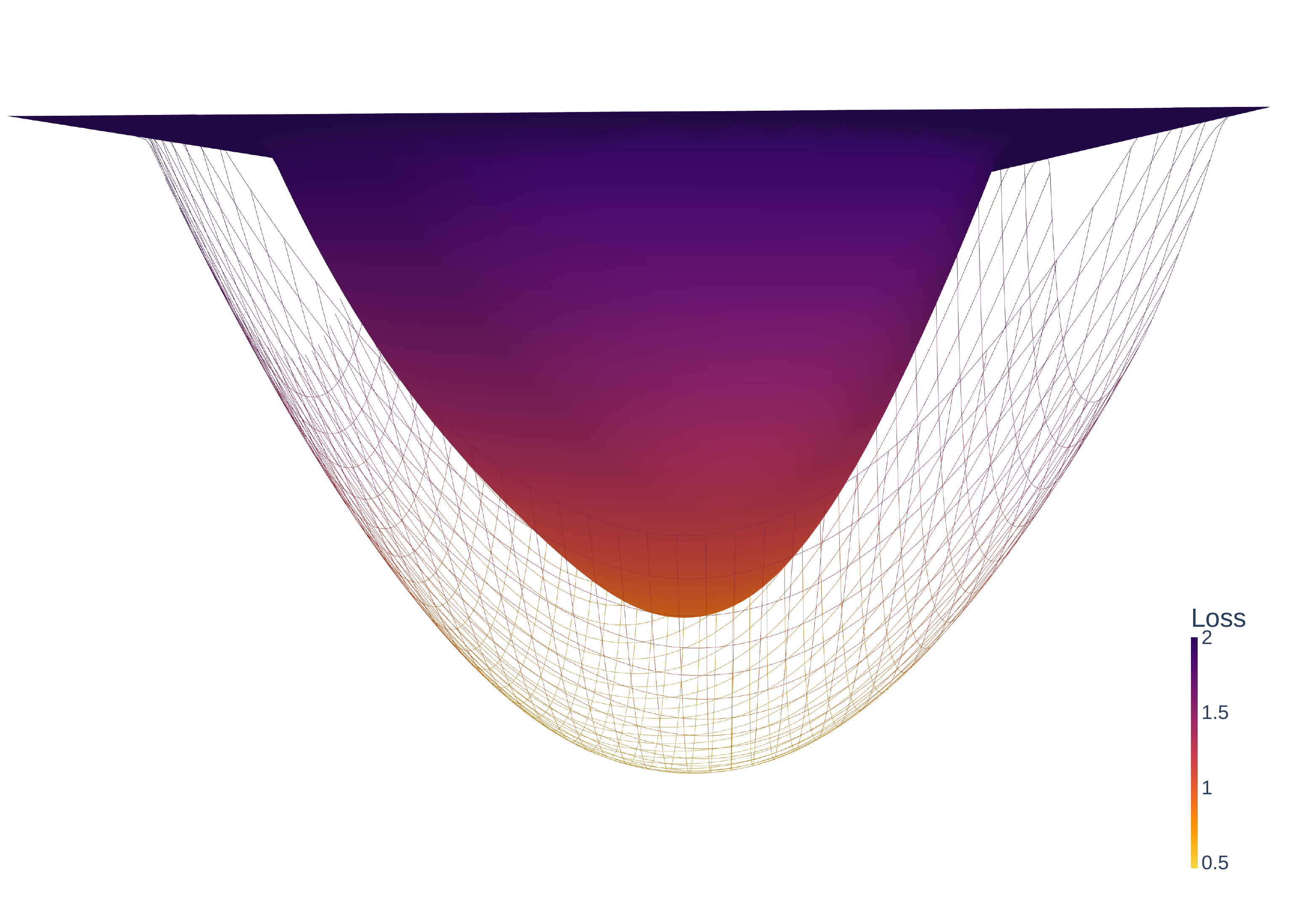}}
    \hfill
    \subfloat[][\cifarten $\alpha=0.05$ \\ \ours \vs \fedsmoo]{\includegraphics[width=.20\linewidth]{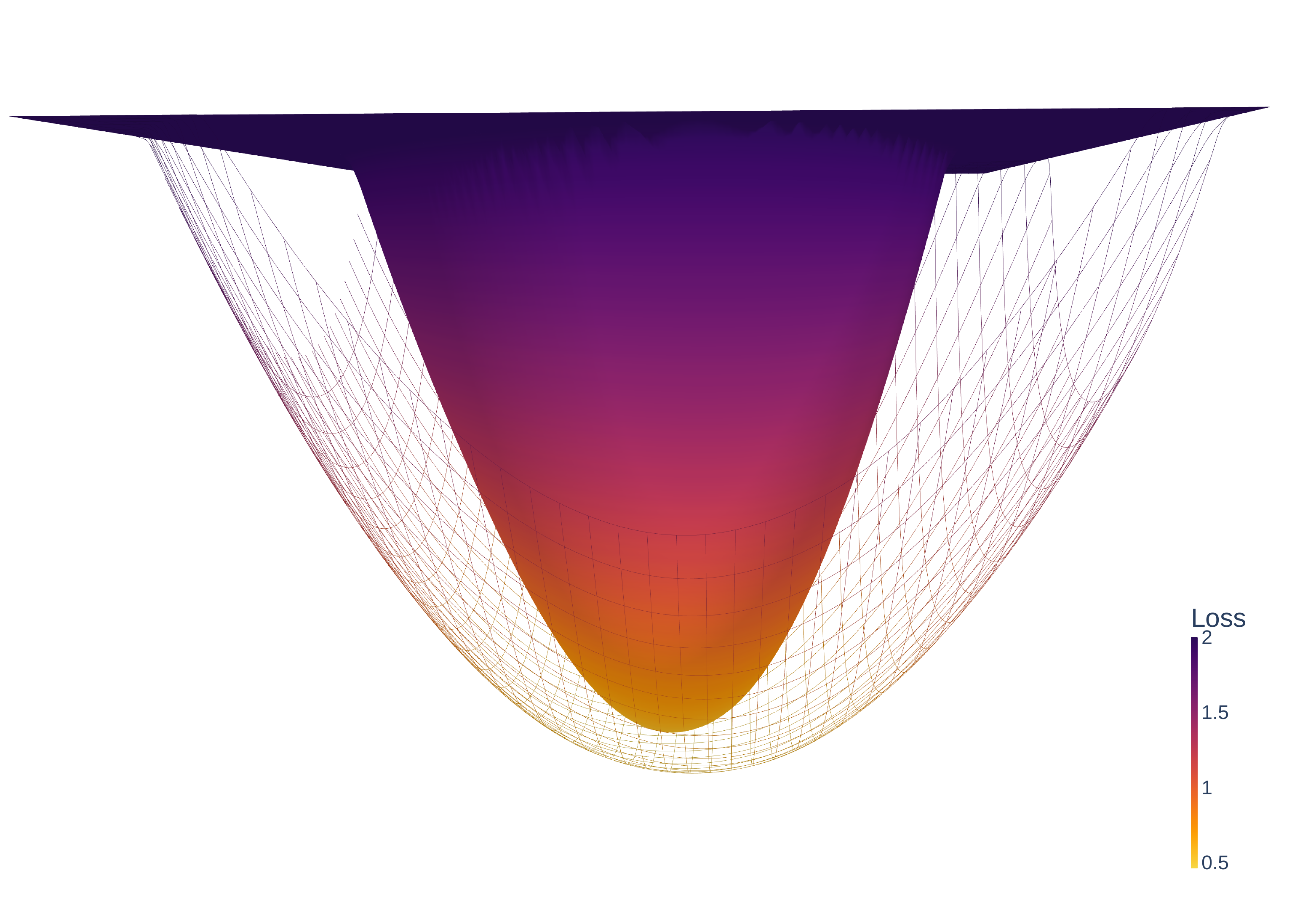}}
    \hfill
    \subfloat[][\cifar $\alpha=0.5$\\ \ours \vs \fedsam]{\includegraphics[width=.20\linewidth]{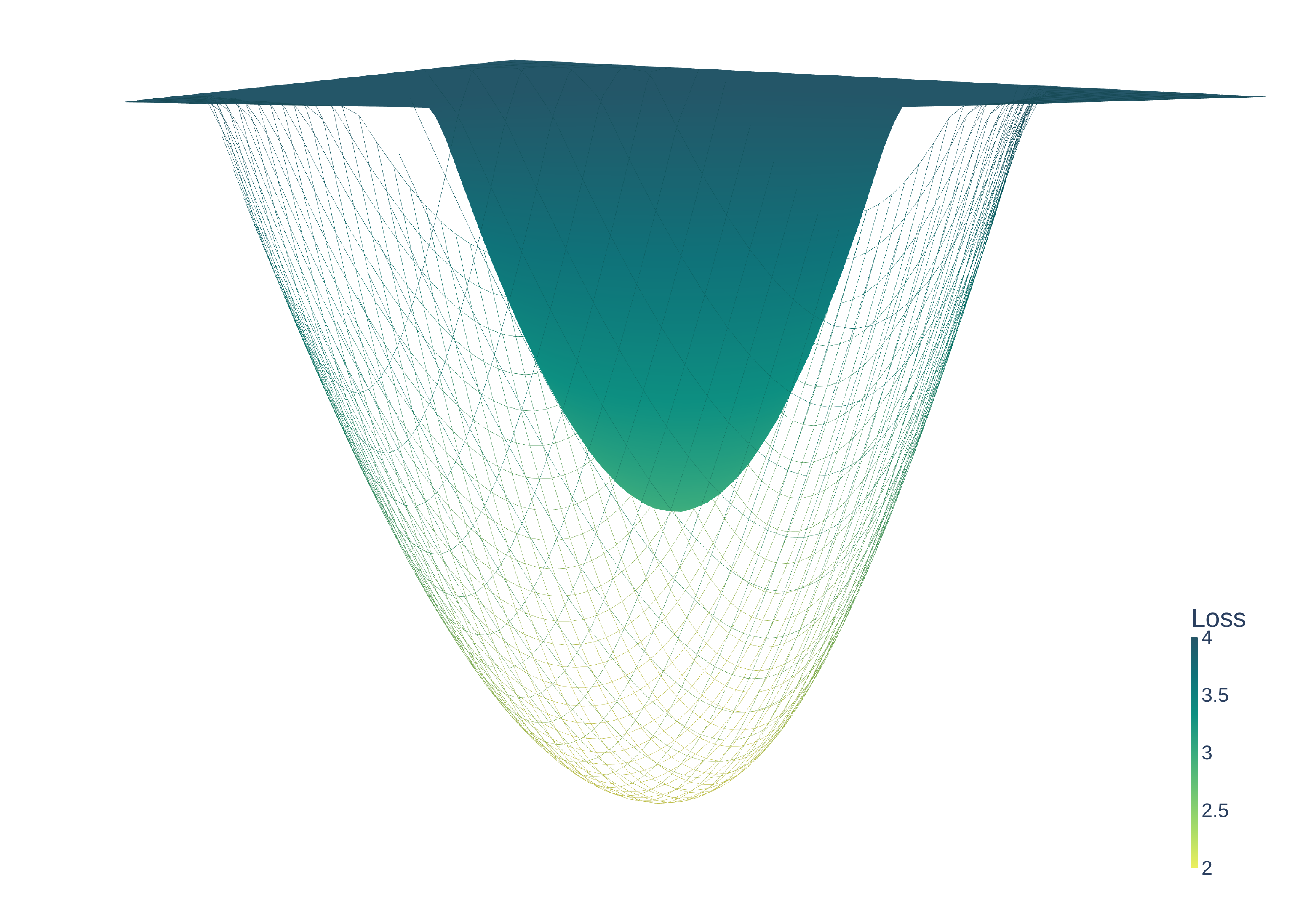}}
    \hfill
    \subfloat[][\cifar $\alpha=0.5$ \\ \ours \vs \fedsmoo]{\includegraphics[width=.20\linewidth]{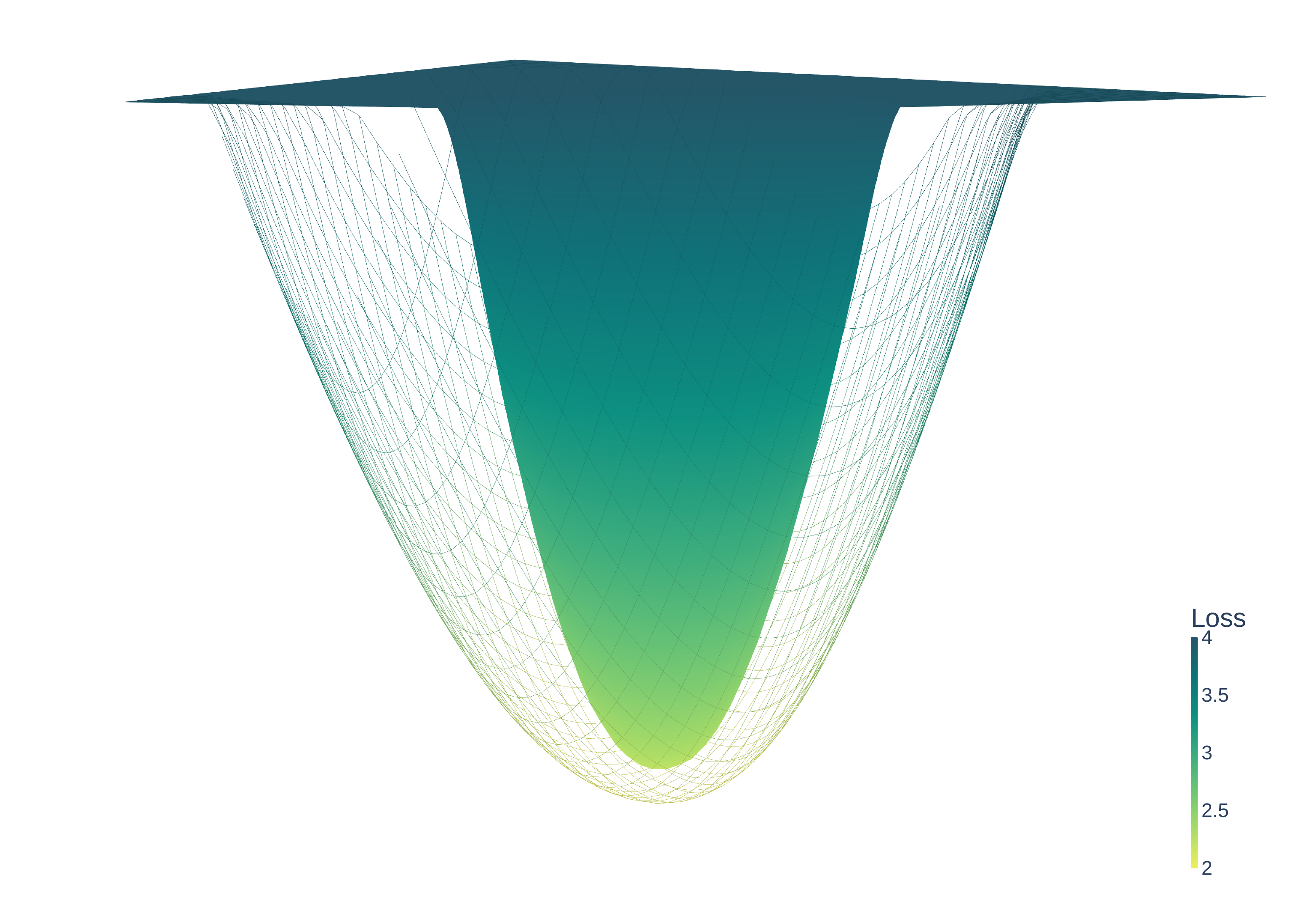}}
    \caption{{Visualization of the loss landscapes of the CNN trained with \textbf{\ours} (\textit{net})  \textbf{and \fedsam or \fedsmoo} (\textit{solid}). Comparison with $\alpha=0.05$ \cifarten (left) and $\alpha=0.5$ \cifar (right) extending \cref{fig:landscape_fedgloss_fedsam}. \textbf{\ours consistently achieves flatter minima and lower loss values in the \textit{global} loss landscape.}}}
    \label{fig:landscapes_fedgloss_fedsam_fedsmoo}
\end{figure}

\paragraph{\ours \vs \fedavg, \fedsam, \fedsmoo.} \cref{fig:landscapes_fedavg_fedgloss} compares the loss landscapes of global models trained with \ours and \fedavg, showing how the former consistently achieves flatter minima and lower loss values in the \textit{global} loss landscapes. This confirms the behaviors already appreciated in \cref{fig:landscape_fedgloss_fedsam}. In addition, 
\cref{fig:landscapes_fedgloss_fedsam_fedsmoo} shows the global loss surfaces of \ours' solutions against  models trained with \fedsam and \fedsmoo. These plots extend \cref{fig:landscape_fedgloss_fedsam} with the less heterogeneous scenarios $\alpha=0.05$ for \cifarten and $\alpha=0.5$ on \cifar. They confirm \ours' effectiveness in reaching flatter and lower-loss solutions with respect to its main direct competitors.

\begin{figure}[t]
    \centering
    \captionsetup{font=small}
    \captionsetup[sub]{font=scriptsize}
    \subfloat[ \fedavg (\textit{solid}) \vs \ours (\textit{net})][\cifarten $\alpha=0.05$\\\ours (\textit{net}) \vs\\\fedavg (\textit{solid})]{\includegraphics[width=.15\linewidth]{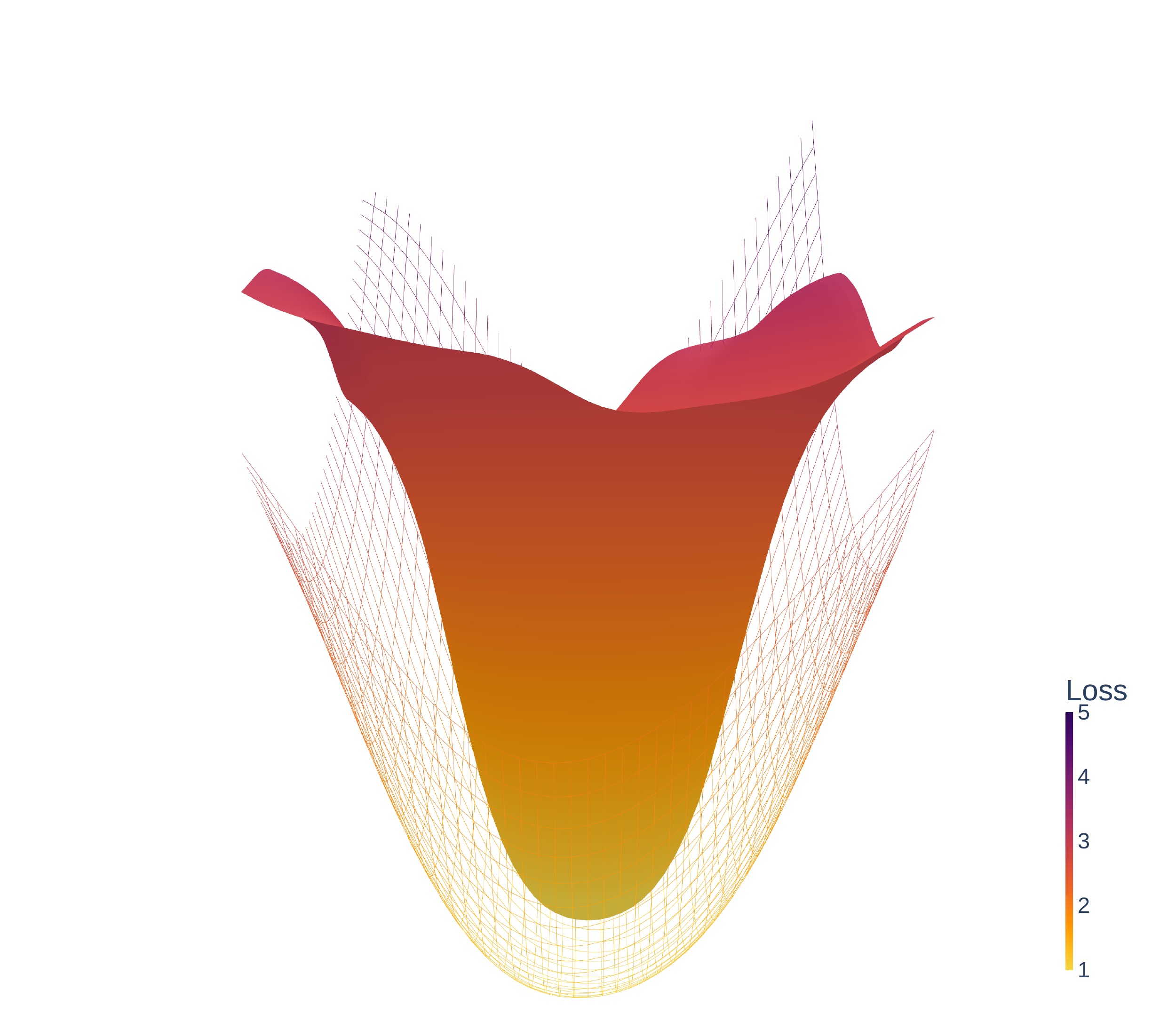}}
    \hfill
    \subfloat[\fedsam (\textit{solid}) \vs \ours (\textit{net})][\cifarten $\alpha=0.05$\\ \ours (\textit{net}) \vs\\\fedsam (\textit{solid})]{\includegraphics[width=.15\linewidth]{figures/landscapes/rn_c10_fedsam_fedgloss.pdf}}
    \hfill
    \subfloat[\fedsmoo (\textit{solid}) \vs \ours (\textit{net})][\cifarten $\alpha=0.05$\\ \ours (\textit{net}) \vs\\\fedsmoo (\textit{solid})]{\includegraphics[width=.15\linewidth]{figures/landscapes/rn_c10_fedsmoo_fedgloss.pdf}}
    \subfloat[\fedavg (\textit{solid}) \vs \ours (\textit{net})][\cifar $\alpha=0.5$\\ \ours (\textit{net}) \vs \fedavg (\textit{solid})]{\includegraphics[width=.15\linewidth]{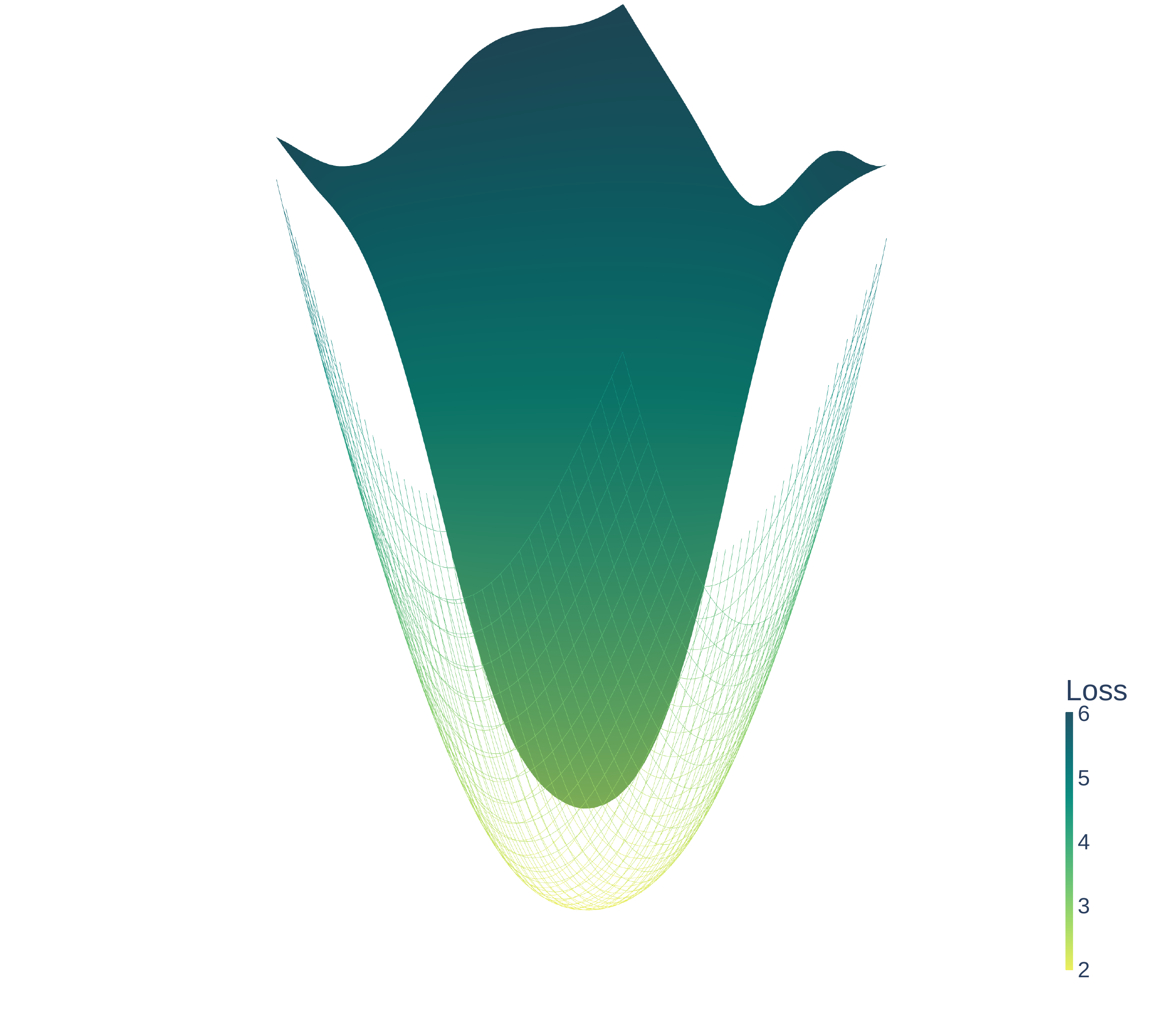}}
    \hfill
    \subfloat[\fedsam (\textit{solid}) \vs \ours (\textit{net})][\cifar$\alpha=0.5$\\ \ours (\textit{net}) \vs \fedsam (\textit{solid})]{\includegraphics[width=.15\linewidth]{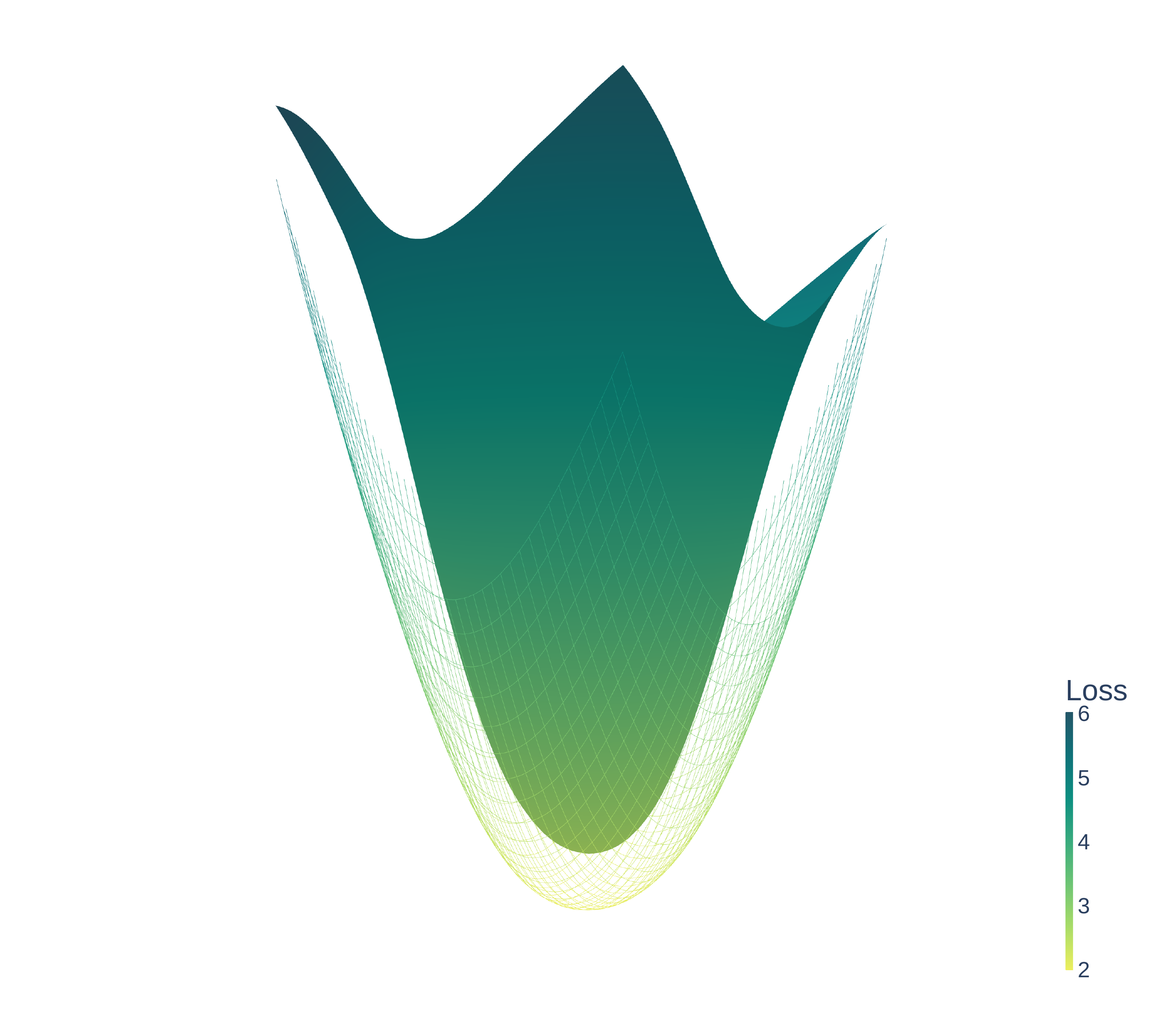}}
    \hfill
    \subfloat[\fedsmoo (\textit{solid}) \vs \ours (\textit{net})][\cifar $\alpha=0.5$ \\\ours (\textit{net}) \vs \fedsmoo (\textit{solid})]{\includegraphics[width=.15\linewidth]{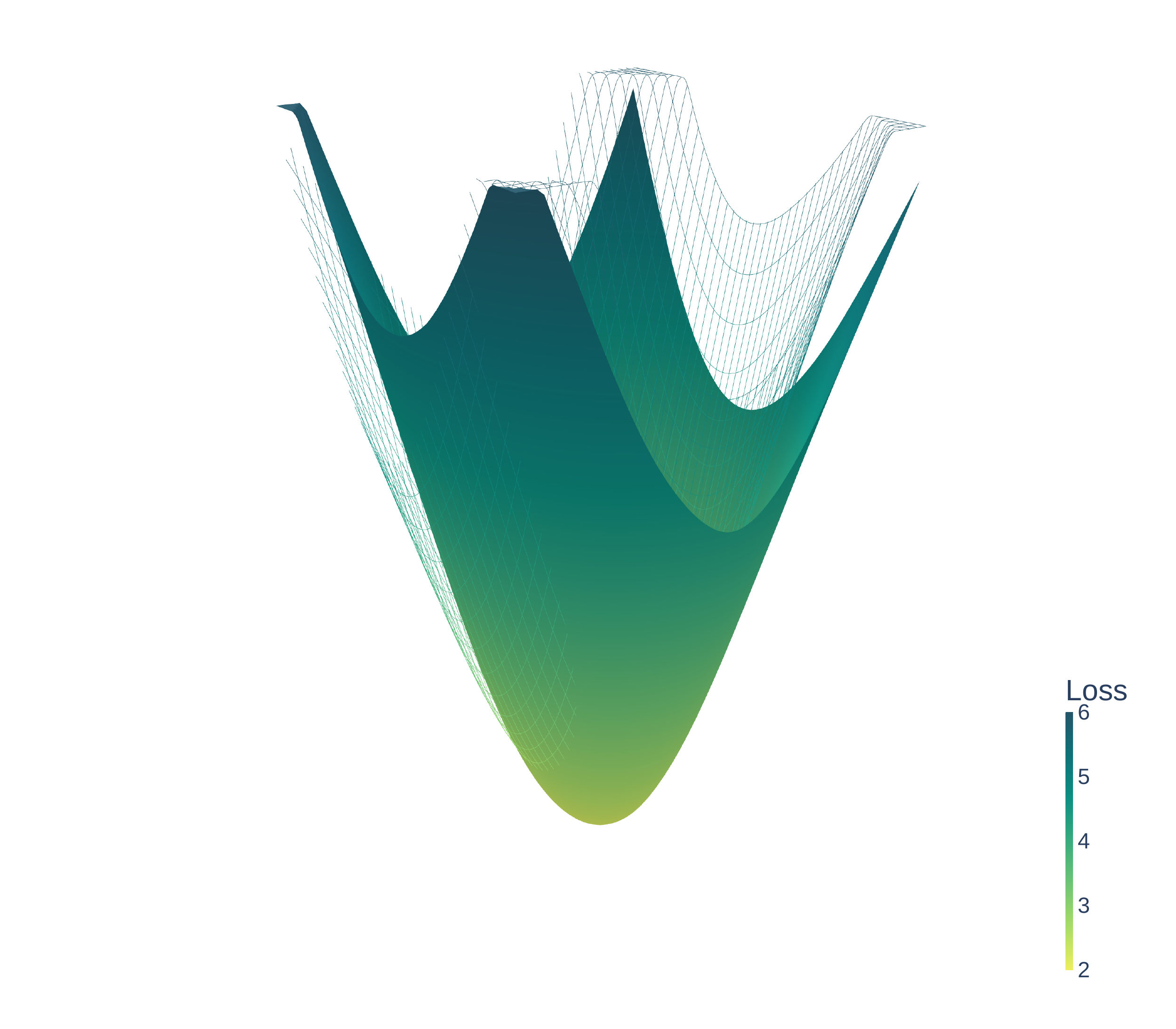}}\\
    \caption{Visualization of loss landscapes of the ResNet18 trained with \textbf{\ours} (\textit{net}) \textbf{and \fedavg, or \fedsam or \fedsmoo} (\textit{solid}). Comparison with  \cifarten $\alpha=0.05$ (\textit{a-c}) and \cifar $\alpha=0.5$ (\textit{d-f}). \textbf{\ours achieves flatter and lower-loss regions in the global landscape}.}
    \label{fig:landscape_c10}
\end{figure}

\paragraph{\ours on ResNet18.} \cref{fig:landscape_c10} extends \cref{fig:landscape_fedgloss_fedsam} from the main paper and shows the flatter loss landscapes reached by \ours when using ResNet18 on \cifarten and \cifar. 

\paragraph{Hessian Eigenvalues.}
\label{app:plots:eigs}

\cref{tab:sota_cnn_eig} reports the values of the maximum Hessian eigenvalues, as already shown in \cref{fig:eigs} in the main paper. 
First, as expected, we note that \sam-based methods achieve flatter minima \wrt the counterpart. Notably, our main competitor \fedsmoo presents higher sharpness than \fedsam in the simpler \cifarten, regardless of the data distribution. In addition, \ours with local \sam achieves the lowest sharpness (\ie, lowest $\lambda_1$) on \textit{all} configurations, outperforming the state of the art and, specifically, \textit{all} sharpness-aware methods. %
\setlength\tabcolsep{5pt}
\begin{table}[!h]
    \captionsetup{font=small}
    \caption{\textbf{\ours \vs the state of the art}, distinguished by local optimizer - \textcolor{NavyBlue}{\sgd} (\textit{top}) and \textcolor{Bittersweet}{\sam} (\textit{bottom}). Comparison in terms of communication cost and maximum Hessian eigenvalue $\lambda_1$. %
    Best results in \textbf{bold}. Model: \underline{CNN}.} %
    \centering    
    \scriptsize
    \begin{tabular}{clcccccccccccc}
        \toprule
         &\multicolumn{1}{c}{\multirow{3}{*}{\textbf{Method}}} & & \multicolumn{3}{c}{\textbf{\cifarten}} && \multicolumn{3}{c}{\textbf{\cifar}} \\ 
        \cline{4-6} \cline{8-10}
         && \textbf{Comm.} & {$\alpha=0$} && {$\alpha=0.05$} && {$\alpha=0$} && {$\alpha=0.5$} \\
        \cline{4-4} \cline{6-6} \cline{8-8} \cline{10-10}
         && \textbf{Cost} & $\lambda_{1} (\downarrow)$ && $\lambda_{1} (\downarrow)$ && $\lambda_{1} (\downarrow)$ && $\lambda_{1} (\downarrow)$ \\ 
        \midrule
        \multirow{5}{*}{\rotatebox[origin=c]{90}{\textcolor{NavyBlue}{Client \textbf{\sgd}}}} & \fedavg & \textcolor{ForestGreen}{\boldmath$1\times$} & $66.23\pm 0.50$ && $71.14\pm 4.07$ && \boldmath$66.30\pm 3.08$ && $68.77\pm 0.96$ \\
        & \fedprox & \textcolor{ForestGreen}{\boldmath$1\times$} & $66.19\pm 0.52$ && $71.41\pm 4.40$ && $66.34\pm 3.75$ && \boldmath$68.63\pm 1.37$\\
        &\feddyn & \textcolor{ForestGreen}{\boldmath$1\times$} & $63.94\pm 4.41$ && $71.44\pm 8.73$ && - && - \\
        &\scaffold & \textcolor{BrickRed}{\boldmath$2\times$} & $166.54\pm 6.93$ && $180.51\pm 30.08$ && - && $120.01\pm 0.76$\\
        \cdashline{2-14}[1pt/3pt] 
        &\textbf{\ours} & \textcolor{ForestGreen}{\boldmath$1\times$} & \boldmath$58.26\pm 3.49$ && \boldmath$56.28\pm 4.19$ && $96.01\pm 9.00$ && $107.35\pm 7.5$\\
        
        \midrule
 
        \multirow{6}{*}{\rotatebox[origin=c]{90}{\textcolor{Bittersweet}{Client \textbf{\sam}}}} &\fedsam & \textcolor{ForestGreen}{\boldmath$1\times$} & $10.35\pm 0.07$ && $9.43\pm 0.28$ && $58.38\pm 2.93$ && $57.54\pm 1.21$ \\
        &\feddyn & \textcolor{ForestGreen}{\boldmath$1\times$} & $10.04\pm 5.38$ && $6.58\pm 0.20$ && -&& -\\
        &\fedspeed & \textcolor{ForestGreen}{\boldmath$1\times$} & $10.92\pm 0.17$ && $9.97\pm 0.12$ && $58.23\pm 3.18$ && $58.00\pm 1.86$ \\
        &\fedgamma & \textcolor{BrickRed}{\boldmath$2\times$} & $4.79\pm 0.20$ && $4.55\pm 0.20$ && - && $99.86\pm 6.74$\\
        &\fedsmoo & \textcolor{BrickRed}{\boldmath$2\times$} & $15.37\pm 1.67$ && $12.57\pm 0.56$ && $28.43\pm 1.97$ && $29.23\pm 0.17$ \\
        \cdashline{2-14}[1pt/3pt]
        &\textbf{\ours} & \textcolor{ForestGreen}{\boldmath$1\times$} & \boldmath$2.03\pm 0.05$ && \boldmath$1.93\pm 0.03$ && \boldmath$17.18\pm 0.97$ && \boldmath${16.22}\pm 0.35$ \\
        \bottomrule
    \end{tabular}%
    \label{tab:sota_cnn_eig}
\end{table}

\subsection{Increasing Convergence Speed}
\label{app:plots:acc}

\begin{figure}[t]
   \centering
   \captionsetup{font=small}
   \captionsetup[sub]{font=small}
    \subfloat[][$\alpha=0$ \sam]{\includegraphics[width=.25\linewidth]{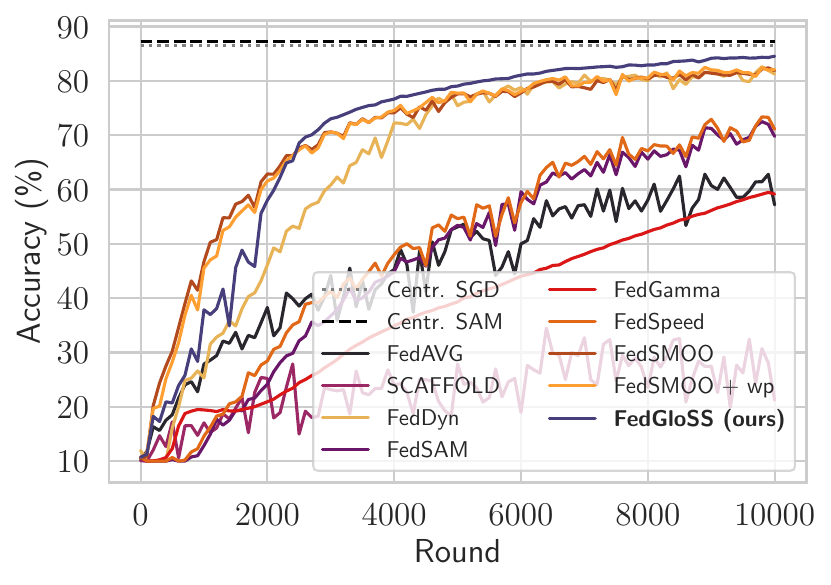}}
    \hfill
    \subfloat[][$\alpha=0$ \sgd]{\includegraphics[width=.25\linewidth]{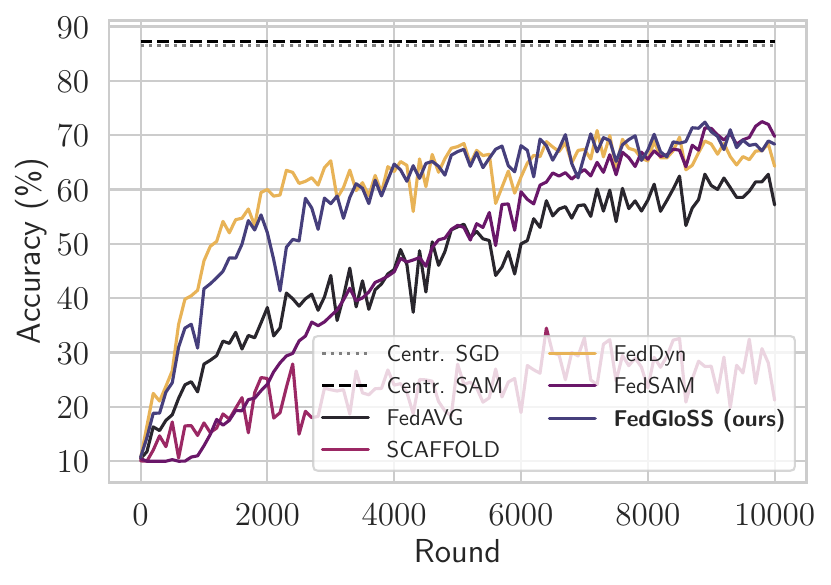}}
    \hfill
    \subfloat[][$\alpha=0.05$ \sam]{\includegraphics[width=.25\linewidth]{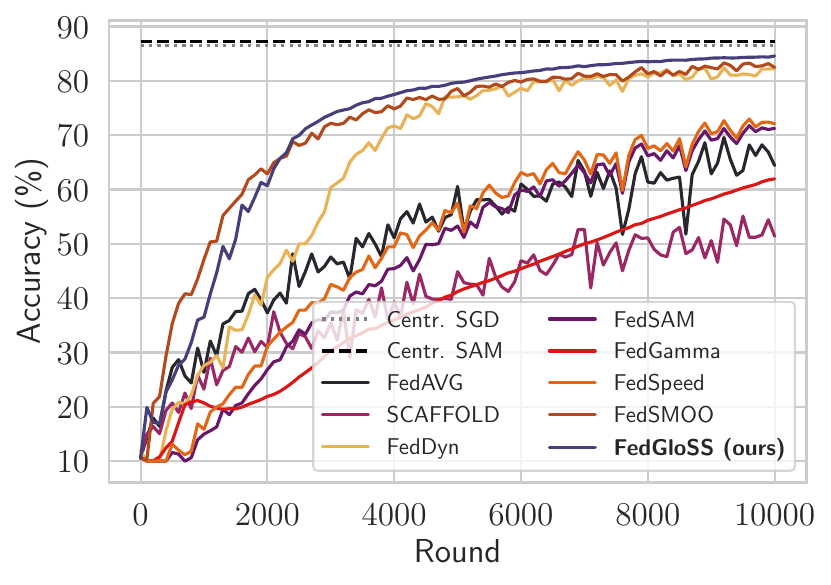}}
    \hfill
    \subfloat[][$\alpha=0.05$ \sgd]{\includegraphics[width=.25\linewidth]{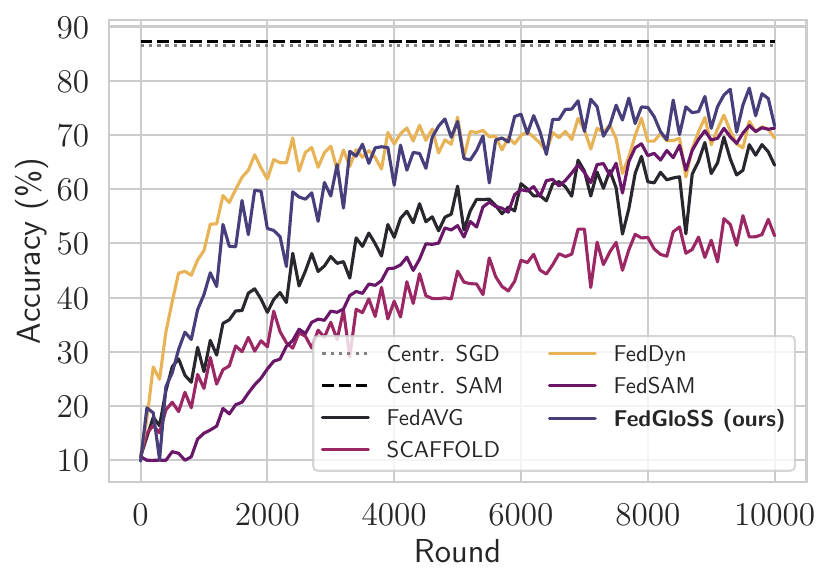}}
    \caption{\cifarten with varying degrees of heterogeneity ($\alpha\in\{0,0.05\}$). Results of centralized runs (\textit{dashed lines}) added as reference. Comparison of \ours with state-of-the-art approaches, distinguished in \sam-based methods (\textbf{a}, \textbf{c}) and \sgd-based ones (\textbf{b}, \textbf{d}). \ours consistently achieves the best performance. Model: CNN.}
   \label{fig:acc_c10}
\end{figure}

\begin{figure}[t]
   \centering
   \captionsetup{font=small}
   \captionsetup[sub]{font=small}
    \subfloat[][$\alpha=0$ \sam]{\includegraphics[width=.25\linewidth]{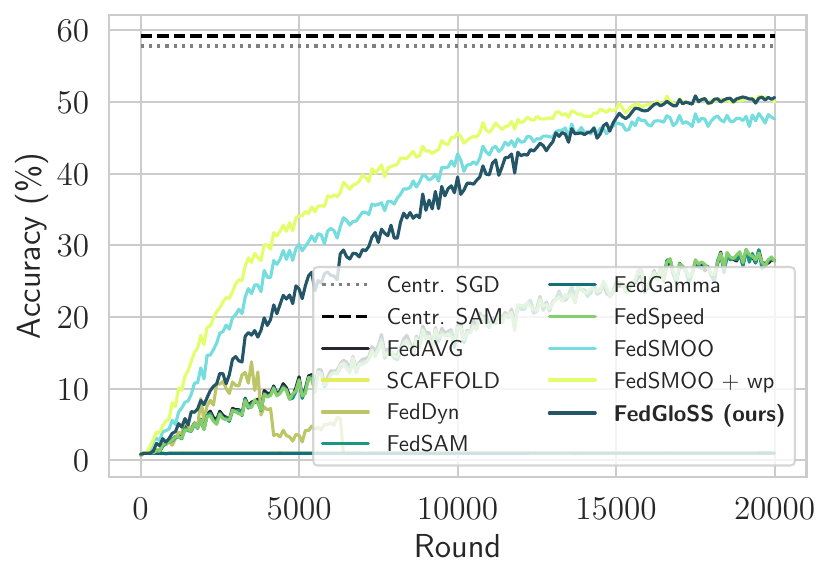}}
    \hfill
    \subfloat[][$\alpha=0$ \sgd]{\includegraphics[width=.25\linewidth]{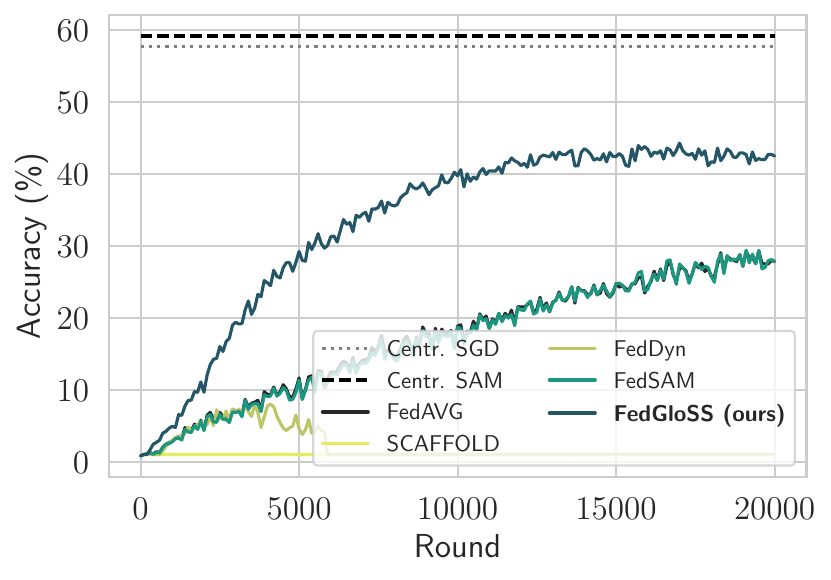}}
    \hfill
    \subfloat[][$\alpha=0.5$ \sam]{\includegraphics[width=.25\linewidth]{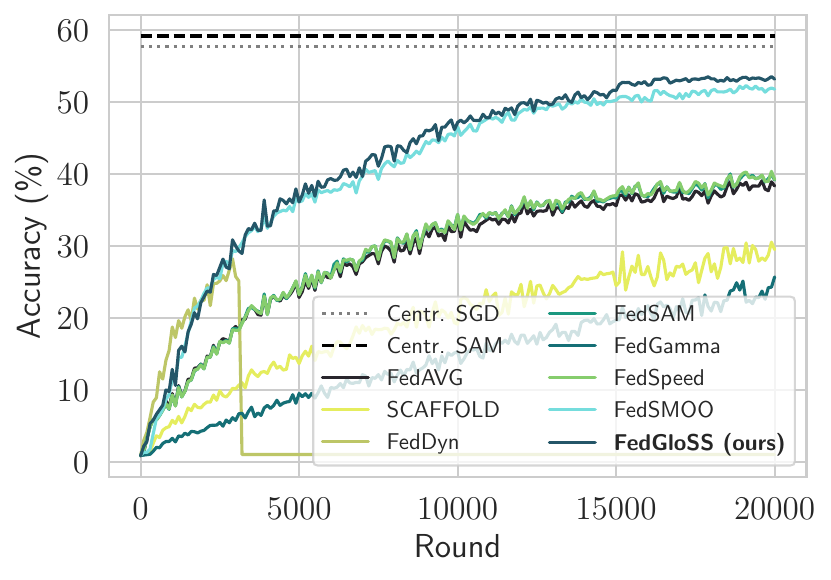}}
    \hfill
    \subfloat[][$\alpha=0.5$ \sgd]{\includegraphics[width=.25\linewidth]{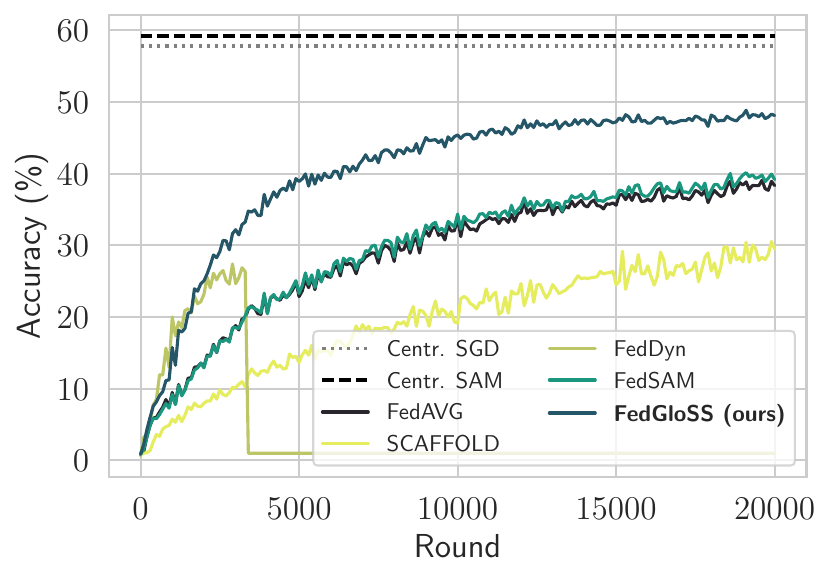}}
    \caption{\cifar with varying degrees of heterogeneity ($\alpha\in\{0,0.5\}$) with \textbf{CNN}. Results of centralized runs (\textit{dashed lines}) added as reference. Comparison of \ours with state-of-the-art approaches, distinguished in \sam-based methods (\textbf{a}, \textbf{c}) and \sgd-based ones (\textbf{b}, \textbf{d}). \ours consistently achieves the best performance.}
   \label{fig:acc_c100}
\end{figure}

\begin{figure}[t]
   \centering
   \captionsetup{font=small}
   \captionsetup[sub]{font=small}
    \subfloat[][\cifar $\alpha=0.5$ \sam]{\includegraphics[width=.25\linewidth]{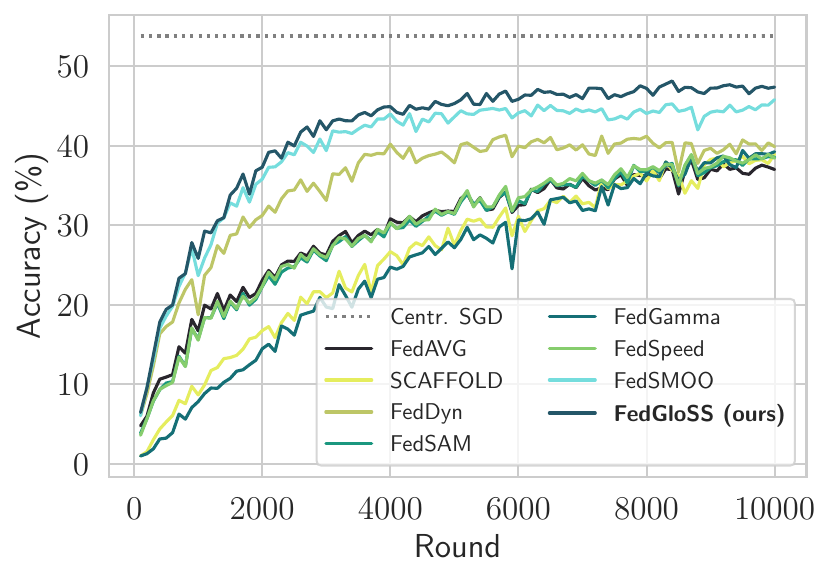}}
    \hfill
    \subfloat[][\cifar $\alpha=0.5$ \sgd]{\includegraphics[width=.25\linewidth]{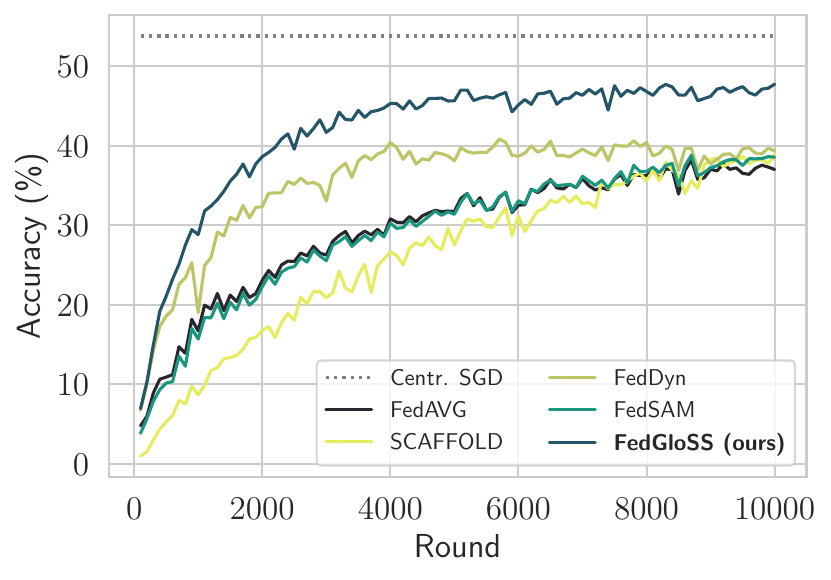}}
    \hfill
    \subfloat[][\cifarten $\alpha=0.05$ \sam]{\includegraphics[width=.25\linewidth]{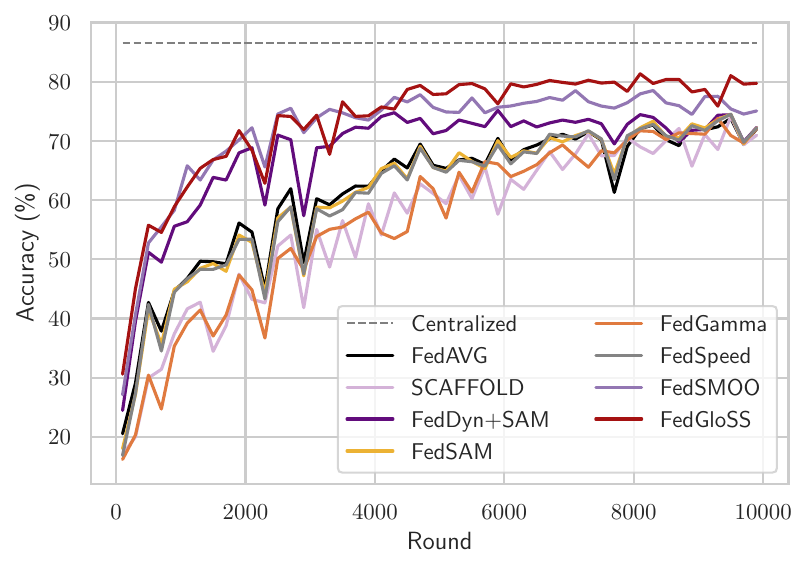}}
    \hfill
    \subfloat[][\cifarten $\alpha=0.05$ \sgd]{\includegraphics[width=.25\linewidth]{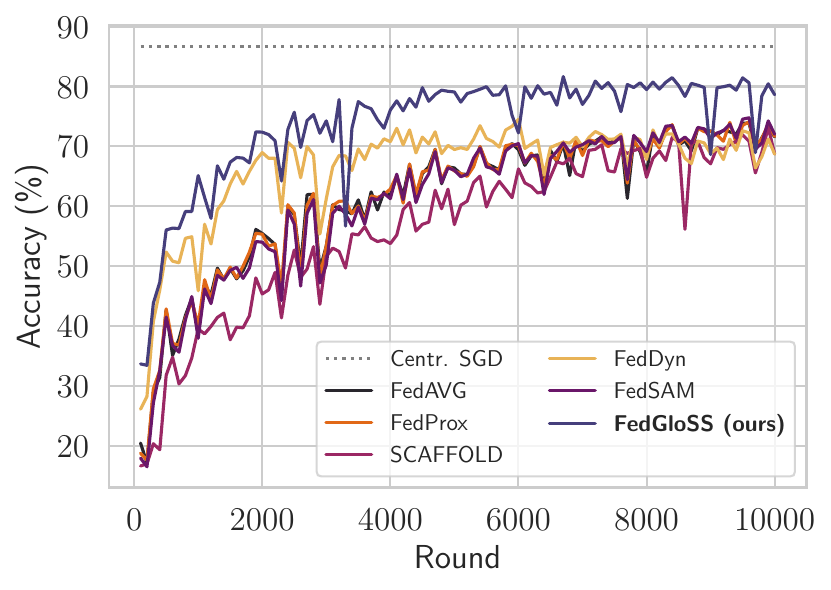}}
    \caption{Accuracy trends with \textbf{ResNet18} on \cifar (\textit{left}) and \cifarten (\textit{right}). Comparison of \ours with state-of-the-art approaches, distinguished in \sam-based methods (\textbf{a}, \textbf{c}) and \sgd-based ones (\textbf{b}, \textbf{d}). \ours consistently achieves the best performance, both in terms of final accuracy and convergence speed.}
   \label{fig:acc_c100_rn18}
\end{figure}

\cref{fig:acc_c10,fig:acc_c100} show the accuracy trends of \ours compared to state-of-the-art methods for heterogeneous FL on \cifarten and \cifar respectively. For a clearer understanding, we distinguish between \sam-based and \sgd-based methods depending on the used local optimizer. For a fair comparison, we report \fedsmoo with and without the scheduling of $\rho$ (\textit{+wp} in the figure). For additional details on the scheduling, refer to \cref{app:exp_details}. \ours consistently achieves the best performances and convergence speedup in each group. In addition, we remind that \ours communicates \textit{half} the number of bits at each round \wrt \fedsmoo. \cref{fig:acc_c100_rn18} reports the results obtained with ResNet18 on both datasets: \ours consistently achieves the best speed of convergence and final accuracy, with both \sam and \sgd as \textsc{ClientOpt}.

\subsection{Effectiveness in Multiple Scenarios and with Several Model Architectures}
\label{app:settings:hom}
As already shown in \cref{sec:exp}, \ours outperforms the state of the art in terms of speed of convergence, final performance in accuracy, flatness of the solution and communication efficiency. This remains true across multiple datasets (\cifar, \cifarten, \gld) and model architectures (CNN, ResNet18, MobileNetv2). The main paper focuses on results in more challenging heterogeneous FL scenarios. The next paragraphs discuss the behavior of \ours in increasingly more homogeneous settings.

\paragraph{\ours in moderately heterogeneous settings.}
The choice of the Dirichlet's concentration parameter $\alpha$ in the main text aligns with the FL literature \citep{acar2021federated,caldarola2022improving,hsu2020federated,sun2023dynamic,sun2023fedspeed}, where $\alpha < 1$ is commonly used to evaluate challenging settings. Moreover, for $\alpha > 0.6$, differences between methods become less pronounced \citep{zaccone2023communication}. To further analyze the impact of varying heterogeneity levels on \ours, we consider $\alpha \in \{1, 5, 10\}$ on \cifarten. \cref{tab:other-alpha} presents a comparison of \ours against several state-of-the-art FL algorithms. Consistently, \ours achieves the highest accuracy across all tested scenarios, except for $\alpha = 5$ with SGD as the local optimizer, where it is outperformed by \scaffold. Notably, even in this case, \ours exhibits the lowest maximum Hessian eigenvalue, highlighting its stability.

\setlength\tabcolsep{1pt}
\begin{table}[t]
\captionsetup{font=small}
\caption{\ours against SOTA algorithms in increasingly more homogeneous settings on \cifarten, compared in terms of accuracy (\%) and maximum Hessian eigenvalue $\lambda_1$. Best results in \textbf{bold}. Model: {CNN}.}
    \centering
    \footnotesize
    \begin{tabular}{clccccccccc}
    \toprule
        & \multirow{2}{*}{\textbf{Method}} && \multicolumn{2}{c}{$\alpha=1$} && \multicolumn{2}{c}{$\alpha=5$} && \multicolumn{2}{c}{$\alpha=10$} \\
    \cline{4-5} \cline{7-8} \cline{10-11}
        &&& Accuracy & $\lambda_1$ && Accuracy & $\lambda_1$ && Accuracy & $\lambda_1$ \\
    \midrule
        \multirow{5}{*}{\rotatebox[origin=c]{90}{\textcolor{NavyBlue}{\textbf{\textsc{Sgd}}-based}}} & \fedavg && $76.4 \pm 0.2$ & $68.2 \pm 2.7$ && $83.7 \pm 0.4$ & $63.3 \pm 2.9$ && $82.4 \pm 0.1$ & $70.6 \pm 0.3$\\
        & \fedprox && $76.3 \pm 0.2$ & $69.5 \pm 0.9$ && $83.8 \pm 0.4$ & $63.0 \pm 3.6$ && $82.4 \pm 0.2$ & $71.8 \pm 2.0$\\
        & \feddyn && $77.1 \pm 0.1$ & $64.5 \pm 4.9$ && $82.3 \pm 0.3$ & $48.3 \pm 1.6$ && $82.2 \pm 0.1$ & $58.1 \pm 3.5$\\
        & \scaffold  && $79.6 \pm 0.1$ & $62.7 \pm 2.7$ && \boldmath$84.5 \pm 0.2$ & $47.4 \pm 3.0$ && $83.8 \pm 0.2$ & $58.4 \pm 2.9$\\
    \cdashline{2-11}[1pt/3pt]
        & \textbf{\ours} && \boldmath$80.31 \pm 0.4$ & \boldmath$40.0 \pm 5.2$ && $83.5 \pm 0.2$ & \boldmath$8.5 \pm 0.3$ && \boldmath$84.8 \pm 0.3$ & \boldmath$41.4 \pm 2.2$\\
    \midrule
        \multirow{6}{*}{\rotatebox[origin=c]{90}{\textcolor{Bittersweet}{\textbf{\textsc{Sam}}-based}}} & \fedsam && $77.7 \pm 0.2$ & $25.6 \pm 0.02$ && $83.5 \pm 0.1$ & $19.7 \pm 0.1$ && $83.2 \pm 0.2$ & $22.1 \pm 1.4$\\
        & \feddyn && $80.5 \pm 0.2$ & $21.0 \pm 0.8$ && $82.9 \pm 0.4$ & $16.3 \pm 0.5$ && $84.1 \pm 0.1$ & $17.3 \pm 0.3$\\
        & \fedspeed && $76.2 \pm 0.5$ & $8.3 \pm 0.1$ && $83.6 \pm 0.1$ & $20.3 \pm 0.5$ && $78.2 \pm 0.2$ & $8.6 \pm 0.2$\\
        & \fedgamma && $70.3 \pm 1.0$ & $4.8 \pm 0.2$ && $83.5 \pm 0.1$ & $16.0 \pm 0.6$ && $76.4 \pm 0.3$ & $6.4 \pm 0.1$\\
        & \fedsmoo && $85.2 \pm 0.2$ & $9.8 \pm 0.2$ && $86.2 \pm 0.2$ & $6.0 \pm 0.2$ && $86.8 \pm 0.3$ & $6.2 \pm 0.2$\\
    \cdashline{2-11}[1pt/3pt]
        & \textbf{\ours} && \boldmath$85.6 \pm 0.2$ & \boldmath$2.0 \pm 0.04$ && \boldmath$86.9 \pm 0.2$ & \boldmath$4.5 \pm 0.02$ && \boldmath$87.1 \pm 0.1$ & \boldmath$4.1 \pm 0.03$\\
    \bottomrule
    \end{tabular}
    \label{tab:other-alpha}
\end{table}

\paragraph{\ours in homogeneous settings.}
\cref{tab:hom} evaluates \ours  against the main methods \fedavg, \fedsam, \feddyn and \fedsmoo in homogeneous FL settings. Here, client gradients are naturally more aligned due to reduced client drift \citep{karimireddy2020scaffold}. We thus test \ours with and without ADMM for global consistency. As expected, \ours achieves similar accuracy with or without ADMM, particularly when using SAM as the local optimizer. However, ADMM facilitates convergence to flatter minima (evidenced by lower $\lambda_1$ values) by aligning local and global convergence points. Notably, \ours achieves the flattest minima (lowest $\lambda_1$) across both datasets, and the best accuracy on the more complex \cifar. While \fedsmoo achieves slightly higher accuracy on \cifarten, \ours finds a flatter minimum and achieves competitive accuracy with significantly lower communication costs (halved).

\begin{table}[h]
\setlength\tabcolsep{4pt}
    \centering
    \captionsetup{font=small}
    \caption{\ours against SOTA FL methods on homogeneous \textsc{Cifar} settings, compared in terms of communication costs, accuracy (\%) and maximum Hessian eigenvalue $\lambda_1$.  Best result in \textbf{bold} and second best \underline{underlined}. Model: {CNN}.}
    \footnotesize
    \begin{tabular}{lcccccccc}
    \toprule
         \multirow{2}{*}{\textbf{Method}} & \textbf{Comm.} & \multirow{2}{*}{\textbf{ADMM}} & \multicolumn{2}{c}{\textbf{\cifarten} \boldmath$\alpha=100$} && \multicolumn{2}{c}{\textbf{\cifar} \boldmath$\alpha=1000$} \\
         \cline{4-5} \cline{7-8}
         & \textbf{Cost} && Accuracy & $\lambda_1$ && Accuracy & $\lambda_1$\\
         \midrule
         \fedavg & \textcolor{ForestGreen}{\boldmath$1\times$} & \xmark & 84.0 & 68.4 && 50.1 & 49.4 \\
         \fedsam & \textcolor{ForestGreen}{\boldmath$1\times$} & \xmark & 84.7 & 36.2 && 53.4 & 32.6 \\
         \feddyn (\sgd) & \textcolor{ForestGreen}{\boldmath$1\times$} & \cmark & 83.8 & 47.8 && 51.9 & 91.7 \\
         \feddyn (\sam) & \textcolor{ForestGreen}{\boldmath$1\times$} &  \cmark & 84.5 & 27.9 && 52.5 & 46.0 \\
         \fedsmoo & \textcolor{BrickRed}{\boldmath$2\times$} & \cmark & \textbf{85.1} & \underline{6.4} && 53.9 & 24.6 \\
         \midrule
         \multirow{2}{*}{\ours (\sgd)} &  \textcolor{ForestGreen}{\boldmath$1\times$} & \xmark & 84.0 & 67.7 && 50.5 & 50.8 \\
          &  \textcolor{ForestGreen}{\boldmath$1\times$} & \cmark & 83.1 & 7.1 && 51.7 & 47.9 \\
         \hdashline
         \multirow{2}{*}{\ours (\sam)} &  \textcolor{ForestGreen}{\boldmath$1\times$} & \xmark & \underline{84.8} & 36.2 && \textbf{55.8} & \underline{13.9} \\
          &  \textcolor{ForestGreen}{\boldmath$1\times$} & \cmark & \underline{84.8} & \textbf{2.8} && \underline{55.7} & \textbf{11.8} \\
    \bottomrule
    \end{tabular}
    \label{tab:hom}
\end{table}

\subsection{Reducing Communication Costs}
\label{app:others:comm_eff}

\paragraph{Analysis on communication cost.} \cref{tab:comm_eff_others} extends the analysis presented in \cref{subsec:comm} by evaluating the communication costs across all scenarios considered in this work. Notably, since \ours maintains the per-round communication cost of \fedavg, it remains advantageous even when matching the convergence speed of the best-performing algorithm (\fedsmoo), due to its \emph{halved communication costs}.

\setlength\tabcolsep{1pt}
\begin{table}[h]
\captionsetup{font=small}
\caption{Communication costs comparison %
\wrt \textsc{FedAvg}. ``-'' for not reached accuracy, \quotes{\xmark} for non-convergence.}
    \centering
    \tiny
    \resizebox{\linewidth}{!}{
    \begin{tabular}{clccccccccccccccccccccc}
    \toprule
        & \multirow{4}{*}{\textbf{Method}} && \multicolumn{11}{c}{\textbf{CNN}} && \multicolumn{5}{c}{\textbf{ResNet18}} && \multicolumn{2}{c}{\textbf{MobileNetv2}} \\
    \cline{4-14} \cline{16-20} \cline{22-23}
        &&& \multicolumn{5}{c}{\textsc{Cifar-10}} && \multicolumn{5}{c}{\textsc{Cifar-100}} && \multicolumn{2}{c}{\textsc{Cifar-10}} && \multicolumn{2}{c}{\textsc{Cifar-100}} && \multicolumn{2}{c}{\multirow{2}{*}{\textsc{Landmarks-User-160k}}} \\
    \cline{4-8} \cline{10-14} \cline{16-17} \cline{19-20}
        &&& \multicolumn{2}{c}{$\alpha=0$} && \multicolumn{2}{c}{$\alpha=0.05$} && \multicolumn{2}{c}{$\alpha=0$} && \multicolumn{2}{c}{$\alpha=0.5$} && \multicolumn{2}{c}{$\alpha=0.05$} && \multicolumn{2}{c}{$\alpha=0.5$}\\
    \cline{4-5} \cline{7-8} \cline{10-11} \cline{13-14} \cline{16-17} \cline{19-20} \cline{22-23}
        &&& \textit{Rounds} & \textit{Cost} && \textit{Rounds} & \textit{Cost} && \textit{Rounds} & \textit{Cost} && \textit{Rounds} & \textit{Cost} && \textit{Rounds} & \textit{Cost} && \textit{Rounds} & \textit{Cost} && \textit{Rounds} & \textit{Cost} \\
    \midrule
        \multirow{5}{*}{\rotatebox[origin=c]{90}{\textcolor{NavyBlue}{\textbf{\textsc{Sgd}}-based}}} & \textsc{FedAvg} && $10k \ (1 \times)$ & $1B$ && $10k \ (1 \times)$ & $1B$ && $20k \ (1 \times)$ & $1B$ && $20k \ (1 \times)$ & $1B$ && $10k \ (1 \times)$ & $1B$ && $10k \ (1 \times)$ & $1B$ && $1.3k \ (1 \times)$ & $1B$ \\
        & \textsc{FedProx} && $7.6k \ (1.3 \times)$ & $0.8B$ && $7.9k \ (1.3 \times)$ & $0.8B$ && $18.7k \ (1.1 \times)$ & $0.9B$ && $18.6k \ (1.1 \times)$ & $0.9B$ && $8.8k \ (1.1 \times)$ & $0.9B$ && $8.3k (1.2 \times)$ & $0.8B$ && - & - \\
        & \textsc{FedDyn} && $2k \ (5 \times)$ & \boldmath$0.2 B$ && $1.9k \ (5 \times)$ & \boldmath$0.2B$ && \multicolumn{2}{c}{\xmark} && \multicolumn{2}{c}{\xmark} && - & - && $3.5k \ ( 2.9 \times)$ & $0.4B$ && - & - \\
        & \textsc{Scaffold} && - & - && - & - && - & - && - & - && - & - && $8.9k \ (1.1 \times)$ & $1.8B$ && \multicolumn{2}{c}{\xmark} \\
    \cdashline{2-23}[1pt/3pt] 
        & \textbf{\textsc{FedGloSS}} && $3.4k \ (2.9 \times)$ & $0.3B$ && $3.8k \ (2.6 \times)$ & $0.4B$ && $5k \ (4 \times)$ & \boldmath$0.3B$ && $4.7k \ (4.3 \times)$ & \boldmath$0.2B$ && $2.4k \ (4.2 \times)$ & \boldmath$0.2B$ && $1.9k \ (5.3 \times)$ & \boldmath$0.2B$ && - & - \\
    \midrule
        \multirow{6}{*}{\rotatebox[origin=c]{90}{\textcolor{Bittersweet}{\textbf{\textsc{Sam}}-based}}} & \textsc{FedSam} && $6.3k \ (1.6 \times)$ & $0.6B$ && $7.8k \ (1.3 \times)$ & $0.8B$ && $18.3k \ (1.1 \times)$ & $0.9B$ && $16.3k \ (1.2 \times)$ & $0.8B$ && $9.2k \ (1.1 \times)$ & $0.9B$ && $7.8k \ (1.3 \times)$ & $0.8B$ && $1.3k \ (1 \times)$ & $1B$ \\
        & \textsc{FedDyn} && $3k \ (3.3 \times)$ & $0.3B$ && $4.2k \ (2.4 \times)$ & $0.4B$ && \multicolumn{2}{c}{\xmark} && \multicolumn{2}{c}{\xmark} && $4.1k \ (2.4 \times)$ & $0.4B$ && $3.5k \ (2.9 \times)$ & $0.4B$ && - & - \\
        & \textsc{FedSpeed} && $6.3k \ (1.6 \times)$ & $0.6B$ && $6.9k \ (1.4 \times)$ & $0.7B$ && $18.3k \ (1.1 \times)$ & $0.9B$ && $15.7k \ (1.3 \times)$ & $$0.8B$$ && $8.3k \ (1.2 \times)$ & $0.8B$ && $8.3k \ (1.2 \times)$ & $0.8B$ && $1.3k \ (1 \times)$ & $1B$ \\
        & \textsc{FedGamma} && - & - && - & - && - & - && - & - && $9.3k \ (1.1 \times)$ & $1.9B$ && $8.1k \ (1.2 \times)$ & $1.6B$ && \multicolumn{2}{c}{\xmark} \\
        & \textsc{FedSmoo} && $1.9k \ (5.3 \times)$ & $0.4B$ && $2.2k \ (4.5 \times)$ & $0.4B$ && $4.5k \ (4.4 \times)$ & $0.5B$ && $6.5k \ (3.1 \times)$ & $0.7B$ && $2.4k \ (4.2 \times)$ & $0.5B$ && $2.3k \ (4.3 \times)$ & $0.5B$ && $200 \ (6.5 \times)$ & $0.3B$ \\
    \cdashline{2-23}[1pt/3pt] 
        & \textbf{\textsc{FedGloSS}} && $2.2k \ (4.5 \times)$ & \boldmath$0.2B$ && $2.2k \ (4.5 \times)$ & \boldmath$0.2B$ && $6.3k \ (3.2 \times)$ & \boldmath$0.3B$ && $5.2k \ (3.8 \times)$ & \boldmath$0.3B$ && $2.4k \ (4.2 \times)$ & \boldmath$0.2B$ && $1.9k \ (5.3 \times)$ & \boldmath$0.2B$ && $200 \ (6.5 \times)$ & \boldmath$0.2B$ \\
    \bottomrule
    \end{tabular}}
    \label{tab:comm_eff_others}
\end{table}

\begin{figure}[t]
    \centering
    \captionsetup{font=small}
    \includegraphics[width=0.5\linewidth]{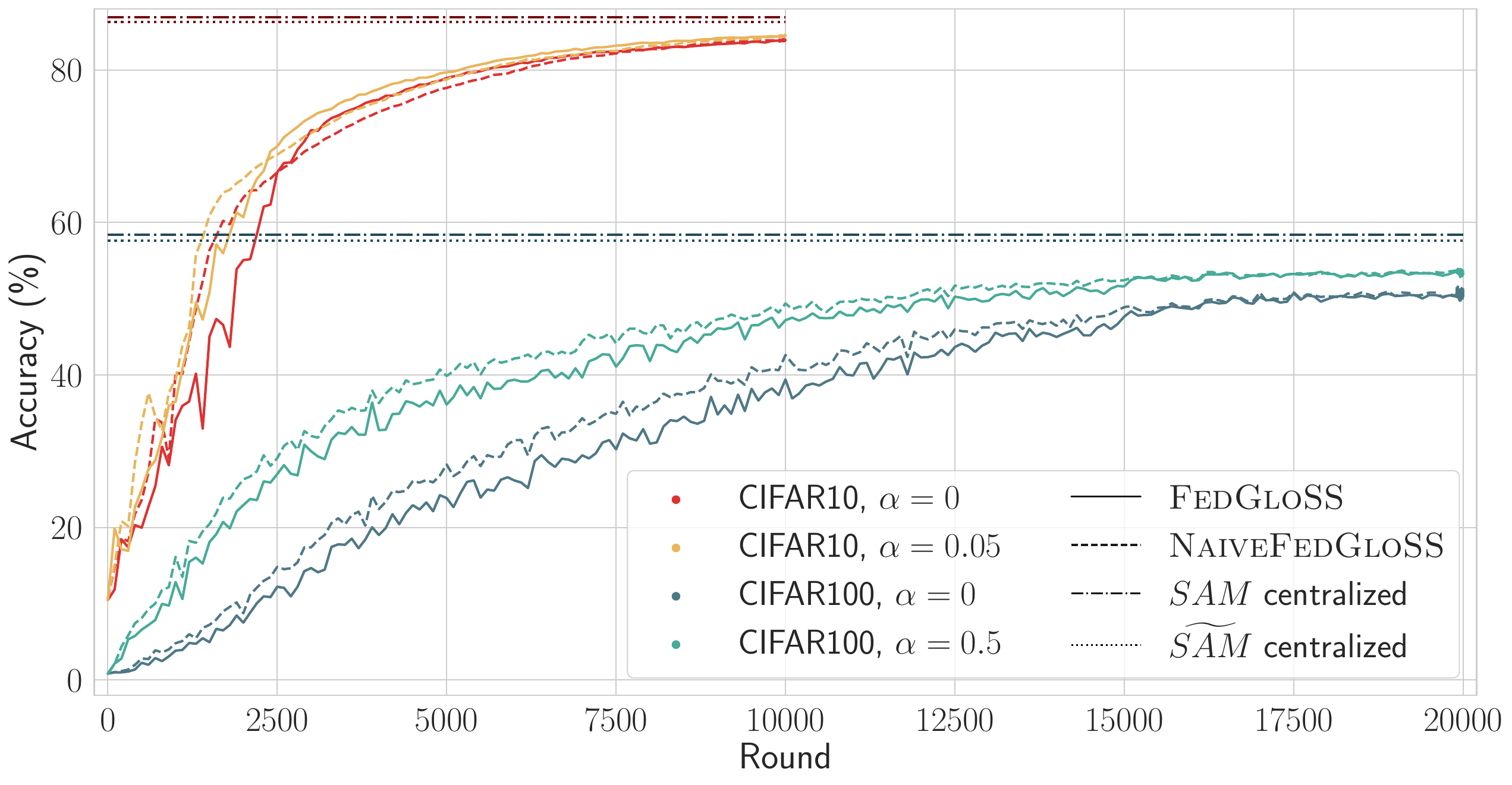}
    \caption{{\textbf{Accuracy trends of \ours \vs \naiveours.} The comparison  includes the centralized upper bounds of \sam and $\widetilde{\text{\sam}}$ (the adaptation of \ours' strategy to the centralized scenario). CNN on \cifarten and \cifar with varying heterogeneity degree ($\alpha$).  \naiveours is $\approx 1.1\times$ faster than its efficient alternative \ours in \cifar, while \ours shows increased convergence speed after $\approx 25\%$ of training rounds in \cifarten. However, both methods reach the same accuracy at the of training. These results motivate the choice of \ours over \naiveours.}}
    \label{fig:ours-vs-naive}
\end{figure}

\begin{figure}[t]
    \centering
    \captionsetup{font=small}
    \includegraphics[width=0.5\linewidth]{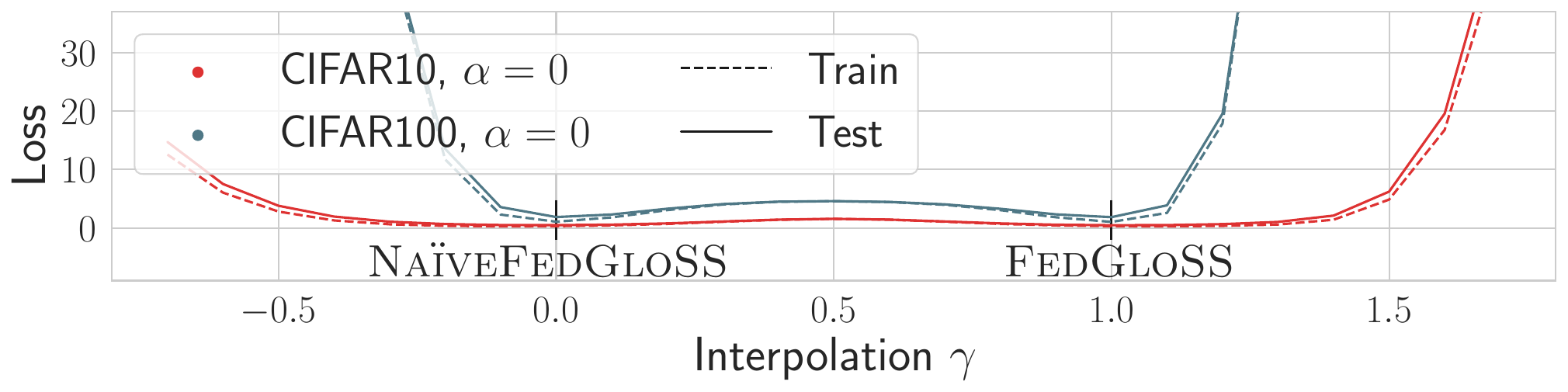}
    \caption{%
    \textbf{Loss barriers} resulting from interpolating \naiveours and \ours' models, which are found in the same basin. \textsc{Cifar} datasets, CNN. Details on the computation in \cref{app:landscape1d}.%
  }
    \label{fig:cmp-naive-v-ours}
\end{figure}

\paragraph{\ours \vs \naiveours.} \cref{fig:ours-vs-naive} deepens the comparison between \ours and its baseline \naiveours, discussed in \cref{subsec:ablation}. In particular, this plot reports the accuracy trends of the two methods, showing that \naiveours is slightly faster ($\approx 1.1\times$) than \ours in \cifar, while \ours surpasses the speed of the baseline after $\approx 25\%$ of training in \cifarten. However, both methods reach the same accuracy at the of training. In addition, it is to be noted that \ours \textit{halves} the communication cost \wrt \naiveours by transmitting half the number of bits at each round, as it eliminates the need to invoke the clients twice to compute the updates. Lastly, our sharpness approximation does not steer the optimization path: models trained with \ours and \naiveours end up in the same basin (no loss barrier), with similar flatness (\cref{fig:cmp-naive-v-ours}). These insights further support our choice of the efficient strategy of \ours over \naiveours.

\paragraph{Convergence speed.} Since all methods are compared over a fixed amount of communication rounds, \textbf{higher final accuracy implies faster convergence}. Since \ours consistently outperforms the other state-of-the-art algorithms taken into account, it is guaranteed to converge faster, as also shown in the accuracy trends (\cref{fig:acc_c10,fig:acc_c100,fig:acc_c100_rn18}).

\subsection{The Impact of Server-side $\rho_s$}
\begin{figure}[h]
    \centering
    \captionsetup{font=small}
    \captionsetup[sub]{font=small}
    \subfloat[][\ours $\rho_s$ with different  architectures\label{fig:serv_rho:acc}]{\includegraphics[width=.35\linewidth]{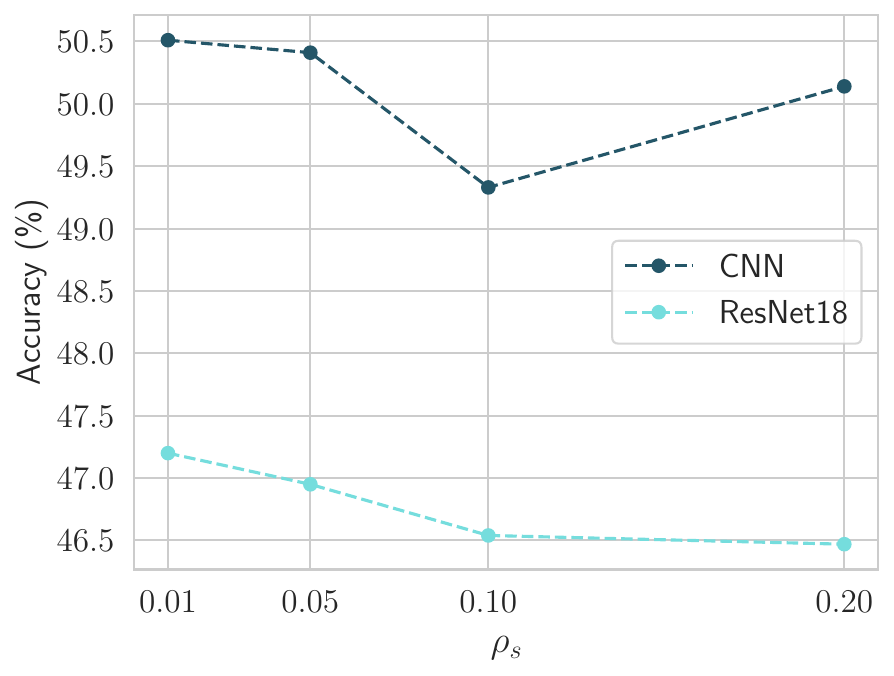}}
    \quad\quad
    \subfloat[][\ours $\rho_s$ \vs $\lambda_1$\label{fig:serv_rho:rho_eig}]{\includegraphics[width=.305\linewidth]{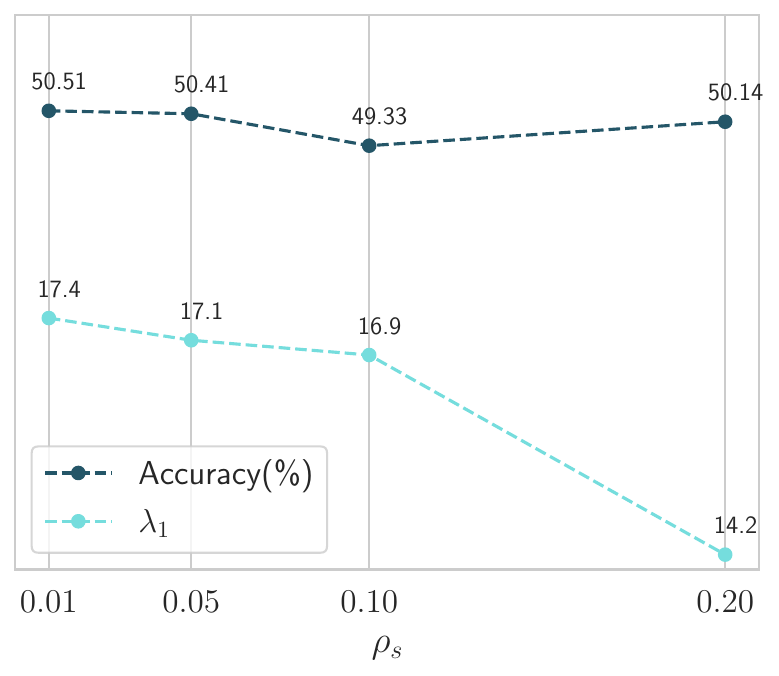}}
    \caption{\cifar. \textbf{(a):} Accuracy (\%) of \ours when varying the server-side \sam $\rho_s$, with different heterogeneity and architecture (CNN with $\alpha=0$ and ResNet18 with $\alpha=0.5$). Smaller values of $\rho_s$ lead to better performances. \textbf{(b):} \ours $\rho_s$ \vs maximum Hessian eigenvalue $\lambda_1$ on CNN with $\alpha=0$. Larger values of $\rho_s$ lead to \textcolor{TealBlue}{lower eigenvalues} with minimum loss in \textcolor{Blue}{accuracy}.}
    \label{fig:serv_rho}
\end{figure}

\cref{fig:serv_rho} analyzes the impact of $\rho_s$ on the performance of the global model, both in terms of accuracy (\cref{fig:serv_rho:acc}) and flatness of the solution (\cref{fig:serv_rho:rho_eig}). In details, \cref{fig:serv_rho:acc} compares the accuracy of the global model trained on \cifar when varying the model architecture (CNN \vs ResNet18) and the data heterogeneity ($\alpha=0$ \vs $\alpha=0.5$). In all the configurations, we note that a smaller value of $\rho_s$ usually leads to the best results. \cref{fig:serv_rho:rho_eig} instead focuses on the CNN in the most heterogeneous setting ($\alpha=0$) and compares the reached accuracy with the corresponding maximum Hessian eigenvalue $\lambda_1$ when varying $\rho_s$. A larger server-side $\rho_s$ corresponds to a smaller $\lambda_1$, \ie, a flatter region in the global loss landscape. 

\section{Implementation Details}
\label{app:exp_details}
This section delves into a comprehensive description of the datasets and models utilized throughout this paper, specifying the Deep Learning framework and the hardware employed (\cref{app:datasets}). 
Additionally, we present the area of the hyper-parameters' space explored during the fine-tuning process in order to yield optimal results (\cref{app:hp}).

\subsection{Datasets and Models}
\label{app:datasets}
\cref{tab:ds-stats} provides a comprehensive overview of each dataset's general statistics. This includes the number of training clients participating in the process and the total number of samples used to construct both the training and test sets.
\begin{table}[b]
    \centering
    \captionsetup{font=small}
    \caption{Datasets' description with their general statistics on the size and number of clients.}
    \footnotesize
    \begin{tabular}{lccccc}
        \toprule
        \textbf{Dataset} & \textbf{Train clients} && \textbf{Train samples} && \textbf{Test samples} \\
        \midrule
        \cifarten & 100 && 50,000 && 10,000 \\
        \cifar & 100 && 50,000 && 10,000 \\
        \gld & 1262 && 164,172 && 19,526 \\
        \bottomrule
    \end{tabular}
    \label{tab:ds-stats}
\end{table}

\subsubsection{\cifarten and \cifar}
\begin{figure}
    \centering
    \captionsetup{font=small}
    \captionsetup[sub]{font=small}
    \subfloat[][\cifarten]{\includegraphics[width=.48\linewidth]{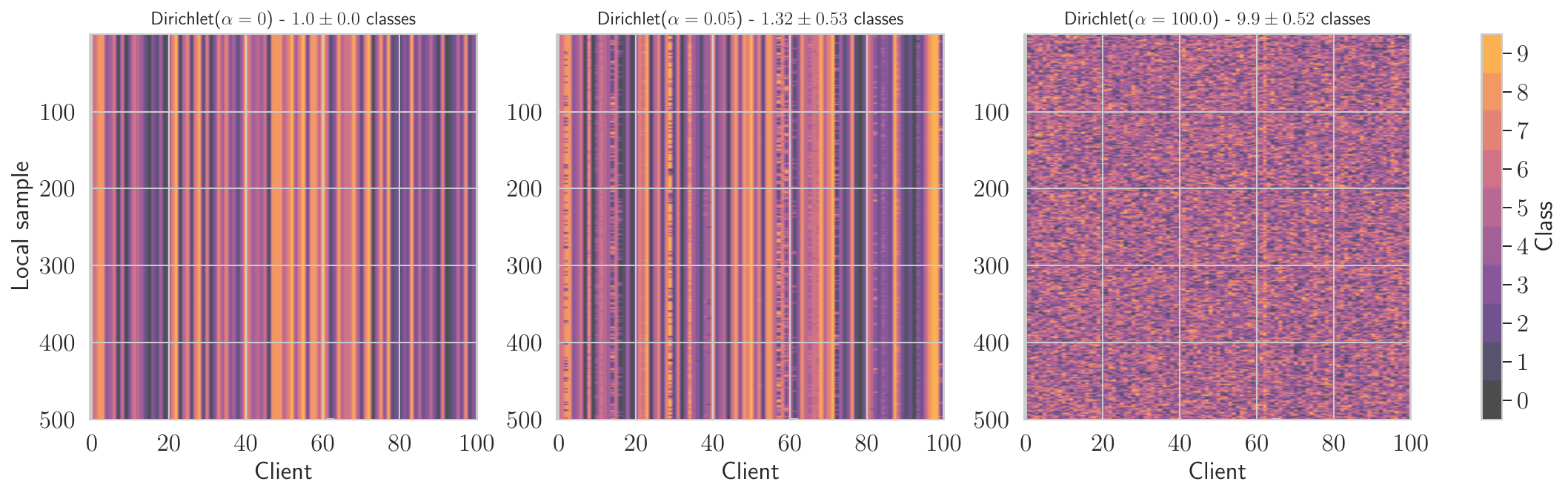} \label{subfig:c10-distr}}
    \subfloat[][\cifar]{\includegraphics[width=.48\linewidth]{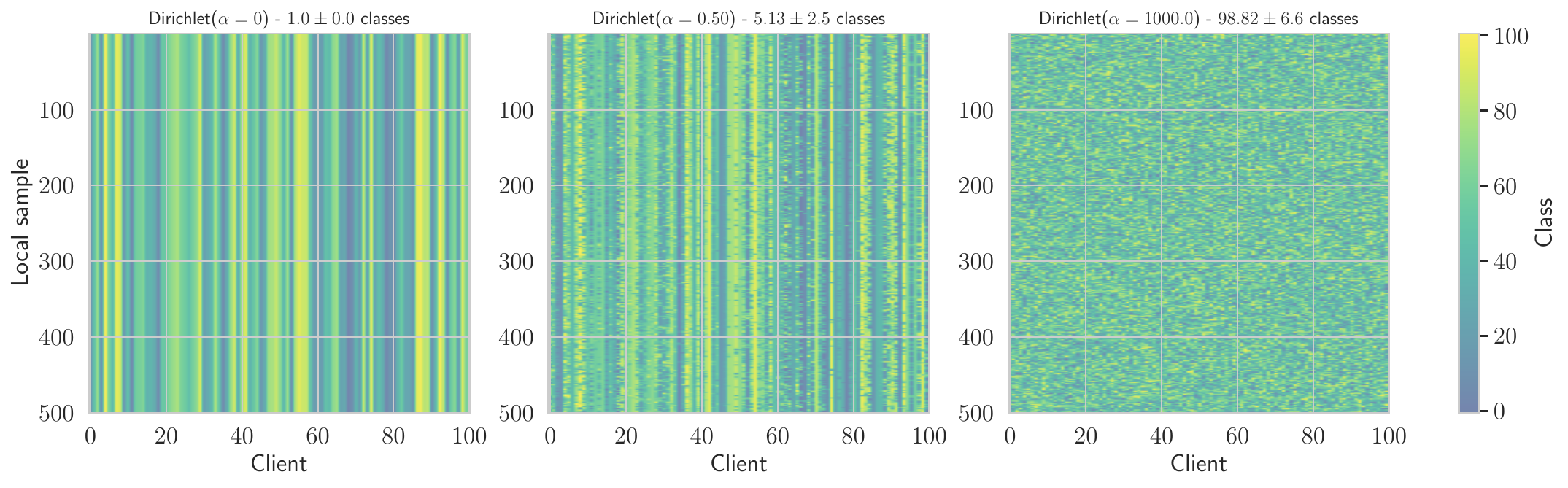} \label{subfig:c100-distr}}\\
    \caption{\cifarten (\emph{left}) and \cifar (\emph{right}) data distribution across clients with the heterogeneity levels tested in the experiments. On top of each chart we report the average number of classes seen by each client.}
    \label{fig:cifar-distributions}
\end{figure}

We adapted these two well-known image classification datasets to the FL scenario by replicating the splits among clients proposed by \citep{hsu2019measuring}.
Both datasets are split evenly among 100 clients, thus each of them has access to 500 data samples. This partitioning is performed according to a Latent Dirichlet Allocation (LDA) on the labels. In practice, each local dataset follows a multinomial distribution drawn according to a symmetric Dirichlet distribution with concentration parameter $\alpha$. The higher the value of this parameter is, the more the local datasets resemble a homogeneous scenario, in the limit case $\alpha=0$ each client has access to one only class of images. In our experiments we tested $\alpha \in \{0, 0.05, 1, 5, 10, 100\}$ and $\alpha \in \{0, 0.5, 1000\}$ for \cifarten and \cifar, respectively. Our choice of $\alpha$ aligns with FL literature \cite{acar2021federated,caldarola2022improving,hsu2020federated,sun2023dynamic,sun2023fedspeed}, where $\alpha < 1$ is standard for testing challenging settings. 
Also, for $\alpha>0.6$, method differences are less apparent \cite{zaccone2023communication}. \cref{subfig:c10-distr,subfig:c100-distr} show how data is distributed across clients in all the experimental settings for these two datasets. 
Both datasets are pre-processed by applying random crops and random horizontal flips.

\paragraph{Models.} We trained a Convolutional Neural Network (CNN) inspired by the LeNet-5 architecture \citep{lecun1998gradient}, as proposed by \citep{hsu2020federated}. The network comprises two 64-channels convolutional layers, both using $5 \times 5$ kernels and followed by $2 \times 2$ max-pooling layers. This is succeeded by two fully-connected layers with 384 and 192 units, respectively. The final output layer is adapted to the specific number of classes in the dataset.

To explore deeper and more expressive architectures, we also trained a ResNet18 \citep{he2015deep} on \cifar with $\alpha=0.5$ and \cifarten with $\alpha=0.05$. We replaced the standard Batch Normalization layers \citep{ioffe2015batch} with Group Normalization layers \citep{wu2018group} due to their demonstrated effectiveness in handling skewed data distributions in FL \citep{hsieh2020non}.
We carried out the experiments with the CNN model using PyTorch \citep{pytorch} and the ResNet18 ones using FedJAX \citep{fedjax2021}.

\subsubsection{\gld}
To achieve a comprehensive understanding of the efficacy of the proposed method, a thorough analysis was undertaken utilizing large-scale real-world datasets.
Specifically, we use the \gld dataset \citep{hsu2020federated}, a repository encompassing 164,172 images depicting 2028 distinct landmarks, distributed among 1262 clients. %

\paragraph{Models.} The model employed for training is a MobileNetV2 \citep{sandler2018mobilenetv2, hsu2020federated}, replacing batch normalization with group normalization layers and pre-trained on ImageNet \citep{deng2009imagenet}. The tests on this dataset were carried out on a cluster of NVIDIA A100 40GB, using our FedJAX codebase.

\subsection{Hyperparameters}
\label{app:hp}
\paragraph{Grid search.} In \cref{tab:base-hp} we report the training hyperparameters associated to each dataset and model pairing, while  \cref{tab:hp} summarizes the hyper-parameters search grid tested for each method (in bold the chosen ones). All runs are averaged over 3 seeds. In addition, following previous works \citep{acar2021federated,caldarola2022improving,caldarola2023window}, we report the averaged accuracy of the last 100 rounds to reduce the noise typical of heterogeneous FL settings.

\paragraph{Local hyperparameters.} As for local hyperparameters, we tested \ours' robustness to $E=2$ and $\eta_l = 0.001$ on \cifarten with $\alpha=0$. A smaller $\eta_l$ had the greatest impact, with \fedavg and \fedsam losing 45\% and 77\% of their performance (more local steps degrade performance in heterogeneous settings), respectively. Notably, \ours loses only 29\%. We attribute this to the pseudo-gradient consistently pointing toward higher-loss regions, ensuring a reliable ascent step approximation regardless of local hyperparameters.

\paragraph{Scheduling of $\rho$.} While running our experiments on \ours, we noticed that a larger value of local $\rho$ allowed to reach the best final accuracy, while a smaller $\rho$ achieved faster convergence in the initial training stages. Following this insight, we schedule the value of $\rho$ for $T_s$ rounds as
\begin{equation*}
    \rho(t) = 
    \begin{cases}
      \rho_0 + \nicefrac{(\rho - \rho_0)}{T_s} \cdot t  & \text{if $t \leq T_s$}\\
      \rho & \text{otherwise,}\\
    \end{cases}
\end{equation*}
starting from the value $\rho_0 = 0.001$. 
\setlength{\tabcolsep}{5pt}
\begin{table}[]
    \centering
    \captionsetup{font=small}
    \caption{General training hyper-parameters common to all methods, distinguished by dataset and model architecture. Symbols: local epochs $E$, local learning rate $\eta$, weight decay $wd$, client-side momentum $\beta_l$, batch size $B$.}
    \small
    \begin{tabular}{lccccccccc}
    \toprule
         \multirow{2}{*}{\textbf{Dataset}} & \multirow{2}{*}{\textbf{Model}} & \multirow{2}{*}{\textbf{Rounds}} & \textbf{Clients} && \multicolumn{5}{c}{\textbf{Client optimization}}\\
         \cline{6-10}
         &&& \textbf{per round} && $E$ & $\eta$ & $wd$ & $\beta_l$ & $B$ \\
         \midrule
         \cifarten & CNN & 10000 & 5 && 1 & \multirow{1}{*}{$10^{-2}$} &$4 \cdot 10^{-4}$ & 0 & 64 \\ \cdashline{1-10}[1pt/3pt]
         \multirow{2}{*}{\cifar} & CNN & 20000 & 5 && 1 & $10^{-2}$ & $4 \cdot 10^{-4}$ &  0 & 64\\
          & ResNet18-GN & 10000 & 10 && 1 &  $10^{-2}$ & $10^{-5}$ & 0.7 & 64 \\
        \cdashline{1-10}[1pt/3pt]
         \gld & \multirow{1}{*}{MobileNetv2} & 1300 & 50 && \multirow{1}{*}{5} & \multirow{1}{*}{0.1} & \multirow{1}{*}{$4 \cdot 10^{-5}$} & \multirow{1}{*}{0} & \multirow{1}{*}{64} \\
    \bottomrule
    \end{tabular}%
    \label{tab:base-hp}
\end{table}

\begin{table}[]
    \centering
    \captionsetup{font=small}
    \caption{Search grid used to find optimal hyper-parameters for each combination of method, dataset and model. We highlight the best performing values in \textbf{bold}.}
    \resizebox{\linewidth}{!}{
    \begin{tabular}{lccccccccc}
        \toprule
            \multirow{2}{*}{\textbf{Method}} & \multirow{2}{*}{\textbf{HParam}} && \multicolumn{2}{c}{\textbf{\cifarten}} && \multicolumn{2}{c}{\textbf{\cifar}} && \multirow{2}{*}{\textbf{\gld}} \\
        \cline{7-8} \cline{4-5}
            & && CNN & ResNet18 && CNN & ResNet18 \\
        \midrule
            \fedsam & $\rho$ && [0.05, 0.1, \textbf{0.15}, 0.2] &[\textbf{0.01}, {0.02}, 0.05]&& [0.005, \textbf{0.01}, 0.02, 0.05] & [\textbf{0.01}, 0.02, 0.05] && [\textbf{0.05}] \\
        \midrule
            \fedprox & $\mu$ && [0.001, 0.01, \textbf{0.1}] &[0.001, \textbf{0.01}, {0.1}] && [0.001, 0.01, \textbf{0.1}] & [0.001, 0.01, \textbf{0.1}] && [0.001, 0.01, \textbf{0.1}] \\
        \midrule
            \multirow{2}{*}{\feddyn} & $\alpha$ && [0.001, \textbf{0.01}, 0.1] & [0.001, \textbf{0.01}, 0.1] && [0.001, \textbf{0.01}, 0.1] & [\textbf{0.01}] && [\textbf{0.001}, 0.01]\\
            & $\rho$ (\sam-based only) && [\textbf{0.15}] & [\textbf{0.01}]&& [0.01, \textbf{0.02}] & [\textbf{0.01}] && [\textbf{0.05}]\\
        \midrule
            \multirow{3}{*}{\fedspeed} & $\rho$ && [0.05, 0.1, \textbf{0.15}, 0.2] &[\textbf{0.01}]&& [0.005, \textbf{0.01}, 0.02, 0.05] & [\textbf{0.01}] && [\textbf{0.05}] \\
            & $\alpha$ && [0.9, \textbf{0.95}, 0.99] &[0.9, \textbf{0.95}] && [0.9, \textbf{0.95}, 0.99] & [0.9, \textbf{0.95}] && [\textbf{0.95}] \\
            & $\lambda$ && [10, \textbf{100}, 1000] &[10, 100, \textbf{1000}]&& [10, 100, \textbf{1000}] & [10, 100, \textbf{1000}] && [100, \textbf{1000}]\\
        \midrule
            \fedgamma & $\rho$ && [\textbf{0.15}] & [\textbf{0.01}]&& [\textbf{0.01}] & [\textbf{0.01}] && [\textbf{0.05}]\\
        \midrule
            \multirow{2}{*}{\fedsmoo} & $\rho$ && [0.05, 0.1, \textbf{0.15}, 0.2] & [\textbf{0.01}] && [0.005, 0.01, 0.05, \textbf{0.1}, 0.2] & [\textbf{0.01}] && [0.05, \textbf{0.1}, 0.2, 0.3]\\
            & $\beta$ && [5, \textbf{10}, 100] & [1, 2, \textbf{5}, 10] && [10, \textbf{100}] & [5, \textbf{10}, 100] && [10, \textbf{50}, 100, 1000] \\
        \midrule
            \multirow{4}{*}{\textbf{\ours (ours)}} & $\rho_s$ && [0.01, 0.1, \textbf{0.15}] &[\textbf{0.01}, 0.05, 0.1, 0.5] && [\textbf{0.01}, 0.05, 0.1, 0.2] & [\textbf{0.01}, 0.05, 0.1, 0.5] && [\textbf{0.005}, 0.01, 0.02] \\
            & $\rho$ && [0.05, 0.1, \textbf{0.15}, 0.2] & [\textbf{0.01}] && [0.05, 0.1, \textbf{0.2}] & [\textbf{0.01}] && [0.05, 0.1, 0.2, \textbf{0.3}] \\
            & $\beta$ && [5, \textbf{10}, 100] & [\textbf{1}, 2, 5, 10]&& [10, \textbf{100}] & [\textbf{5}, 10, 100] && [10, \textbf{50}, 100]\\
            & $T_s$ && [1000, \textbf{2000}, 4000] & [\textbf{0}]&& [1000, 2000, 5000, 10000, \textbf{15000}] & [\textbf{0}] && [\textbf{0}] \\
        \bottomrule
    \end{tabular}}
    \label{tab:hp}
\end{table}

\section{Flatness Analysis}
\label{app:landscape}
This section describes the procedure to compute the visualization of the loss landscapes and the Hessian eigenvalues. 

\subsection{Visualizing 3D Loss Landscapes}
We leverage techniques from \citep{li2018visualizing} to visualize the loss landscapes of our models. We adapt their code to work with our specific datasets and network architectures. The process involves calculating the loss function along random directions in the parameter space. This is achieved by perturbing the model's parameters within a defined range. In our visualizations, we constrain these perturbations to occur within the range of $[-1,1]$ for both the $x$ and $y$ axes. To ensure consistent comparisons across models (\eg, as seen in \cref{fig:landscape_fedgloss_fedsam}), we utilize the same set of random directions for all models.

\subsection{Visualizing 1D Loss Landscapes}
\label{app:landscape1d}
\cref{fig:cmp-naive-v-ours} shows the interpolation of \ours and \naiveours's models. Given their respective weights $\w_\ours$ and $\w_\naiveours$, the interpolation is computed using a coefficient $\gamma$ as
\begin{equation}
    \w = \gamma \cdot \w_\ours + (1 - \gamma) \cdot \w_\naiveours.
\end{equation}
For each $\gamma\in[-1, 2]$, $\w$ is tested on the training or test sets, and the plot reports the computed loss. 
The resulting interpolation indicates that there is no loss barrier between the two models, suggesting they lie within the same basin. Additionally, the 1D geometry of the emerging loss landscape reveals that both models converge to a flat minimum when evaluated on both \cifar and \cifarten.

\subsection{Hessian Eigenvalues for Flatness Measure}
Following prior work \citep{foret2020sharpness,garipov2018loss,li2018visualizing,caldarola2022improving}, we use the spectrum of the Hessian matrix to quantify the \textit{flatness} of the loss landscape. Here, lower maximum eigenvalues correspond to flatter landscapes, implying less sharpness. To compute these eigenvalues (denoted by $\lambda_1$ in the main paper), we employ the stochastic power iteration method \citep{xu2018accelerated} with a maximum of 20 iterations, referring to the code of \citet{hessian-eigenthings}.

\end{document}